%% file: main.tex
\documentclass{article} 
\usepackage{iclr2027_conference,times}

\input{math_commands.tex}

\usepackage{hyperref}
\usepackage{url}

\title{FLOORA: A Human-Aligned Domain-Specific Language Model for Architectural Design}

\newcommand{\core}{\textsuperscript{\ensuremath{\dagger}}}

\newcommand{\corecorr}{\textsuperscript{\ensuremath{\dagger},*}}
\newcommand{\adsk}{\textsuperscript{\ensuremath{\diamond}}}

\author{
Sahand Rezaei-Shoshtari\corecorr
\And 
Patryk Wozniczka\core
\And
Shu Ishida\core
\And
Gregg Streuber\core
\And
Farnoosh Javadi\core
\And 
Jeffrey Landes\core
\And
Angela Ju 
\And 
Muhammad Azam
\And 
Bryan Lim\adsk
\And 
Johan Luttun
\And 
Indrajeet Haldar
\And 
Jonathan Shaw
\And 
Beatriz Guerra
\And 
Ivan Sosnovik
\And
James Stoddart\core
\And 
Robert Giaquinto\core
\And
Adam Gaier\corecorr
\AND
\makebox[\textwidth][c]{\normalfont\textit{Autodesk Research}}
\AND
\makebox[\textwidth][c]{
  \normalfont\footnotesize
  \begin{tabular}{@{}l@{}}
    \textsuperscript{\ensuremath{\dagger}}\,Core contributor\\[0.4ex]
    \textsuperscript{\ensuremath{\diamond}}\,Work done while at Autodesk\\[0.4ex]
    \textsuperscript{*}\,Corresponding authors:
    \texttt{\{sahand.rezaei-shoshtari, adam.gaier\}@autodesk.com}
  \end{tabular}
}
}

\usepackage[nameinlink]{cleveref}
\usepackage{graphicx}
\usepackage{listings}
\usepackage{wrapfig}
\usepackage{subcaption}
\usepackage{tcolorbox}
\usepackage{enumitem}
\usepackage{booktabs}
\usepackage{pdflscape}
\usepackage{booktabs}
\usepackage{adjustbox}
\tcbuselibrary{breakable}
\usepackage{tikz}
\usepackage{xcolor}
\usepackage{inconsolata}
\usetikzlibrary{positioning,arrows.meta}

\newcommand{\tagc}[2]{\textcolor{#1!75!black}{#2}}
\newcommand{\todo}[2][]{%
    \textcolor{orange!80!red}{[TODO\if\relax\detokenize{#1}\relax\else\ (#1)\fi]: #2}
}
\definecolor{dslOrange}{HTML}{D96B1A}
\definecolor{feedbackBlue}{HTML}{1F78A6}
\definecolor{postPurple}{HTML}{6B5FC7}
\definecolor{dslGreen}{HTML}{4F9A6A}
\definecolor{coreTeal}{HTML}{2C9C8F}
\definecolor{massingBlue}{HTML}{2F80C9}
\definecolor{finalRed}{HTML}{D6453D}
\definecolor{softInk}{HTML}{2F3A45}

\crefname{section}{Section}{Sections}
\crefname{algorithm}{Algorithm}{Algorithms}
\crefname{definition}{Definition}{Definitions}
\crefname{assumption}{Assumption}{Assumptions}
\crefname{thm}{Theorem}{Theorems}
\crefname{prop}{Proposition}{Propositions}
\crefname{claim}{Claim}{Claims}
\crefname{equation}{Equation}{Equations}
\crefname{figure}{Figure}{Figures}
\crefname{appendix}{Appendix}{Appendices}
\crefname{table}{Table}{Tables}
\crefname{chapter}{Chapter}{Chapters}
\crefname{lstlisting}{Listing}{Listings}

\lstdefinelanguage{DSL}{
  morekeywords={building,structure,massing,grids,spaces,columns,beams,core_walls,
occupancy_type,storeys,level,elevation,material,height,polygon,gridline,column,beam,wall,size},
  sensitive=true
}
\iclrfinalcopy 
\begin{document}

\maketitle

\begin{abstract}
Foundation models are powerful generators, but many engineering domains require structured representations that general-purpose systems handle poorly. We introduce FLOORA (Floor Layout Optimization with RL Alignment), a family of small domain-specific language (DSL) models for architectural layout generation. With specialized data and alignment, our 0.6B model outperforms much larger frontier models, achieving VLM judge win rates up to \textbf{92.0\%} on out-of-distribution real-world buildings and \textbf{96.0\%} on synthetic buildings. Human evaluations further corroborate these results, with FLOORA selected as the best model in \textbf{89.3\%} of evaluations. FLOORA combines a token-efficient DSL, custom tokenization, domain-specific pretraining, supervised fine-tuning (SFT), and reinforcement learning (RL) with learned human-preference and verifiable rewards. This pipeline improves architectural and geometric validity, supported by extensive empirical evaluation and ablation studies.  Although focused on architecture, our results suggest that similar domain-specific recipes may be useful in other engineering domains with structured, verifiable outputs. Datasets, models, and inference code are available at \href{https://github.com/AutodeskAILab/floora}{\texttt{https://github.com/AutodeskAILab/floora}}.
\end{abstract}

\section{Introduction}
\label{sec:intro}
\input{content/intro}

\par\nopagebreak[4]
\section{Related Work}
\label{sec:related_work}
\input{content/related_work}

\section{Domain-Specific Language (DSL)}
\label{sec:dsl}
\input{content/dsl}

\section{Data Generation and Processing}
\label{sec:data}
\input{content/data}

\section{DSL Pre-training}
\label{sec:pre-training}
\input{content/pre-training}

\section{DSL Post-training}
\label{sec:post-training}
\input{content/post-training}

\section{Results}
\label{sec:results}
\input{content/results}

\section{Limitations}
\label{sec:limitations}
\input{content/limitations}

\section{Conclusion}
\label{sec:conclusion}
\input{content/conclusion}

\input{content/statements}


\bibliography{refs}
\bibliographystyle{iclr2027_conference}

\clearpage
\appendix

\section{Additional Details on the Domain-Specific Language (DSL)}
\label[appendix]{app:dsl}
\input{appendix/dsl}

\section{Additional Details on Data Generation and Processing}
\label[appendix]{app:data}
\input{appendix/data}

\section{Additional Details on the Human Feedback Data}
\label[appendix]{app:human_feedback_collection}
\input{appendix/human_feedback_collection}

\section{Additional Details on the Training Setup}
\label[appendix]{app:training_hparams}
\input{appendix/training_hparams}

\section{Additional Details on the Verifiable Rewards}
\label[appendix]{app:verifiable_rewards}
\input{appendix/verifiable_rewards}

\section{Additional Details on the VLM Judge}
\label[appendix]{app:vlm_judge}
\input{appendix/vlm_judge}

\clearpage
\section{Frontier Model Baselines and Inference Protocol}
\label[appendix]{app:frontier_model_inference}
\input{appendix/frontier_model_inference}

\section{Additional Experimental Results}
\label[appendix]{app:results}
\input{appendix/results}

\clearpage
\section{Ablation Studies}
\label[appendix]{app:ablations}
\input{appendix/ablations}

\end{document}

%% file: math_commands.tex
\usepackage{amsmath,amsfonts,bm}

\def\eqref#1{equation~\ref{#1}}

\def\1{\bm{1}}

\DeclareMathAlphabet{\mathsfit}{\encodingdefault}{\sfdefault}{m}{sl}
\SetMathAlphabet{\mathsfit}{bold}{\encodingdefault}{\sfdefault}{bx}{n}



%% file: content/intro.tex
Large language models (LLMs) \citep{brown2020language} are increasingly evaluated on specialized scientific and engineering tasks \citep{zhou2025engibench,sun2024scieval,heesch2025evaluating,rein2023gpqa}, yet it remains unclear whether such domains are best served by continuing to scale general-purpose models \citep{kaplan2020scaling,hoffmann2022training} or by training smaller models around domain-specific data \citep{beltagy2019scibert,lee2020biobert,gu2021domain,ouyang2022training}.

\begin{figure}[b!]
    \centering
    \includegraphics[width=\textwidth]{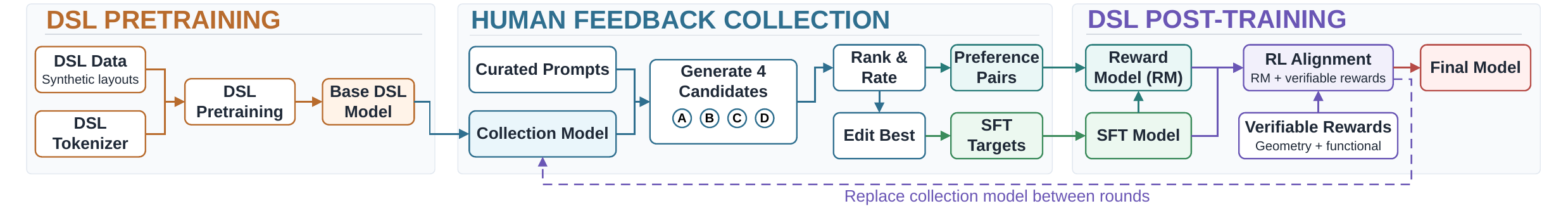}
    \caption{Overview of our DSL pre-training, human feedback collection, and post-training pipeline.}
    \label{fig:rlhf_diagram}
\end{figure}

Architecture, Engineering, and Construction (AEC) provides a useful test case because building designs are structured objects with strict geometric constraints, multiple valid representations, and quality criteria combining verifiable rules with expert judgment. Unlike open-ended text generation, building-design models must produce outputs that can be parsed, validated, and used as geometry. Recent work has benchmarked frontier models on AEC tasks \citep{mankodiya2026aec,liang2026aecbench,kondratenko2026aecv}, revealing both their capabilities and limitations. Complementing this work, we ask whether compact DSL models can match or outperform frontier models on an architectural generation task given the right representation, data pipeline, and post-training recipe. We study multifamily residential layout generation from building massings, synthesizing labeled, geometrically parseable floor-plan spaces that satisfy design constraints, and investigate how expert human feedback aligns these models with architectural judgment.

A central obstacle is representation. Standard building formats such as Industry Foundation Classes (IFC) \citep{buildingSMART_IFC} are too verbose and relational for efficient autoregressive language modeling. We introduce a compact DSL that encodes building metadata, massing geometry, and labeled floor-plan spaces as structured text. It can be deterministically parsed, validated, normalized, and converted back to geometry, enabling architectural generation as language modeling while preserving downstream usability. Compared with IFC, it is up to 15$\times$ more compact while retaining the geometric and semantic information needed for conceptual layout generation.

\begin{figure}[t!]
    \centering
    \includegraphics[width=0.9\textwidth]{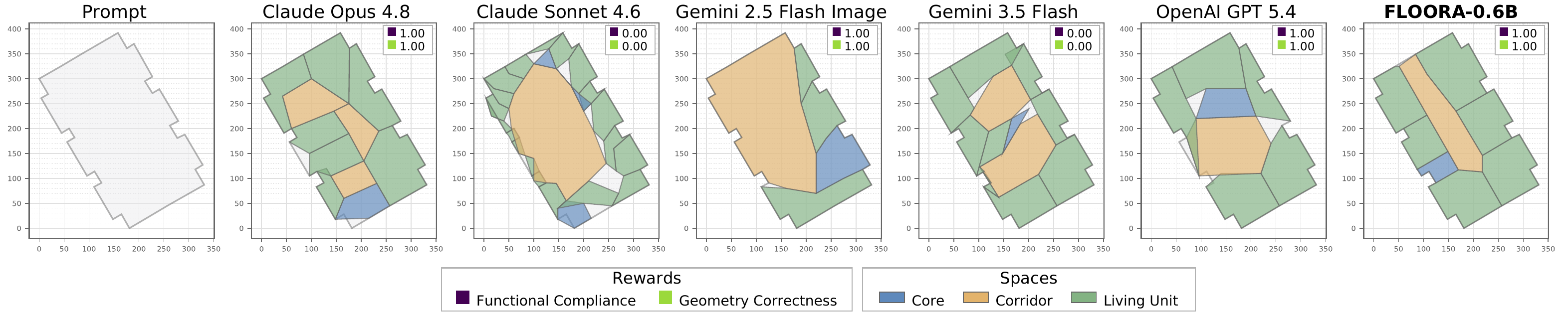}
    \caption{Qualitative comparison of FLOORA-0.6B and frontier models on a real-world test sample. FLOORA outperforms the baselines on geometric and functional checks and architectural quality.}
    \label{fig:qualitative_comparison_main}
\end{figure}

We build a pre- and post-training pipeline around this representation, shown in \cref{fig:rlhf_diagram}. We pre-train compact models on large-scale synthetic architectural data with a DSL tokenizer and evaluate them on out-of-distribution (OOD) real-world OpenStreetMap massings \citep{OpenStreetMap}. Practicing architects then rate, rank, and edit generated layouts, yielding 10k SFT samples and 90k pairwise preferences. Finally, we combine learned and verifiable rewards to optimize architect preferences and explicit design constraints simultaneously. 

Notably, this is not a scaling study. Our models are compact, domain-specific, and optimized for structured generation rather than language ability. To test whether the recipe generalizes across backbones, we evaluate Qwen3 \citep{yang2025qwen3} and Pythia \citep{biderman2023pythia} across model sizes. Our results suggest that, for this structured generation task, domain-specific representation and alignment can partially substitute for scale. A 0.6B FLOORA model outperforms much larger frontier models on our layout-generation benchmarks, with \cref{fig:qualitative_comparison_main} showing a qualitative example.

The closest prior work, \citet{yin2025floorplanllama}, represents room layouts with discrete visual tokens and uses architect feedback for RL. We instead tackle the problem of floor layout by combining an architectural DSL with domain-specific pre-training, enabling direct geometric generation and deterministic verification. In addition, our feedback pipeline uses architect edits for SFT, while our RL pipeline jointly optimizes learned architect preferences and verifiable constraints. To the best of our knowledge, FLOORA is the first architectural post-training framework to jointly optimize a learned reward model with verifiable rewards, requiring careful normalization and balancing of the signals.

Beyond architectural layout generation, our framework suggests a general recipe for specialized domains that combine structured representations, verifiable constraints, and expert-defined quality criteria. More broadly, our results suggest that similar domain-specific approaches may offer an effective alternative to relying primarily on model scale for structured generation tasks. \Cref{fig:recipe} summarizes the full recipe. Our contributions are:

\begin{itemize}[itemsep=2pt, topsep=0pt, parsep=0pt, partopsep=0pt,leftmargin=1em]
    \item We show that a token-efficient architectural DSL and custom tokenizer enable effective domain-specific pre-training and post-training, resulting in generalization to OOD real-world massings.
    \item We develop an architect-guided post-training pipeline that uses direct edits and pairwise preferences to align synthetically pre-trained models with actual design preferences.    
    \item We show that combining learned architect preferences with verifiable rewards generally improves layout generation, with greater reward model benefits at larger scales.
    \item We demonstrate that compact domain-specific models can outperform substantially larger frontier models across verifiable metrics, VLM judge pairwise comparisons, and human evaluations.
    \item We release the synthetic and OSM datasets, trained models, and inference code for future research.
\end{itemize}

%% file: content/related_work.tex
\input{figures/tikz/recipe}

\paragraph{Machine Learning for Architectural Layouts.}
Optimization and machine learning for architectural layouts have a long history~\citep{weber2022automated}, typically focusing on room arrangement within homes; we instead model living units, corridors, and cores on multifamily floor plates. Recent generative approaches differ substantially in how they encode geometry. Coordinate-sequence methods use standard model vocabularies: \citet{galanos2023architext} represent type labels and corner coordinates, \citet{leng2023tell2design} and \citet{yin2025floorplan} quantize coordinates, \citet{luo2024dstruct2design} output full-precision polygon vertices, and \citet{de2025tokenizing} combine categorical and continuous room features. Other approaches introduce specialized representations. \citet{armen2024scenescript} generate quantized parametric commands with a custom tokenizer, \citet{klimenko2026hypergraphformer} represent floor plans as space-partition trees from which geometry is recovered procedurally, and \citet{qin2026tokenization} encode massings and rooms using a VQ-VAE. Most closely related, \citet{yin2025floorplanllama} represent unit plans with discrete visual tokens and post-train using an architect-preference reward model.

We take a middle ground between unconstrained coordinate sequences and representations that encode geometry implicitly. Our DSL represents layouts directly as compact sequences of labeled closed polygons. Rather than enforcing geometric validity through the representation itself, we use training data and reinforcement learning to learn geometric correctness and architectural validity.

\paragraph{Domain-Specific Language (DSL) Models.} 
A large body of work has explored adapting pretrained language models to specialized domains, motivated by distributional differences in terminology, structure, and domain knowledge. Early approaches primarily focused on domain-specific pre-training or continued pretraining of encoder models \citep{lee2020biobert, beltagy2019scibert, huang2019clinicalbert, araci2019finbert, chalkidis2020legal, gu2021domain, gupta2022matscibert}. More generally, domain-adaptive pretraining has been shown to improve downstream performance by continuing training on data drawn from the target distribution \citep{gururangan2020don}. More recent work extends domain specialization to generative LLMs, typically through domain-focused continued pretraining of general-purpose models \citep{luo2022biogpt,taylor2022galactica,wu2023bloomberggpt,wu2024pmc,chen2023meditron,labrak2024biomistral, xie2024efficient, colombo2024saullm, yang2023fingpt}, spanning biomedical, scientific, financial, and legal applications. 

DSL models remain limited in AEC. \citet{li2025cad} combine CAD-specific pretraining with instruction tuning for parametric 3D generation, while \citet{govindarajan2025cadmium} fine-tune code language models on sequential CAD data. In architecture, \citet{lin2026qwen} fine-tune a general-purpose LLM on BIM-derived question-answering and reasoning data. In contrast, our work provides an end-to-end pipeline for generative architectural design, spanning domain-specific pre-training through expert-guided post-training including supervised fine-tuning and reinforcement learning.

%% file: figures/tikz/recipe.tex
\begingroup
\definecolor{Ink}{HTML}{172235}
\definecolor{Muted}{HTML}{6F7B8B}
\definecolor{RepFill}{HTML}{E6EEFF}
\definecolor{RepAccent}{HTML}{456DB5}
\definecolor{DataFill}{HTML}{DFF4EF}
\definecolor{DataAccent}{HTML}{2D7B6B}
\definecolor{PreFill}{HTML}{FFF0D7}
\definecolor{PreAccent}{HTML}{A36A16}
\definecolor{HumanFill}{HTML}{FFE4DE}
\definecolor{HumanAccent}{HTML}{B65A46}
\definecolor{SftFill}{HTML}{EEE7FA}
\definecolor{SftAccent}{HTML}{7257A1}
\definecolor{RewardFill}{HTML}{F8E2EE}
\definecolor{RewardAccent}{HTML}{A34972}
\definecolor{RlFill}{HTML}{E7F2DE}
\definecolor{RlAccent}{HTML}{587D42}

\newcommand{\recipegap}{2.2mm}
\newcommand{\recipeleftpad}{0.5mm}
\newcommand{\reciperightpad}{0.5mm}
\newcommand{\recipestepleftpad}{0mm}
\newcommand{\recipestepgap}{0.55mm}
\newcommand{\recipestepnumberwidth}{2.7mm}
\newcommand{\recipetitlegap}{1.0mm}

\tikzset{
recipe stage/.style={
rounded corners=2.2mm,
line width=0.65pt,
minimum height=3.3cm,
inner sep=0pt,
anchor=north west
},
recipe flow/.style={
-{Stealth[length=2.1mm,width=1.5mm]},
draw=Muted!58,
line width=0.75pt
}
}

\newcommand{\recipestep}[3]{%
\noindent
\hspace*{\recipestepleftpad}%
\makebox[\recipestepnumberwidth][r]{%
\color{#1}\fontsize{5.9}{6.4}\selectfont\bfseries #2%
}%
\hspace{\recipestepgap}%
\parbox[t]{%
\dimexpr\linewidth-\recipestepleftpad-\recipestepnumberwidth-\recipestepgap\relax
}{%
\raggedright\color{Ink}\fontsize{7.2}{7.15}\selectfont #3%
}%
\par\vspace{0.06mm}%
}

\newcommand{\recipebody}[5]{%
\begin{minipage}[t][2.68cm][t]{#1}
\vspace{-2.2mm}
\noindent
\hspace*{\recipeleftpad}%
\begin{minipage}{\dimexpr#1-\recipeleftpad-\reciperightpad\relax}
\noindent
\tikz[baseline=(n.base)]\node[
rounded corners=1.1mm,
fill=#4,
text=white,
minimum width=0.45cm,
minimum height=0.45cm,
inner sep=0pt,
font=\fontsize{6.8}{6.8}\selectfont\bfseries
] (n) {#2};%
\hspace{\recipetitlegap}%
\parbox[c]{\dimexpr\linewidth-0.45cm-\recipetitlegap\relax}{%
\color{Ink}\fontsize{7.8}{8.3}\selectfont\bfseries #3%
}%
\par\vspace{-0.5mm}
\hyphenpenalty=10000
\exhyphenpenalty=10000
#5
\end{minipage}
\end{minipage}%
}

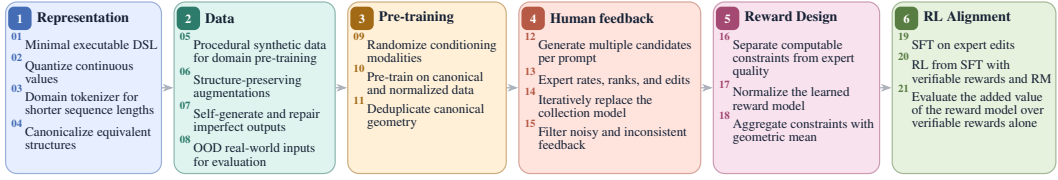
\begin{figure}[t!]
\centering
\resizebox{\linewidth}{!}{%
\begin{tikzpicture}[x=1cm,y=1cm]
  \node[
    recipe stage,
    minimum width=3.00cm,
    fill=RepFill,
    draw=RepAccent!55
  ] (rep) at (0,2.68) {
    \recipebody{3.00cm}{1}{Representation}{RepAccent}{
      \recipestep{RepAccent}{01}{Minimal executable DSL}
      \recipestep{RepAccent}{02}{Quantize continuous values}
      \recipestep{RepAccent}{03}{Domain tokenizer for shorter sequence lengths}
      \recipestep{RepAccent}{04}{Canonicalize equivalent structures}
    }
  };
  \node[
    recipe stage,
    minimum width=3.10cm,
    fill=DataFill,
    draw=DataAccent!55
  ] (data) at ([xshift=\recipegap]rep.north east) {
    \recipebody{3.10cm}{2}{Data}{DataAccent}{
      \recipestep{DataAccent}{05}{Procedural synthetic data for domain pre-training}
      \recipestep{DataAccent}{06}{Structure-preserving augmentations}
      \recipestep{DataAccent}{07}{Self-generate and repair imperfect outputs}
      \recipestep{DataAccent}{08}{OOD real-world inputs for evaluation}
    }
  };
  \node[
    recipe stage,
    minimum width=3.05cm,
    fill=PreFill,
    draw=PreAccent!55
  ] (pre) at ([xshift=\recipegap]data.north east) {
    \recipebody{3.05cm}{3}{Pre-training}{PreAccent}{
      \recipestep{PreAccent}{09}{Randomize conditioning modalities}
      \recipestep{PreAccent}{10}{Pre-train on canonical and normalized data}
      \recipestep{PreAccent}{11}{Deduplicate canonical geometry}
    }
  };
  \node[
    recipe stage,
    minimum width=3.50cm,
    fill=HumanFill,
    draw=HumanAccent!55
  ] (human) at ([xshift=\recipegap]pre.north east) {
    \recipebody{3.50cm}{4}{Human feedback}{HumanAccent}{
      \recipestep{HumanAccent}{12}{Generate multiple candidates per prompt}
      \recipestep{HumanAccent}{13}{Expert rates, ranks, and edits}
      \recipestep{HumanAccent}{14}{Iteratively replace the collection model}
      \recipestep{HumanAccent}{15}{Filter noisy and inconsistent feedback}
    }
  };

  \node[
    recipe stage,
    minimum width=3.20cm,
    fill=RewardFill,
    draw=RewardAccent!55
  ] (reward) at ([xshift=\recipegap]human.north east) {
    \recipebody{3.20cm}{5}{Reward Design}{RewardAccent}{
      \recipestep{RewardAccent}{16}{Separate computable constraints from expert quality}
      \recipestep{RewardAccent}{17}{Normalize the learned reward model}
      \recipestep{RewardAccent}{18}{Aggregate constraints with geometric mean}
    }
  };
  \node[
    recipe stage,
    minimum width=3.15cm,
    fill=RlFill,
    draw=RlAccent!55
  ] (rl) at ([xshift=\recipegap]reward.north east) {
    \recipebody{3.15cm}{6}{RL Alignment}{RlAccent}{
      \recipestep{RlAccent}{19}{SFT on expert edits}
      \recipestep{RlAccent}{20}{RL from SFT with verifiable rewards and RM}
      \recipestep{RlAccent}{21}{Evaluate the added value of the reward model over verifiable rewards alone}
    }
  };

  \draw[recipe flow] (rep.east) -- (data.west);
  \draw[recipe flow] (data.east) -- (pre.west);
  \draw[recipe flow] (pre.east) -- (human.west);
  \draw[recipe flow] (human.east) -- (reward.west);
  \draw[recipe flow] (reward.east) -- (rl.west);
\end{tikzpicture}
}
\caption{High-level recipe for training and post-training a domain-specific language model.}
\label{fig:recipe}
\end{figure}
\endgroup

%% file: content/dsl.tex

To enable scalable LLM training for architectural design, we develop a DSL and parser library that represent building geometry as compact structured text. Unlike Industry Foundation Classes (IFC), whose verbose structure produces long sequences, our DSL captures only the information needed for conceptual design in a semantically structured format optimized for token efficiency and deterministic parsing. The DSL encodes building massing, floor-plan spaces, and metadata as human-readable text suitable for LLMs. The parser provides full roundtrip support, allowing generated DSL to be parsed, validated, normalized, and converted back to geometry. Together, these components bridge free-form LLM outputs and structurally valid building representations for both training data preparation and post-generation validation. See \cref{app:dsl} for details.

\cref{fig:dsl_vis_render_app} and \cref{lst:dsl_example} in \cref{app:dsl} show example DSL representations and encodings. The DSL in \cref{fig:dsl_example_a} uses 642 characters versus over 10,000 for equivalent IFC, a 15$\times$ reduction, while preserving the geometric and semantic information needed for conceptual floor-plan generation.

%% file: content/data.tex

\subsection{Synthetic Data}
\label{sec:synthetic_data}
Data synthesis proceeds in four stages, each broadening the data distribution. First, we derive labeled floor plans from TileGPT \citep{gaier2024generative,villaggi2024tilegpt}, a quality-diversity system for high-performing tile-based layouts, and project their geometry into the DSL. Second, we apply scale augmentation to diversify massing and space dimensions beyond the tile grid. Third, we sample massings outside the tile-based distribution, generate layouts with a model checkpoint trained on the preceding synthetic data, and repair near-miss outputs, retaining only those that pass a correctness verifier. This enables generation on free-form massings beyond the tile-based distribution. The verifier follows the verifiable reward criteria from \cref{sec:verifiable_rewards}. Fourth, we apply rotation augmentation, sampling 20 angles per layout so that the model sees both near-identity perturbations and large reorientations; it runs last, after the processing steps of \cref{sec:data_processing}, so that no rotated variant of a training layout can reach evaluation. Both scale and rotation augmentations are structure-preserving. The synthetic corpus contains 4.1M samples before rotation augmentation and 82M after. \cref{app:data_generation} provides additional details.

\begin{figure}[b!]
    \centering
    \hspace{3em}
    \begin{subfigure}[b]{0.3\textwidth}
        \centering
        \includegraphics[width=\textwidth]{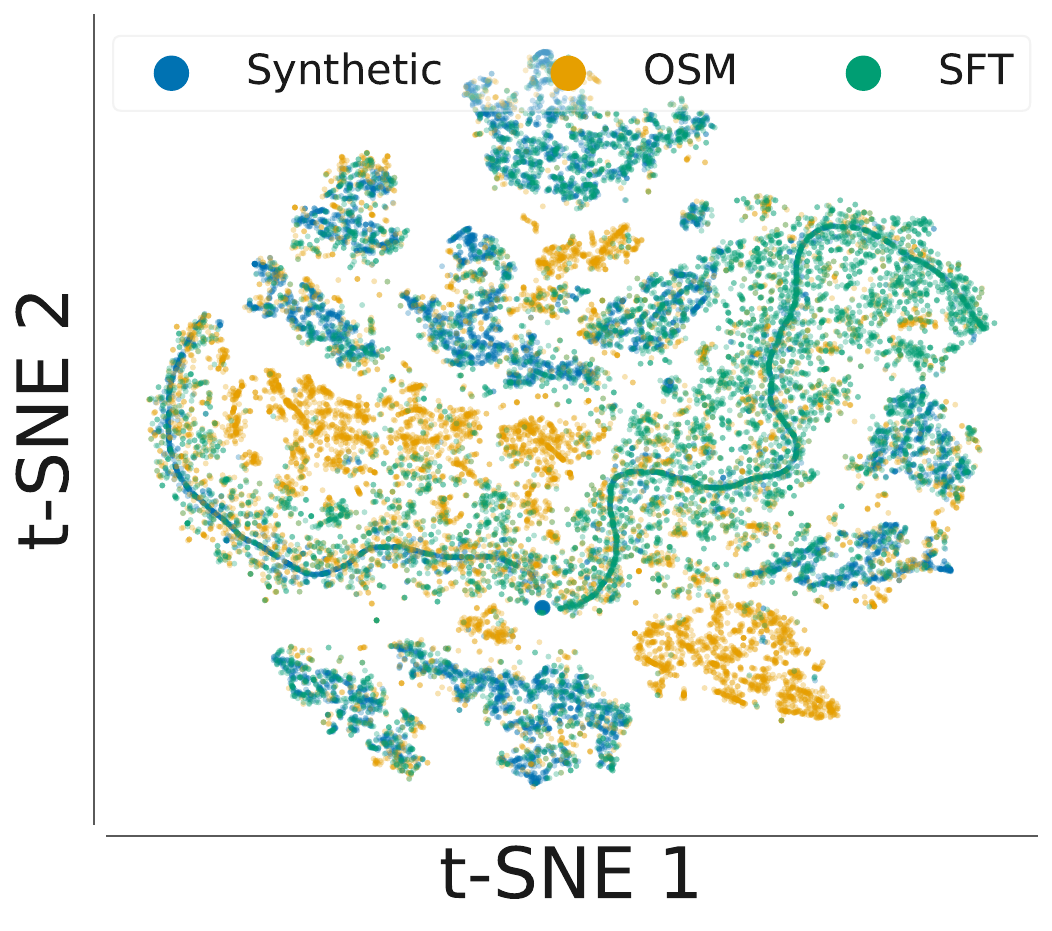}
        \caption{t-SNE of massing polygons.}
        \label{fig:tsne_data}
    \end{subfigure}
    \hfill
    \begin{subfigure}[b]{0.45\textwidth}
        \centering
        \includegraphics[width=\textwidth]{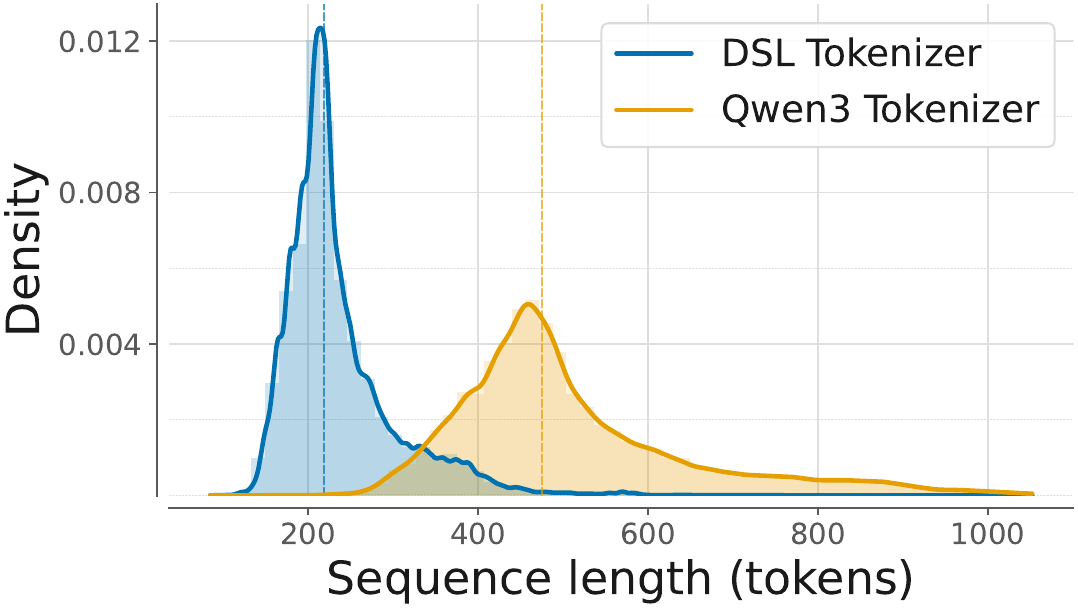}
        \caption{Sequence length distributions.}
        \label{fig:seq_lengths}
    \end{subfigure}
    \hspace{3em}
    \caption{\textbf{(a)} t-SNE visualization of massing polygons across synthetic, OSM, and SFT datasets, indicating the distribution shift between synthetic and real-world massing geometries. \textbf{(b)} Tokenized sequence length distributions for the pre-training dataset using the DSL tokenizer and the original Qwen3 tokenizer, showing that the DSL tokenizer reduces sequence lengths by 54\%, on average.} 
    \label{fig:data}
\end{figure}



\subsection{Real-World Data}
\label{sec:osm_data}

We collect 14k real-world multifamily building footprints from OpenStreetMap (OSM) \citep{OpenStreetMap} across more than 50 major North American cities. The dataset contains building massings and metadata converted to our DSL. Because OSM lacks ground-truth space layouts, it is used for evaluation rather than training. \cref{fig:tsne_data} visualizes the massings with t-SNE \citep{van2008visualizing}, showing substantial differences between the synthetic and OSM massing distributions, with many OSM samples lying outside the synthetic distribution. This motivates its use for out-of-distribution evaluation.
See \cref{app:data_diversity} for t-SNE details and \cref{app:data_samples} for data samples.

\subsection{Data Processing}
\label{sec:data_processing}

\paragraph{Data Quantization.}
Continuous data is quantized using a bucketing strategy. We limit prompts and completions to a physical extent of $102.4\mathrm{m} \times 102.4\mathrm{m}$ and divide this domain into 1024 buckets of $0.1\mathrm{m}$, effectively quantizing layouts onto a $10\mathrm{cm} \times 10\mathrm{cm}$ grid. This resolution is sufficient for the expected architectural applications. Continuous values are recovered by mapping each bucket to a fixed value within its range, with error scaling linearly with the grid spacing.

\paragraph{Data Normalization.}
A floor plan can have different DSL strings due to polygon start vertices, space ordering, and absolute position. We therefore emit each sample in canonical form. Using the parser's winding-order normalization (see \cref{app:dsl_overview}), we translate each sample so the minimum coordinate is the origin, start each polygon's vertex sequence at the vertex nearest the origin, and order spaces deterministically. The same transformation is applied at inference to match the training distribution. Canonicalization removes design-irrelevant variation and makes deduplication exact.

\paragraph{Deduplication and Splitting.}
Procedural generation can produce duplicates that inflate evaluation if shared across splits. We therefore hash each sample's canonical form and remove from validation and test any sample whose hash occurs in training. Rotation augmentation is applied after splitting, preventing training footprints from appearing in evaluation as exact or rotated duplicates.

%% file: content/pre-training.tex
DSL pre-training adapts general-purpose models to the DSL for syntactically valid generation.

\input{figures/tikz/prompt_completion}

\paragraph{Prompt Structure and Modality Randomization.}
Training samples use three DSL control tags: context, generate, and completion, as shown in \cref{fig:prompt_structure}. Context provides conditioning information such as building metadata, massing geometry, and optional partial layouts; generate specifies the target modality; and completion marks the autoregressive target. Models are trained with standard next-token prediction. During pre-training, modalities are randomly split between context and completion, with subsets of spaces provided as context and the remainder withheld for generation. This teaches both full and partial layout synthesis across varied modality combinations.

\paragraph{Custom Tokenizer and Vocabulary.}

We use a custom tokenizer specialized for the DSL, with a vocabulary of DSL keywords, delimiters, and quantized numeric tokens. As shown in \cref{fig:seq_lengths}, this substantially shortens sequences relative to general-purpose tokenizers, yielding a pre-training corpus of 17.6B tokens after augmentation. We ablate the impact of DSL-specific tokenization on training and downstream generation in \cref{sec:results_ablation} and \cref{app:ablation_tokenizer}.

\paragraph{Base Models.}
To test whether our recipe generalizes across model families and scales, we pre-train Qwen3 \citep{yang2025qwen3} (0.6B, 1.7B) and Pythia \citep{biderman2023pythia} (70M, 160M, 410M, 1.4B). Resizing the embedding layers to the smaller DSL vocabulary reduces effective parameter counts, while all remaining weights are initialized from the corresponding base models. The Chinchilla scaling rule of approximately 20 training tokens per parameter \citep{hoffmann2022training} suggests a compute-optimal scale near 0.9B parameters for our 17.6B-token corpus, though this estimate was developed for general-purpose pre-training. In our experiments, performance largely saturates above $\sim$0.4B parameters on synthetic data and improves only modestly at larger scales on OSM, motivating our focus on compact models up to $\sim$1.4B effective parameters. We use a fixed effective batch size and tune learning rates per model; see \cref{app:pretraining_hparams} for details.

%% file: figures/tikz/prompt_completion.tex
\begin{figure}[b!]
\centering
\resizebox{\linewidth}{!}{
\begin{tikzpicture}[
  font=\ttfamily\footnotesize,
  block/.style args={#1/#2}{
    draw=#1!82!black,
    fill=#1!7,
    rounded corners=1.15pt,
    line width=0.42pt,
    inner xsep=2.4pt,
    inner ysep=1.1pt,
    align=left,
    text width=#2
  }
]

\node[block=feedbackBlue/8.5cm, minimum height=1.3cm] (context) {
\tagc{feedbackBlue}{<context>}\\
\hspace{0.7em}\tagc{dslOrange}{<building>}...\tagc{dslOrange}{</building>}  \; \tagc{coreTeal}{<structure>}...\tagc{coreTeal}{</structure>}\\[-2.0pt]
\hspace{0.7em}\tagc{massingBlue}{<massing>}...\tagc{massingBlue}{</massing>}\quad
\tagc{dslGreen}{<space>}...\tagc{dslGreen}{</space>}\\[-2.0pt]
\tagc{feedbackBlue}{</context>}
};

\node[block=dslOrange/2.5cm, minimum height=1.3cm, right=4.2mm of context] (generate) {%
\tagc{dslOrange}{<generate>}\\[-2.0pt]
\hspace{0.7em}\tagc{dslGreen}{<space>}\\[-2.0pt]
\tagc{dslOrange}{</generate>}%
};

\node[block=postPurple/6.5cm, minimum height=1.3cm, right=4.2mm of generate] (completion) {
\tagc{postPurple}{<completion>}\\[-2.0pt]
\hspace{0.7em}\tagc{dslGreen}{<space>} spaces \{ polygon ... \} \tagc{dslGreen}{</space>}\\[-2.0pt]
\tagc{postPurple}{</completion>}
};

\path
  (context.east) to node[midway, font=\sffamily\footnotesize, text=softInk!65] {$+$}
  (generate.west);

\path
  (generate.east) edge[
    -{Latex[length=1.85mm,width=1.32mm]},
    line width=0.48pt,
    draw=postPurple!72!black
  ]
  (completion.west);

\end{tikzpicture}
}
\caption{Structured prompt and completion format used for DSL pre-training and post-training.}
\label{fig:prompt_structure}
\end{figure}
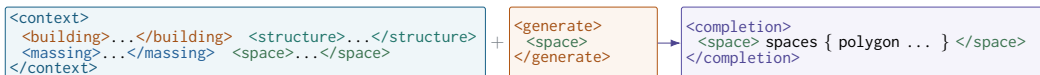

%% file: content/post-training.tex
The base model is limited by its reliance on synthetic data. Although DSL pre-training enables parseable outputs, the model inherits simplified design patterns and often produces architecturally weak layouts, with recurring issues in circulation, core placement, unit proportions, geometry, and labeling. We therefore combine human feedback alignment with verifiable rewards: practicing architects provide supervision that corrects synthetic-data biases and reflects professional judgment, while rule-based rewards enforce automatically verifiable geometric and functional constraints.

\subsection{Human Feedback Alignment}
\label{sec:feedback_collection}
We collect human feedback through an expert annotation workflow producing ratings, rankings, and corrected layouts (\cref{fig:rlhf_diagram}). For each massing, sampled from curated synthetic and manually created footprints, the model generates four candidate floor plans.

Architectural labelers rate each candidate on a 5-point Likert scale, rank them with ties permitted, and edit the highest-ranked layout. Edited layouts provide SFT targets, rankings yield pairwise preferences for reward modeling, and ratings filter noisy or inconsistent preferences. Feedback is collected iteratively across multiple rounds, with SFT and RL updates between rounds so that later collection uses progressively improved models. In total, we collect 10k SFT samples and 90k pairwise preference examples. Our setup follows standard human feedback and preference-based alignment practices \citep{ouyang2022training,rlhf2026lambert}; see \cref{app:human_feedback_collection} for additional details.

\subsection{Verifiable Rewards}
\label{sec:verifiable_rewards}
While human feedback captures holistic design quality and architectural judgment, verifiable rewards automatically enforce precise geometric and functional constraints. By construction, reward terms are bounded to $[0,1]$ and penalty terms to $[-1,0]$. \cref{app:verifiable_rewards} provides additional details.

\paragraph{Geometric Correctness.} This reward measures whether the generated floor plan satisfies fundamental spatial constraints. It comprises three checks: spaces must be contained within the building massing, spaces must not overlap one another, and the massing footprint must be fully occupied by spaces. Each check produces a continuous score, and the scores are combined via geometric mean. 

\paragraph{Functional Compliance.} This reward evaluates whether the generated floor plan satisfies the requirements of residential buildings. It checks that the numbers of cores and corridors fall within acceptable ranges, that a sufficient fraction of living units meet a minimum area threshold and are properly labeled. The individual check scores are aggregated using a geometric mean.


\paragraph{Soft Overlong Penalty.} This reward penalizes overlong completions to discourage unnecessarily verbose or degenerate generations \citep{yu2026dapo}. It is applied only as a penalty for completions exceeding the target length budget and does not provide positive reward for shorter completions. 

\subsection{Post-Training Methodology}
\label{sec:post_setup}

\paragraph{Supervised Fine-Tuning.}
The first post-training stage applies SFT on architect-edited layouts. Examples are formatted as prompt-completion pairs, with the conditioning context in the prompt and the target DSL sequence in the completion. The loss is computed only on completion tokens, and the SFT data is augmented using the same rotational transformations as pre-training.

\paragraph{Reward Model Training.}
The reward model (RM) is initialized from the SFT checkpoint by replacing the language modeling head with a scalar reward head. It is trained on architect-ranked pairwise preferences using the Bradley--Terry objective \citep{bradley1952rank,ouyang2022training}: $-\log \sigma \left(r_\theta(x,y_w) - r_\theta(x,y_l)\right)$, where $y_w$ and $y_l$ are the preferred and rejected completions for prompt $x$. The preference data is also augmented with rotational transformations.

\paragraph{Reinforcement Learning.}
The final stage performs RL using GRPO \citep{shao2024deepseekmath}, initialized from the SFT checkpoint, with prompts sampled from the synthetic and SFT data mixture. The reward function combines a frozen RM with verifiable rewards. Let $R_{\mathrm{RM}}$ denote the RM score and let $\{R_i\}_{i=1}^{N}$ denote the set of verifiable rewards described in \cref{sec:verifiable_rewards}. The overall reward is $R = \lambda_{\mathrm{RM}} \hat{R}_{\mathrm{RM}} + \sum_{i=1}^{N} \lambda_i R_i$, where $\lambda_{\mathrm{RM}}$ and $\lambda_i$ are weighting coefficients. Since the RM score is unconstrained while verifiable rewards are bounded by construction, we normalize the RM score as:
\begin{equation}
    \hat{R}_{\mathrm{RM}} = \tanh\left( \frac{R_{\mathrm{RM}}}{\alpha} \right),
    \label{eq:rm_normalization}
\end{equation}
where $\alpha$ is a fixed normalization scale. This bounds the RM contribution to $[-1,1]$ and prevents it from dominating the optimization objective. We note that the normalization also affects the effective preference scale observed by the policy during RL, which can alter the calibration of the RM. We therefore perform ablations over the normalization scale and report the results in \cref{sec:results_ablation}.

All post-training stages update the full set of model parameters rather than using parameter-efficient fine-tuning methods. Similarly to the pre-training setup, the effective batch size is held constant across model scales within each post-training stage, while learning rates are tuned separately for each model and stage. Additional training details and hyperparameters are provided in \cref{app:post_hparams}.

%% file: content/results.tex
\subsection{Evaluation Methodology}
\label{sec:evaluation_method}
We evaluate on 2k prompts from each of the synthetic and OSM test sets using the verifiable rewards in \cref{sec:verifiable_rewards}. We report pass@1/3/5, where a completion succeeds only if all checks pass. We also use a VLM judge to compare rendered layouts from the same input massing based on architectural quality, geometric validity, and functional usability. Each pair is evaluated in both candidate orders, with disagreements counted as ties. We use Gemini 3.5 Flash as the judge because it performs best among the frontier models in \cref{sec:frontier_comparison}. We additionally validate this judge against architect-implied preferences from the human-feedback dataset, finding 77.5\% agreement with human feedback across 10k pairs; see \cref{app:vlm_judge} for details. To the best of our knowledge, there is no publicly available specialized model that can be directly evaluated on the same massing-conditioned floor-plan layout generation task. We therefore compare FLOORA-0.6B (Qwen3) against frontier models using verifiable pass@k, VLM judge win rates on synthetic and OSM test prompts, and human evaluations. See \cref{app:frontier_model_inference} for the frontier-model inference protocol.

Because pre-training is costly, we train each base model with one seed. For post-training, we train each setting with 3 seeds. We evaluate all resulting checkpoints with 5 evaluation seeds, reporting aggregate performance with 95\% confidence intervals (CIs). Frontier baselines are API-only models, so we evaluate each with a single inference run under a fixed prompting and decoding protocol.

\begin{figure}[b!]
    \centering
    \hspace{1em}
    \begin{subfigure}[b]{0.45\textwidth}
        \centering
        \includegraphics[width=\textwidth]{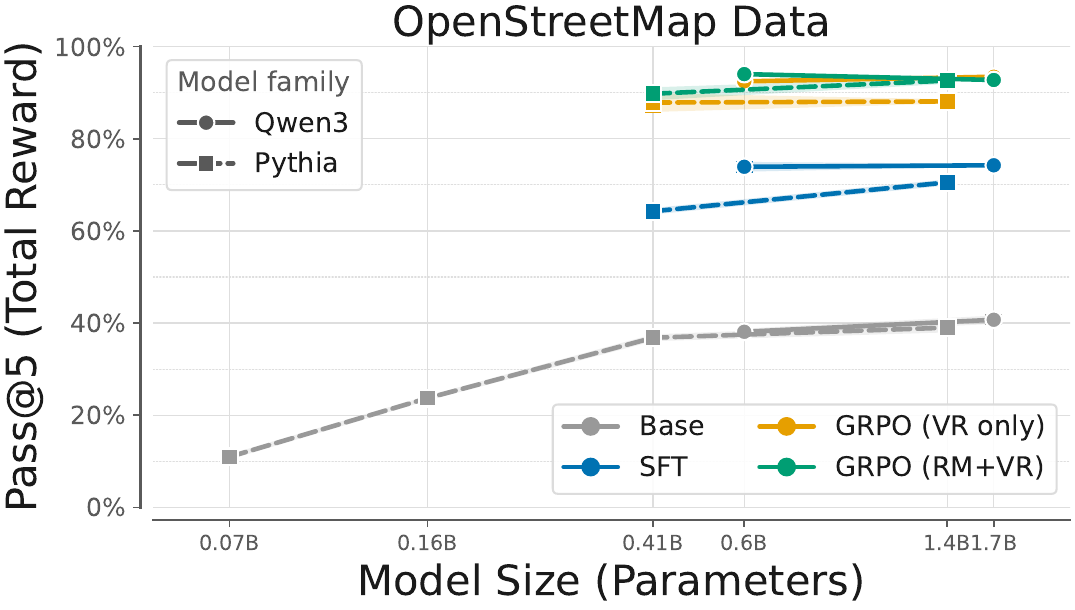}
        \caption{Pass@5 on the OSM test set.}
    \end{subfigure}
    \hfill
    \begin{subfigure}[b]{0.45\textwidth}
        \centering
        \includegraphics[width=\textwidth]{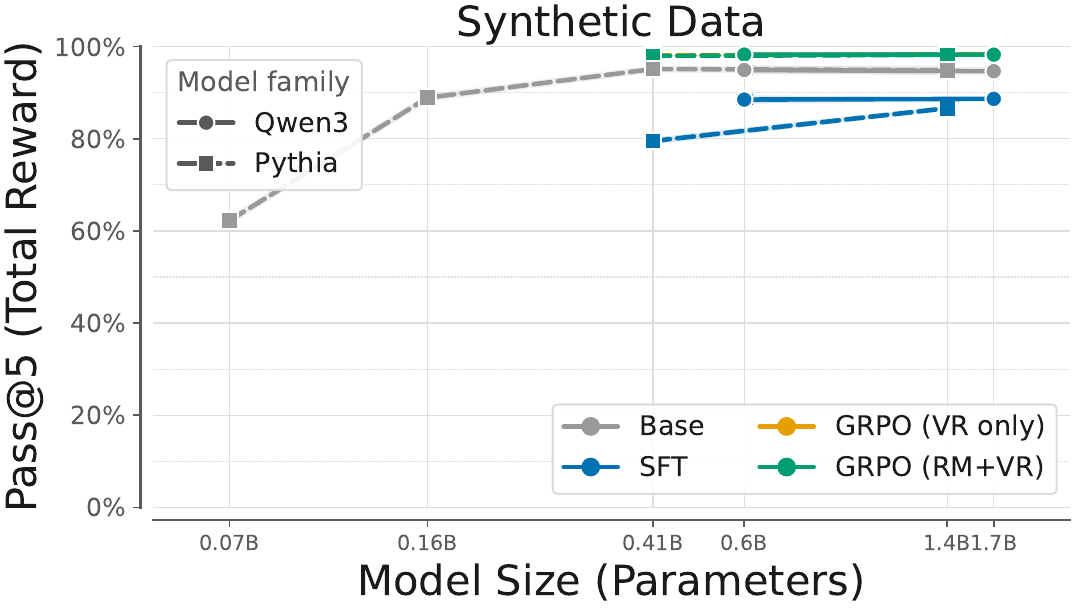}
        \caption{Pass@5 on the synthetic test set.}
    \end{subfigure}
    \hspace{1em}
    \caption{Pass@5 for FLOORA models across model sizes and training stages, where success requires all checks to pass. Shading shows 95\% CIs across 3 training and 5 evaluation seeds.}
    \label{fig:results_main}
\end{figure}

\subsection{Main Results}
\label{sec:results_main}

\paragraph{Pre-Training.}
Full results for the base models are presented in \cref{fig:results_pt_eval} and \cref{tab:pt_osm,tab:pt_synthetic} in \cref{app:results_pt}. Models with fewer than 0.4B parameters fail to effectively capture the underlying building patterns, leading to poor performance even on the synthetic dataset. Although larger models achieve near-saturated performance on the synthetic benchmark, their generalization capability continues to improve on OSM, which serves as an out-of-distribution evaluation setting for the base models. Given the inadequate performance of smaller models, such as Pythia-70M and Pythia-160M, we do not perform any post-training experiments on these models.

\begin{wrapfigure}{r}{0.38\textwidth}
\vspace{-10pt}
    \centering
    \includegraphics[width=0.38\textwidth]{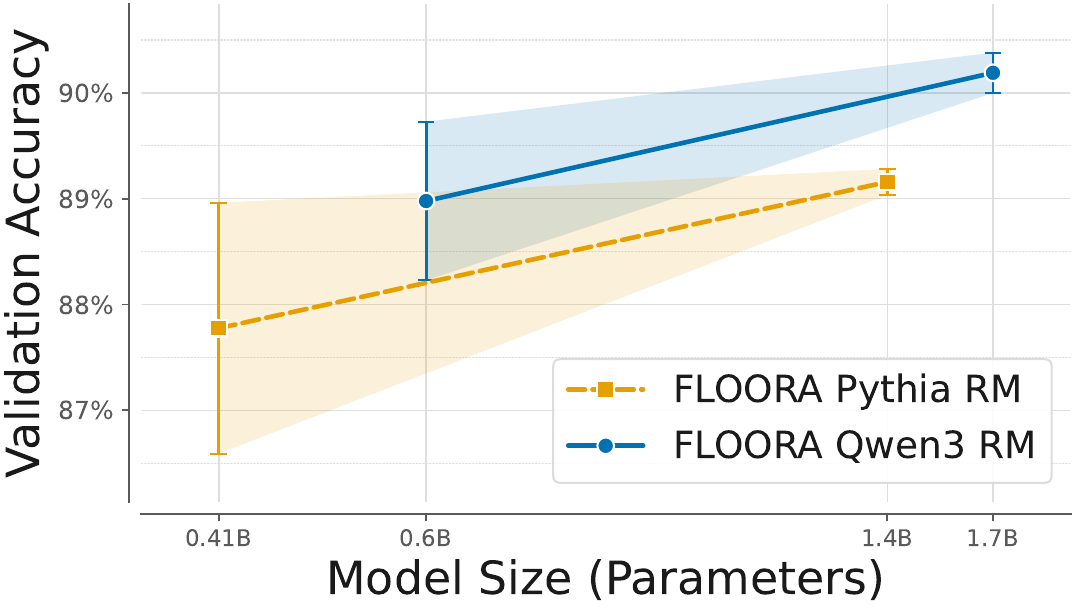}
    \caption{Reward model validation accuracy, reported with 95\% CIs.}
\label{fig:rm_accuracy}
\vspace{-15pt}
\end{wrapfigure}

\paragraph{Supervised Fine-Tuning.}
Full SFT results are presented in \cref{fig:sft_results} in \cref{app:results_sft}. SFT significantly improves performance on the OSM test set across model scales, showing that architect-edited layouts provide an effective alignment signal for real-world building footprints. Since OSM is out-of-distribution relative to the synthetic and SFT data (see \cref{fig:tsne_data}), these gains suggest improved generalization beyond the procedural pre-training distribution.

On the synthetic test set, we observe a decrease in performance after SFT, which we attribute to distribution shift between architect-corrected SFT targets and the procedural synthetic layouts used during pre-training (\cref{fig:tsne_data}). Overall, these results suggest that SFT functions primarily as a domain-alignment step, trading some in-distribution synthetic performance for better out-of-distribution generalization and closer alignment with expert architectural judgment.

\paragraph{Reward Model Training.}
We evaluate each RM on held-out pairwise preference data, reporting validation accuracy in \cref{fig:rm_accuracy} as the fraction of pairs where the architect-preferred layout receives a higher score. All RMs achieve high accuracy, capturing architect preference signals. Accuracy also increases with model size, suggesting that larger RMs better capture architectural preferences. Additional results are provided in \cref{app:results_rm}.

\begin{wrapfigure}{r}{0.5\textwidth}
\vspace{-10pt}
    \centering
    \includegraphics[width=0.5\textwidth]{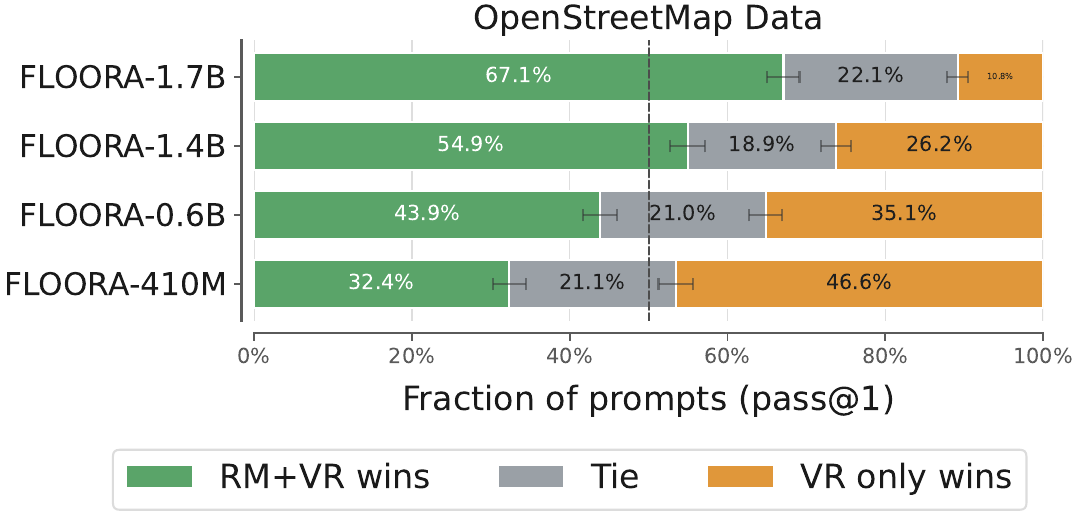}
    \caption{Pairwise VLM judge outcomes for each GRPO (RM+VR)  model against its matched GRPO (VR only) counterpart. Bars show the fraction of prompts where RM+VR wins, ties, or loses. As model size increases, the RM+VR variant is increasingly preferred. Error bars indicate 95\% Wilson intervals over the evaluation prompts.}
\label{fig:winrates_rlhf}
\vspace{-10pt}
\end{wrapfigure}

\paragraph{Reinforcement Learning.}
\cref{fig:results_main} reports Pass@5 for the base and post-trained models, with full pass@1/3/5 results in \cref{fig:results_main_app,tab:rl_osm,tab:rl_synthetic} in \cref{app:results_rl}. Both GRPO variants substantially improve OSM and synthetic performance, resolving many validity and constraint-satisfaction errors left by the base and SFT models. These gains are consistent across datasets and model scales. GRPO also recovers the SFT drop on synthetic data, likely caused by distribution shift, while preserving the OSM gains from SFT.

We further isolate the effect of the learned reward model by comparing GRPO (RM+VR) with matched GRPO (VR only) models using the VLM judge. As shown in \cref{fig:winrates_rlhf}, RM+VR is scale-dependent: it underperforms VR only for Pythia-410M, but becomes increasingly preferred with model capacity, reaching the largest margin for Qwen3-1.7B.  This suggests that the reward model can improve qualitative architectural preferences beyond VR-only training, but only when both the reward model and policy have enough capacity to learn and act on this preference signal.
Qualitative results in \cref{app:results_qualitative_progression} show the relative benefits of RM+VR in the coherence of space organization, unit proportions, core placement, and circulation quality.

\begin{figure}[t!]
    \hspace{2em}
    \centering
    \begin{subfigure}[b]{0.4\textwidth}
        \centering
        \includegraphics[width=\textwidth]{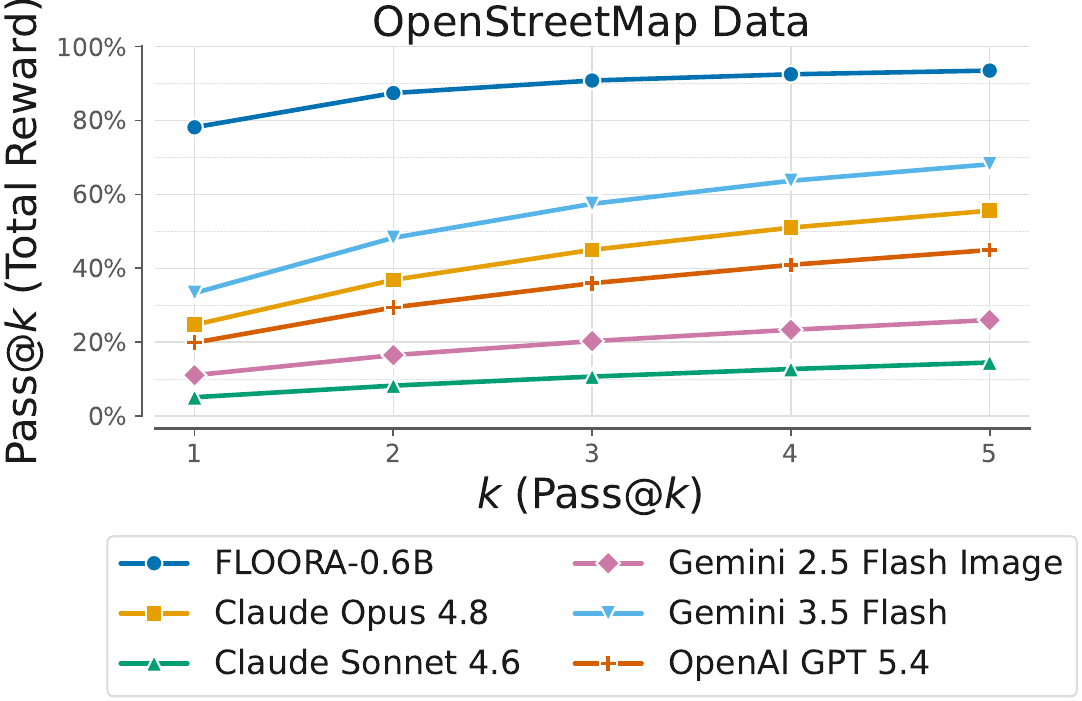}
        \caption{Pass@k on the OSM test set.}
        \label{fig:passk_frontier}
    \end{subfigure}
    \hfill
    \begin{subfigure}[b]{0.53\textwidth}
        \centering
        \includegraphics[width=\textwidth]{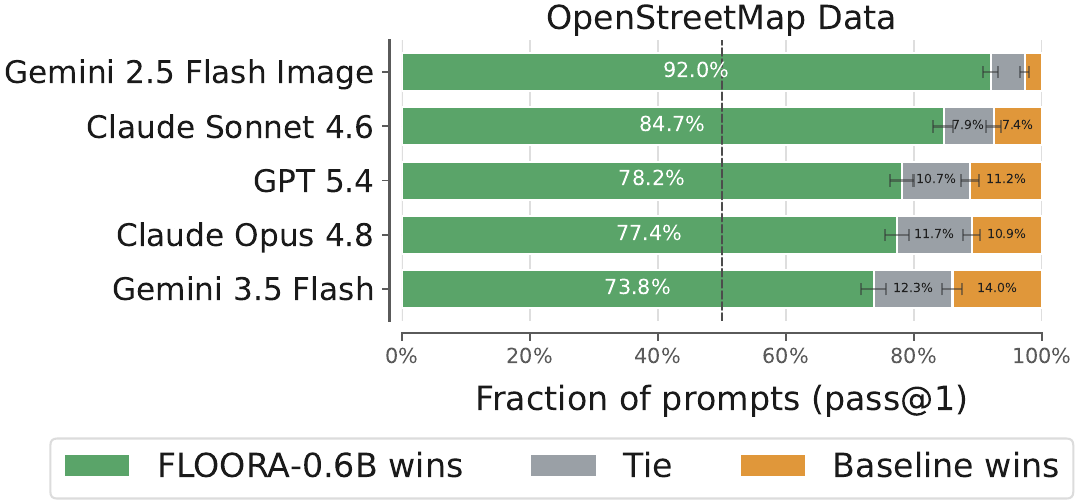}
        \caption{Win rates on the OSM test set.}
        \label{fig:winrates_frontier}
    \end{subfigure}
    \hspace{2em}
    \caption{\textbf{(a)} Pass@k comparison of FLOORA-0.6B (Qwen3) and frontier baselines, where success requires passing both geometric and functional checks. \textbf{(b)} Pairwise VLM judge outcomes for FLOORA-0.6B (Qwen3) against frontier baselines. Bars show the fraction of prompts for which FLOORA wins, ties, or loses. Our model substantially outperforms all evaluated frontier baselines.}
\end{figure}

\subsection{Comparison with Frontier Models}
\label{sec:frontier_comparison}
\begin{wrapfigure}{r}{0.5\textwidth}
\vspace{-10pt}
    \centering
    \includegraphics[width=0.5\textwidth]{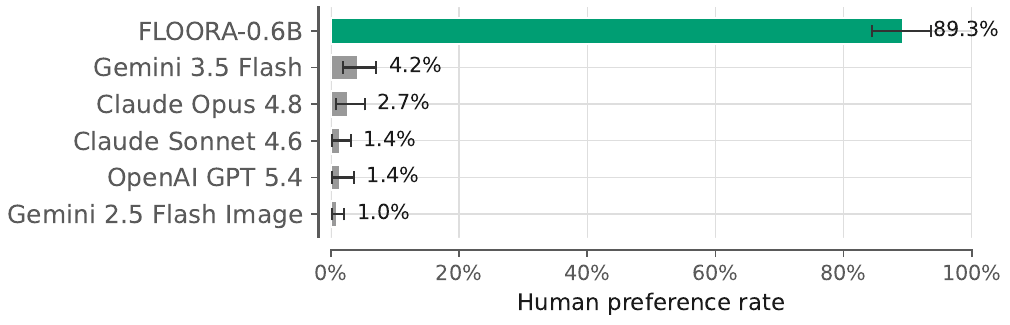}
    \caption{Human evaluation results across 100 test inputs with 95\% CIs. FLOORA is selected as the best model in 89.3\% of evaluations.}
\label{fig:human_eval}
\vspace{-10pt}
\end{wrapfigure}

We compare FLOORA-0.6B against several frontier models using the protocol in \cref{app:frontier_model_inference}. FLOORA substantially outperforms all baselines across pass@k (\cref{fig:passk_frontier}) and is preferred by the pairwise VLM judge at win rates of 73.8\textendash92.0\% (\cref{fig:winrates_frontier}). We additionally conduct human evaluations on 100 test inputs, comprising 50 OSM and 50 synthetic samples, with 10 labelers. FLOORA is selected as the best model in 89.3\% of evaluations (\cref{fig:human_eval}). Additional results and details are provided in \cref{app:results_frontier}. Thus, FLOORA's advantage extends beyond verifier-based pass@k metrics to visual and architectural quality. Qualitative comparisons in \cref{fig:qualitative_comparison_main} and \cref{app:results_qualitative_frontier} show more coherent circulation, regular living-unit subdivisions, and plausible unit proportions. Some frontier outputs still pass geometric and functional checks while exhibiting practical issues such as weak corridor connectivity or irregular unit geometry.

\subsection{Ablation Studies}
\label{sec:results_ablation}
We ablate key design choices and feedback data size using FLOORA-0.6B as the reference.


\paragraph{RM Normalization.} \cref{app:ablation_reward_model} ablates the RM normalization scale $\alpha$ in \cref{eq:rm_normalization}. Without normalization, the RM score dominates the bounded verifiable rewards. Performance is robust for $\alpha \in [1,4]$, while $\alpha=10$ degrades verifiable rewards. We select $\alpha=3$ as a balanced setting that preserves high verifiable rewards while improving the RM signal.

\paragraph{Original vs. DSL Tokenizers.} \cref{app:ablation_tokenizer} ablates tokenizer choice. The DSL tokenizer reduces mean sequence length by 54\% (\cref{fig:seq_lengths}), shrinking the augmented pre-training corpus from 38.5B to 17.6B tokens, wall-clock pre-training time by 31\%, and GPU-hours by 66\%. It also improves downstream OSM pass@5 by 6.5–7.2 percentage points for the GRPO models.

\paragraph{Human Feedback Scale.} \cref{app:ablation_feedback_scale} studies performance as human feedback increases during iterative collection. Because model updates between rounds couple data quantity with collection-model quality, we use cumulative chronological subsets reflecting the feedback available at each stage. Performance improves, with larger early gains and diminishing returns at larger scales. Although this does not isolate data quantity alone, it reflects the practical collection setting and suggests that performance can indicate when additional annotation offers limited marginal benefit.

%% file: content/limitations.tex
Our models are designed for structured architectural generation rather than broad natural language understanding, and should not be interpreted as general-purpose design assistants. We also intentionally do not study agentic workflows, since such systems typically rely on repeated inference from large frontier models and therefore retain the high cost and latency that our compact domain-specific approach seeks to reduce. 
Finally, our data is deliberately scoped to multifamily residential floor-plan generation conditioned on a building massing. While our model generalizes well across diverse real-world building footprints from OpenStreetMap (pass@5 of 94\%), performance may be lower on highly irregular massings that are underrepresented in the training and post-training data.


%% file: content/conclusion.tex
We presented a compact domain-specific modeling recipe for architectural layout generation, combining a token-efficient DSL, domain-specific pre-training, and architect-guided post-training with learned and verifiable rewards. Our results show that a small DSL model can outperform significantly larger frontier models on the evaluated architectural layout generation benchmarks, across automatic metrics, VLM judge comparisons, and human evaluations. Ablations highlight the importance of representation design, reward model normalization, and human feedback, and show that reward model based post-training benefits from larger model capacity. More broadly, our results suggest that similar domain-specific representations, constraints, and expert-guided alignment strategies may be useful for other structured engineering generation tasks.

%% file: content/statements.tex
\subsection*{AI use statement}
In this work, we used generative AI tools to assist in the implementation of methods and experiments. For synthetic data generation, we used an intermediate checkpoint of our own trained model to generate layouts. We also used a VLM as a judge for portions of the experimental evaluation and used inference APIs of generative AI models as experimental baselines. Additionally, we used generative AI tools to create or edit software code, draft and edit parts of the research paper, improve readability, grammar, and word choice, rewrite and paraphrase text, and assist with the related work section, including summarizing and analyzing existing literature.

We did not use generative AI tools to generate research ideas, propose or refine the central research hypotheses of this work, interpret experimental results, or draw conclusions from the results. The interpretation of the results and the conclusions presented in the paper were developed by the authors.

All AI-assisted code was reviewed and verified by the authors, and all AI-assisted text included in the paper was reviewed and edited by the authors. We take responsibility for the final content of this work, including text, claims or artifacts produced with the aid of generative AI.

\subsection*{Ethics Statement}
The human-feedback component of this work involved professional architects who were engaged and compensated through a BIM consultancy contractor to evaluate, rank, and edit model-generated architectural layouts. The real-world evaluation data are derived from publicly available OpenStreetMap building footprints and do not contain private architectural plans or personally identifiable information. FLOORA is narrowly scoped to structured, conceptual multifamily residential floor-plan generation. We therefore expect the potential for misuse, as well as privacy and security concerns, to be limited compared with general-purpose or agentic systems.

\subsection*{Reproducibility Statement}
To support reproducibility, we provide detailed descriptions of the DSL and parser in \cref{app:dsl}, synthetic data generation and processing in \cref{app:data}, the human-feedback collection and processing pipeline in \cref{app:human_feedback_collection}, training settings and hyperparameters in \cref{app:training_hparams}, verifiable reward definitions in \cref{app:verifiable_rewards}, the VLM-judge evaluation protocol in \cref{app:vlm_judge}, and the frontier-model inference protocol in \cref{app:frontier_model_inference}. Additional quantitative results and ablation studies are provided in \cref{app:results,app:ablations}.

We release the synthetic and OSM datasets, post-trained model checkpoints, and inference code as supplementary materials to further support future research. The human-feedback data used for supervised fine-tuning and reward model training and the training code are not included in the release. The paper and appendices document the corresponding data collection, processing, training procedures, hyperparameters, reward formulations, and evaluation protocols to facilitate reproduction of the reported experiments.

%% file: appendix/dsl.tex
\subsection{Overview}
\label[appendix]{app:dsl_overview}
To represent building designs as structured text suitable for LLM training and generation, we developed a custom DSL and an accompanying parser library. The DSL encodes a building design as a collection of independent modalities, each covering a distinct architectural or structural aspect. Modalities are compact, keyword-delimited text blocks using integer coordinates (in quantized format) and enumerated vocabulary, designed to be token-efficient, human-readable, and directly consumable by LLMs as both input context and generation target. The library provides full roundtrip support; raw DSL text produced by a model can be parsed, structurally validated, semantically normalized, and re-emitted in canonical form, forming the backbone of both training data preparation and post-generation validation. The library is organized into three layers:

\paragraph{Grammar.} Each modality is specified by a formal Parsing Expression Grammar (PEG) grammar written using TextX \citep{dejanovic2017textx}, a Python framework that automatically derives parsers and in-memory object models from grammar definitions. Grammars share coordinate and size primitives reused across modalities.

\paragraph{Parser.} Each grammar has a corresponding parser class that loads the grammar at runtime, invokes TextX to parse input text, and post-processes the result into a typed domain model. Post-processing enforces polygon winding-order normalization and resolves symbolic cross-references (e.g., edge names referencing vertices by ID).

\paragraph{Model.} Each modality has a typed Python dataclass that holds validated geometric and semantic data, performs structural integrity checks, supports JSON serialization with backward-compatible field aliasing, and can re-emit canonical DSL text.

\subsection{Modality Reference}
This section describes each modality, its purpose, and a minimal example.

\paragraph{Building and Structure.} The \emph{building} is the global metadata about the building: occupancy type, number of storeys, and a reference floor level with its elevation. This modality provides the high-level design intent and serves as the context header for a generation sequence. The \emph{structure} specifies the primary structural material system. Together with building, this establishes the engineering premise of the design before any geometry is introduced. 

\paragraph{Massing.} The \emph{massing} describes the 3D volumetric envelope as a polygonal footprint sharing a single floor-to-floor height.

\paragraph{Space.} The \emph{space} represents the floor-plan geometry as a flat list of labeled polygons. Each polygon carries a semantic space type and a sequence of counter-clockwise 2D vertices. This is the simplest geometric representation, pure boundary coordinates with no explicit topology, making it the preferred format for floor-plan generation tasks. The supported space types are core, corridor, and living unit.

\paragraph{Space Graph.} The \emph{space graph} is the topological counterpart to spaces. In this representation, a floor plan is modeled as a graph: named vertices store 2D coordinates, named edges connect pairs of vertices through symbolic references, and spaces are defined as ordered traversals of edges. TextX cross-reference resolution enforces referential integrity by ensuring that every edge reference in a space boundary points to a declared edge and that every edge endpoint refers to a declared vertex.

Shared walls between adjacent rooms are represented as a single directed edge rather than duplicated overlapping boundaries, making the spacegraph the canonical representation for downstream structural and topological analysis. Since spaces are a simplified abstraction of the spacegraph, spaces are used to train the LLM, while spacegraphs are used for parsing and validation.

\subsection{Example DSL}
\begin{figure}[h!]
    \centering
    \hspace{2em}
    \begin{subfigure}[b]{0.4\textwidth}
        \centering
        \includegraphics[width=\textwidth]{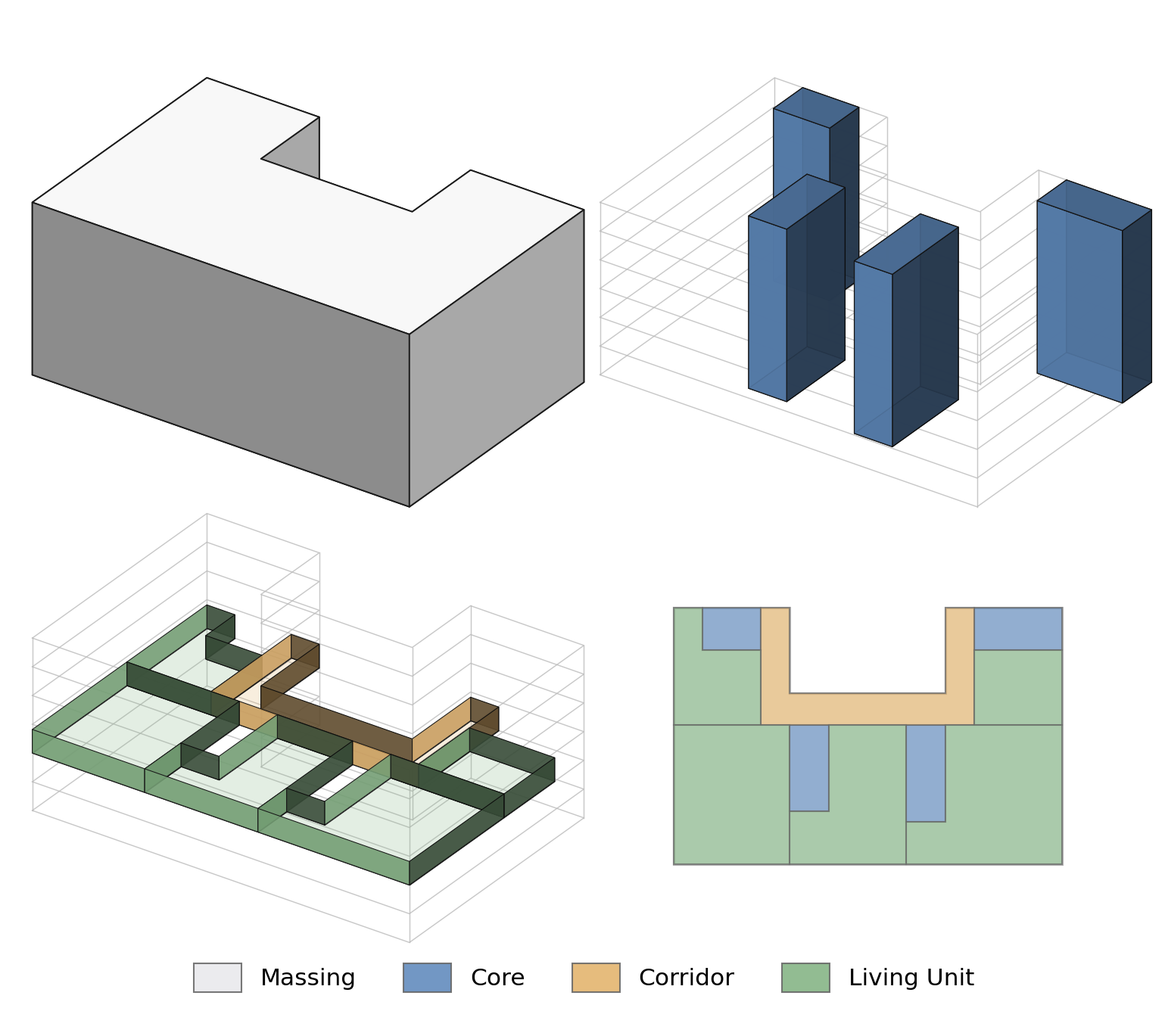}
        \caption{}
        \label{fig:dsl_example_a}
    \end{subfigure}
    \hfill
    \begin{subfigure}[b]{0.4\textwidth}
        \centering
        \includegraphics[width=\textwidth]{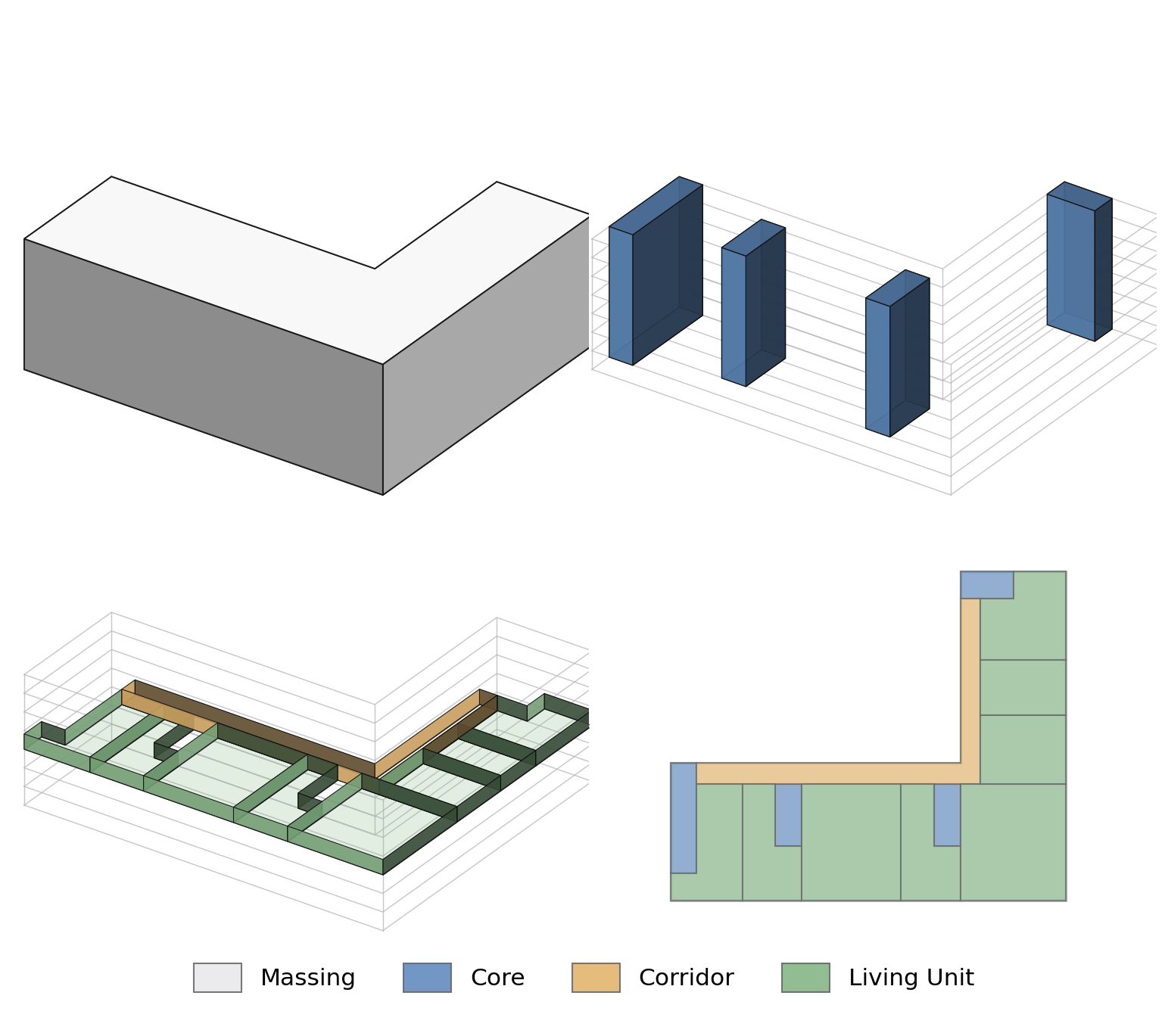}
        \caption{}
        \label{fig:dsl_example_b}
    \end{subfigure}
    \hspace{2em}
    \caption{Examples of DSL representation of multifamily residential buildings, illustrating the massing, vertical cores, living units, corridors, and corresponding 2D floor-plan layout.}
    \label{fig:dsl_vis_render_app}
\end{figure}

\cref{fig:dsl_vis_render_app,lst:dsl_example} show examples of the DSL representation used for building layout generation. The DSL encodes each design using structured text blocks for building metadata, structural material, massing geometry, and labeled floor-plan spaces. In \cref{lst:dsl_example}, the \texttt{massing} block defines the exterior footprint and height, while the \texttt{spaces} block specifies the cores, corridor, and living units as labeled polygons. \cref{fig:dsl_vis_render_app} visualizes the same example geometrically, illustrating how the compact textual encoding maps directly to the massing and 2D floor-plan layout.

\begin{lstlisting}[
    language=DSL,
    caption={Examples of DSL encodings of multifamily residential buildings, including building metadata, structural material, massing geometry, and floor-plan spaces.},
    label={lst:dsl_example}
]
# DSL for Example (a)
building { occupancy_type multifamily_residential storeys 6 level 2 elevation 60 } 
structure { material wood_frame }
massing {
    height 30 polygon 0,0 416,0 416,275 291,275 291,183 124,183 124,275 0,275 
} 
spaces {
    polygon core 322,229 416,229 416,275 322,275 
    polygon core 249,45 291,45 291,149 249,149 
    polygon core 124,57 166,57 166,149 124,149 
    polygon core 31,229 93,229 93,275 31,275 
    polygon corridor 93,149 322,149 322,275 291,275 291,183 124,183 124,275 93,275 
    polygon living_unit 322,149 416,149 416,229 322,229 
    polygon living_unit 249,0 416,0 416,149 291,149 291,45 249,45 
    polygon living_unit 124,0 249,0 249,149 166,149 166,57 124,57 
    polygon living_unit 0,149 93,149 93,229 31,229 31,275 0,275 
    polygon living_unit 0,0 124,0 124,149 0,149 
}

# DSL for Example (b)
building { occupancy_type multifamily_residential storeys 7 level 3 elevation 90 } 
structure { material reinforced_concrete }
massing {
    height 30 polygon 0,0 610,0 610,508 448,508 448,212 0,212 
} 
spaces { 
    polygon core 0,42 40,42 40,212 0,212 
    polygon core 162,84 203,84 203,180 162,180 
    polygon core 407,84 448,84 448,180 407,180 
    polygon core 448,466 529,466 529,508 448,508 
    polygon corridor 40,180 478,180 478,466 448,466 448,212 40,212 
    polygon living_unit 0,0 112,0 112,180 40,180 40,42 0,42 
    polygon living_unit 112,0 203,0 203,84 162,84 162,180 112,180 
    polygon living_unit 203,0 356,0 356,180 203,180 
    polygon living_unit 356,0 448,0 448,84 407,84 407,180 356,180 
    polygon living_unit 448,0 610,0 610,180 448,180 
    polygon living_unit 478,180 610,180 610,286 478,286 
    polygon living_unit 478,286 610,286 610,371 478,371 
    polygon living_unit 478,371 610,371 610,508 529,508 529,466 478,466 
}
\end{lstlisting}

\subsection{Industry Foundation Classes (IFC)}


Industry Foundation Classes (IFC) is an open industry standard file format for encoding and exchanging AEC BIM data ~\citep{buildingSMART_IFC}. The IFC specification defines a data model for encoding a complete BIM model, including: high-level object classes (like IfcWall, IfcSlab, and IfcSpace), non-geometric material and property values, and low-level geometry, like vectors, points, lines, polygons, meshes, and BRep solids.

IFC files are ASCII-based and utilize line number references to link components of an element definition, meaning the definition of an XYZ point referenced by the geometry defining a BIM element may involve tens or hundreds of intermediate references and span thousands of lines. IFC file size is also highly dependent on the modality of element geometry representation, where the mesh-modeled elements can add hundreds or thousands of lines just to define the vertex and face geometry compared to a geometrically equivalent but compact BRep representation. The verbosity and highly variable structure of IFC presents challenges to direct encoding with transformer architecture, and the multi-layer referential structure means a generated file might be syntactically correct but unparsable by BIM editing software.

%% file: appendix/data.tex
\subsection{Data Generation}
\label[appendix]{app:data_generation}

The synthetic corpus is built in four stages. Each stage targets a distributional limitation of the previous one, and every stage emits DSL in the canonical form, described in \cref{app:data_processing}. Stages 1 to 3 produce the layouts; stage 4 is applied last after deduplication and split generation.

\subsubsection{Stage 1: Layouts Derived from TileGPT}
\label[appendix]{app:stage1_tilegpt}
TileGPT \citep{gaier2024generative,villaggi2024tilegpt} is a generative design system for tile-based architectural layouts. It uses MAP-Elites, a quality-diversity search algorithm, to produce a collection of high-performing layouts spanning user-defined design attributes, fine-tunes a language model on that collection, and refines generated conceptual layouts into constraint-compliant detailed layouts using Wave Function Collapse, a constraint-satisfaction procedure. The diversity and validity of the stage-1 layouts therefore derive from prior work.

Our contribution at this stage is the projection into the DSL. Each TileGPT layout is a three-dimensional array of tile identifiers with transformation flags, which we resolve against a tile catalog into a node-and-feature geometric representation: a deduplicated node set with tolerance-based corner merging, and labeled features referencing those nodes. We then consolidate collinear walls, resolve the labeled spaces into closed polygons, and emit the building, structure, massing, and spaces modalities. Because tiles meet on shared edges, this step must merge coincident vertices and eliminate degenerate segments; without it, polygons that are visually correct produce self-intersecting or zero-area geometry in the DSL.

\subsubsection{Stage 2: Scale Augmentation}
\label[appendix]{app:stage2_scale}
Stage-1 layouts inherit the tile module dimensions, so space sizes are drawn from a small discrete set. We rescale $94\%$ of stage-1 layouts, drawing an independent scale factor for each of the two horizontal axes uniformly from $[0.604, 1.396]$, i.e.\ $1 \pm 0.396$. Because the two factors are drawn independently, the operation both changes overall size and perturbs aspect ratio, producing layouts whose massing and space dimensions vary continuously rather than over the discrete tile set. The bound is what keeps the result usable: beyond it, scaling yields living units and corridors whose proportions no longer correspond to buildable space.

\subsubsection{Stage 3: Inference and Repair on Free-Form Massings}
\label[appendix]{app:stage3_repair}

Stages 1 and 2 cannot produce massings outside the tile-based distribution, yet interactive use requires competence on free-form outlines. Stage 3 closes this gap by generating massings directly, asking an intermediate model to fill them, and repairing the outputs that are close to correct.

\paragraph{Massing Sampling.}
We sample rectangles, squares, and L-shapes, each in orthogonal and non-orthogonal variants. Rather than drawing shape parameters independently, we enumerate a full-factorial grid over the parameters of each family (width and height, and for L-shapes the notch proportion in each axis and the notch corner) and select a subset of grid points for generation, which gives systematic coverage of the parameter space rather than the clustering that independent uniform sampling produces. L-shape notch proportions are restricted to $[0.4,0.65]$ of each axis, which keeps both wings thick enough to contain habitable space. Non-orthogonal variants are produced by perturbing each vertex independently with probability $0.7$ by an offset of up to 20--25 grid units, with any perturbation rejected if it would duplicate an existing vertex. Rectangles and squares span 100--550 grid units ($10$--$55\mathrm{m}$) per dimension and L-shapes 150--550 grid units ($15$--$55\mathrm{m}$).

\paragraph{Candidate Generation.}
For each sampled massing we query an intermediate checkpoint of our own model for a spaces completion.

\paragraph{Repair Operator.}
Intermediate-model completions on unfamiliar massings are typically close to valid but leave small uncovered slivers, overshoot the outline, or omit a space. The repair operator converts the completion to a layout graph, adjusts it to fit the massing, and converts it back to DSL. It proceeds in a fixed order: remove spaces lying entirely outside the outline; clip spaces crossing it; divide large interior unused areas proportionally among adjacent spaces; insert new living units where the remaining free area admits them; iteratively extend spaces toward the outline; and finally merge any leftover small unused areas into adjacent spaces. Both iterative phases stop when the layout passes the admission checks below, when an iteration fails to improve coverage, or at a bounded iteration count. Operating on a graph rather than on independent polygons means shared boundaries between adjacent spaces move together; in addition, each merge or extension is tested against neighbouring spaces beforehand and skipped if it would produce a significant overlap.

\paragraph{Admission Verifier.}
A repaired layout enters the corpus only if it satisfies all of the following, evaluated on the repaired geometry with $\epsilon = 10^{-3}$:

\begin{enumerate}
    \item \emph{Coverage.} The intersection-over-union of the union of spaces with the massing is within $\epsilon$ of 1, and the ratio of total space area to massing area does not exceed $1 + \epsilon$.
    \item \emph{Composition.} At least one core, at least one corridor, and at least one living unit are present.
    \item \emph{Shape quality.} No living unit or core contains a narrow neck, detected by morphological erosion after simplification. This rejects spaces that are nominally large enough but not usable.
    \item \emph{Fidelity to the model output.} Relative to the pre-repair completion, no corridor or core loses more than $20\%$ of its area and no living unit loses more than $50\%$. This prevents the repair operator from silently replacing the model's design with its own.
\end{enumerate}

Layouts failing any check are discarded rather than corrected further.

This verifier is a formal specification of geometric and functional correctness for multifamily layouts, and it is the same specification that the verifiable rewards in \cref{sec:verifiable_rewards} implement as a reward signal. We regard this as a design decision rather than an incidental overlap: a single definition of correctness is applied consistently when admitting training data and when scoring policy rollouts. It does mean that verifier-based metrics on synthetic data are not independent evidence. Every sample in the corpus passes the coverage filter of \cref{app:deduplication}, and stage-3 samples additionally pass the composition and shape checks above, so in-distribution base performance is correspondingly high: our 0.6B base model reaches 90.2\% pass@1 on the synthetic test set, which measures how well it has internalized a specification it was trained toward rather than performance against an independently chosen metric. The informative measurement is on the real-world footprints of \cref{sec:osm_data}, which are used neither to generate nor to filter training data. There the same model reaches 23.8\% pass@1 (\cref{app:results_pt}), and the post-training gains reported in \cref{sec:results} are measured from that baseline.

\subsubsection{Stage 4: Rotation Augmentation}
\label{app:stage4_rotation}

Rotation augmentation is applied last, after the filtering, deduplication, and splitting described in \cref{app:data_processing}, so that no rotated variant of a training layout can reach the validation or test split. Writing the angle distribution of rotation augmentation as a mixture,
$$
\{\theta_i\}_{i=1}^{20}
\sim
4\,\operatorname{Uniform}(0^\circ,10^\circ)
+12\,\operatorname{Uniform}(10^\circ,350^\circ)
+4\,\operatorname{Uniform}(350^\circ,360^\circ),
$$
the two narrow components supply near-identity perturbations, which teach insensitivity to small orientation changes, while the wide component supplies large reorientations. Rotated coordinates are re-quantized, and a rotated variant is discarded if any coordinate falls outside the representable range, so a layout near the extent limit contributes fewer than 20 variants. The same procedure is used for the human-feedback data (\cref{app:feedback_processing}).

\subsubsection{Dataset Composition}
\label[appendix]{app:dataset_composition}

\begin{table}[h]
\centering
\caption{Synthetic corpus composition by generation stage. Counts are before rotation augmentation unless noted.}
\label{tab:synthetic_composition}
\begin{tabular}{llrrr}
\toprule
Stage & Description & Samples & Share of corpus \\
\midrule
1,2 & TileGPT-derived layouts with scale jitter        & 3.9M   & 95\% \\
3 & Inference + repair (admitted)  & 0.2M   & 5\% \\
\midrule
\multicolumn{2}{l}{Total before rotation augmentation} & 4.1M & 100\% \\
\multicolumn{2}{l}{Total after rotation augmentation}  & 82M  & \\
\bottomrule
\end{tabular}
\end{table}

\subsection{Data Processing}
\label[appendix]{app:data_processing}

\subsubsection{Canonicalization}
\label[appendix]{app:canonicalization}

A floor plan has many DSL encodings that carry the same design information: the vertex sequence may start anywhere in the cycle, spaces may be listed in any order, and the sample may sit anywhere in the coordinate range, which is immaterial because the DSL does not model site position. Winding order is already normalized by the parser (\cref{app:dsl_overview}), which emits outer boundaries counter-clockwise and hole boundaries clockwise so that orientation alone distinguishes a boundary from an interior void. The remaining freedom is removed as follows.

\paragraph{Start Vertex.}
The vertex sequence is rotated to begin at the vertex of least Euclidean distance from the origin. Line-like elements are ordered so that the endpoint nearer the origin comes first, with ties broken by smaller $x$ and then smaller $y$.

\paragraph{Translation.}
The sample is translated so that the minimum $x$ and minimum $y$ taken jointly over all planar modalities become zero. The shift is a single integer vector applied uniformly, so relative geometry is unchanged and the sample occupies the low corner of the quantized range.

\paragraph{Space Ordering.}
Space polygons are sorted by label and then by their vertex string, giving a deterministic order independent of the order in which the generator emitted them.

Canonicalization runs on training data and on model inputs at inference time. Two properties follow: textually identical samples are geometrically identical, which makes hash-based deduplication exact; and the model is never asked to spend capacity distinguishing encodings that denote the same building.

\subsubsection{Deduplication and Split Construction}
\label[appendix]{app:deduplication}

\paragraph{Quality Filtering.}
Samples are first scored and filtered on two criteria: any coordinate outside the representable range, and a coverage score combining the intersection-over-union of the space union with the massing and the ratio of total space area to massing area. Samples scoring at or below the threshold are discarded.

\paragraph{Canonical Hash.}
For each surviving sample we form the string consisting of its canonicalized spaces text and its massing text, and hash it with BLAKE2b truncated to 128 bits. Collision probability is negligible at corpus scale. The hash covers geometry only: the building and structure modalities are excluded, so two layouts with identical geometry but different structural material are treated as duplicates and one is retained. This is intentional, since geometry is the property we deduplicate on, but it does mean the corpus contains no pairs that differ solely in structural material.

\paragraph{Within-Split Deduplication.}
Samples are streamed in chunks and the first occurrence of each hash is retained.

\paragraph{Cross-Split Overlap Removal.}
Deduplication within a split does not prevent the same layout appearing in two splits. We therefore treat the training split as the reference and remove from validation and test any sample whose hash occurs in training.

\paragraph{Ordering.}
Rotation augmentation is applied after splitting, so a footprint present in training cannot reach evaluation as a rotated variant, and cross-split overlap removal ensures it cannot reach evaluation as an exact duplicate either.

\subsection{t-SNE Visualization of Massing Geometry}
\label[appendix]{app:data_diversity}

To compare the geometric distributions of the synthetic, OSM, and SFT (described in \cref{sec:feedback_collection,app:human_feedback_collection}) datasets, we project their massing footprints into a shared 2D embedding using t-distributed stochastic neighbor embedding (t-SNE) \citep{van2008visualizing}. The analysis assesses whether the datasets occupy overlapping or distinct regions of footprint shape space. The procedure consists of two stages: shape descriptor construction and dimensionality reduction. Notably, this visualization is computed on the datasets before rotation augmentation is applied.

\paragraph{Shape Descriptor Construction.}
Raw polygon coordinates are not directly comparable across samples because footprints vary in absolute location, scale, and vertex count. We therefore convert each footprint into a normalized fixed-length contour descriptor. First, we translate the polygon to center the footprint at the origin. Second, we normalize scale by dividing all coordinates by $\sqrt{A}$, where $A$ is the polygon area after centering, so that the normalized polygon has unit area. Third, we resample the exterior boundary at uniform arc-length intervals. Given a normalized polygon with perimeter length $L$, we sample $N$ boundary points at distances:
\begin{equation}
    d_i = \frac{iL}{N}, \qquad i = 0, 1, \ldots, N-1.
\end{equation}
The resulting coordinates $\{(x_i, y_i)\}_{i=0}^{N-1}$ are concatenated into a $2N$-dimensional feature vector. This descriptor is invariant to translation and scale. We do not normalize rotation, since orientation and axis alignment are meaningful geometric properties of constructed building footprints. Notably, the visualization is done on the datasets prior to rotation augmentations. 

\paragraph{t-SNE Embedding.}
Feature vectors from all datasets are stacked into a single matrix:
\begin{equation}
    X \in \mathbb{R}^{M \times 2N},
\end{equation}
where $M$ is the total number of sampled footprints. We then apply t-SNE to obtain a 2D embedding. The embedding is initialized with PCA to improve stability and reduce sensitivity to random initialization. \cref{tab:tsne_hyperparameters} summarizes the hyperparameters used in the analysis.

\begin{table}[h]
\centering
\footnotesize
\caption{Hyperparameters used for the t-SNE analysis of massing footprints.}
\label{tab:tsne_hyperparameters}
\begin{tabular}{lc}
    \toprule
    \textbf{Parameter} & \textbf{Value} \\
    \midrule
    Boundary sample points $N$ & 64 \\
    Maximum samples per dataset & 10,000 \\
    t-SNE perplexity & 40 \\
    t-SNE iterations & 1,000 \\
    t-SNE initialization & PCA \\
    \bottomrule
\end{tabular}
\end{table}

\paragraph{Interpretation.}
The resulting embedding, presented in \cref{fig:tsne_data}, contains several smooth, approximately 1D structures. These arise naturally from simple low-complexity footprints. For example, after translation and area normalization, a rectangular footprint is largely parameterized by a single geometric degree of freedom, its aspect ratio $r = W/H$. As $r$ varies, the uniformly sampled boundary points change continuously along the four edges of the rectangle, tracing a smooth 1D manifold in the high-dimensional descriptor space. t-SNE preserves this local structure, causing rectangular footprints to appear as visible lines or arcs in the 2D embedding.

Similarly, L-shaped, T-shaped, and other low-vertex footprints form lower-dimensional submanifolds that appear as shorter curves or branches. In contrast, irregular many-vertex footprints, which are more common in the OSM data, do not follow the same low-complexity structure and are distributed more broadly across the embedding. This pattern supports the interpretation that real-world OSM massings occupy regions of shape space that are only partially covered by the procedurally generated synthetic dataset.

\begin{figure}[b!]
    \centering
    \includegraphics[width=0.9\linewidth]{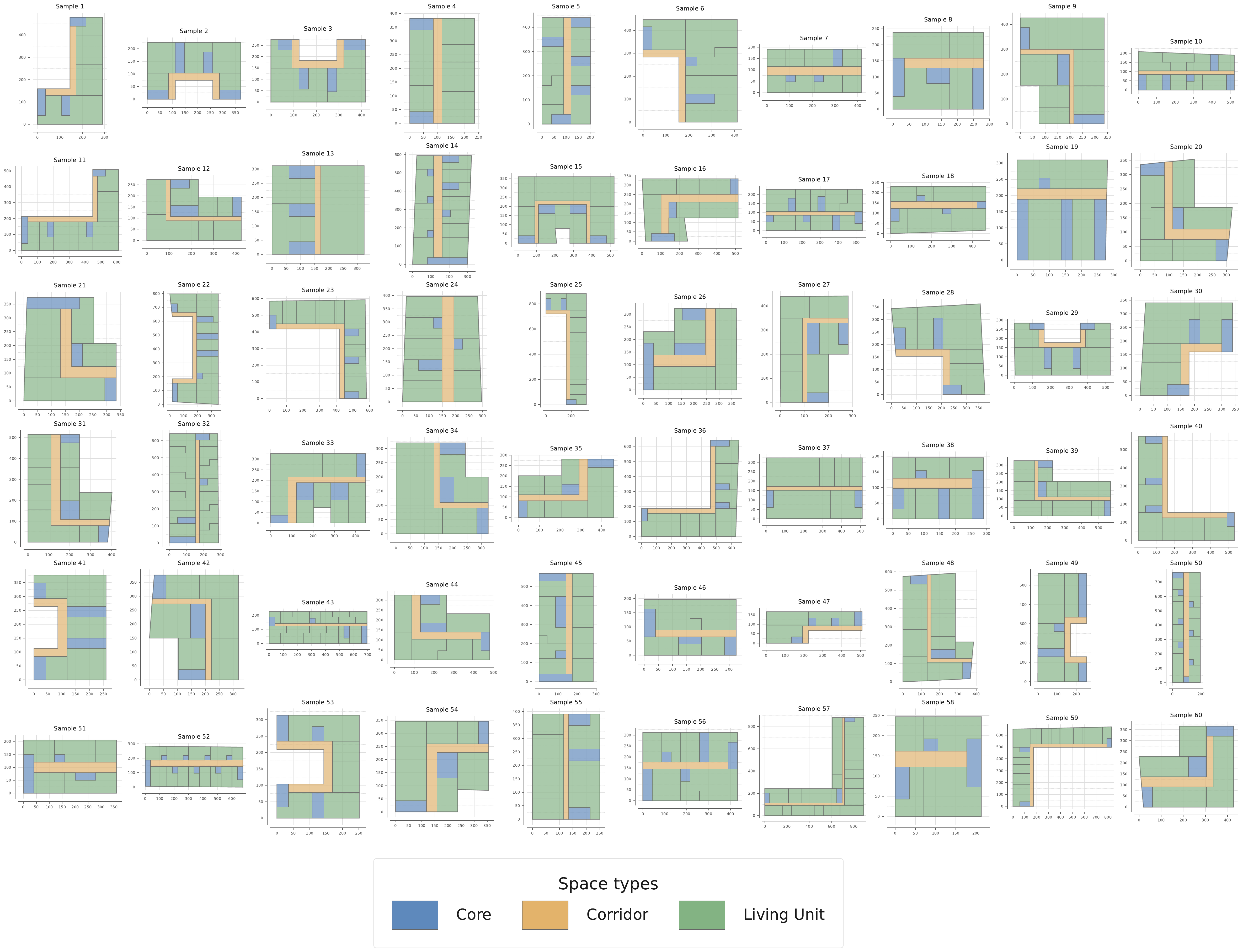}
    \caption{Samples from the \textbf{synthetic} dataset. Each sample contains the building massing, metadata, and a procedurally generated space layout consisting of cores, corridors, and living units.}
    \label{fig:synthetic_samples}
\end{figure}

\begin{figure}[b!]
    \centering
    \includegraphics[width=0.9\linewidth]{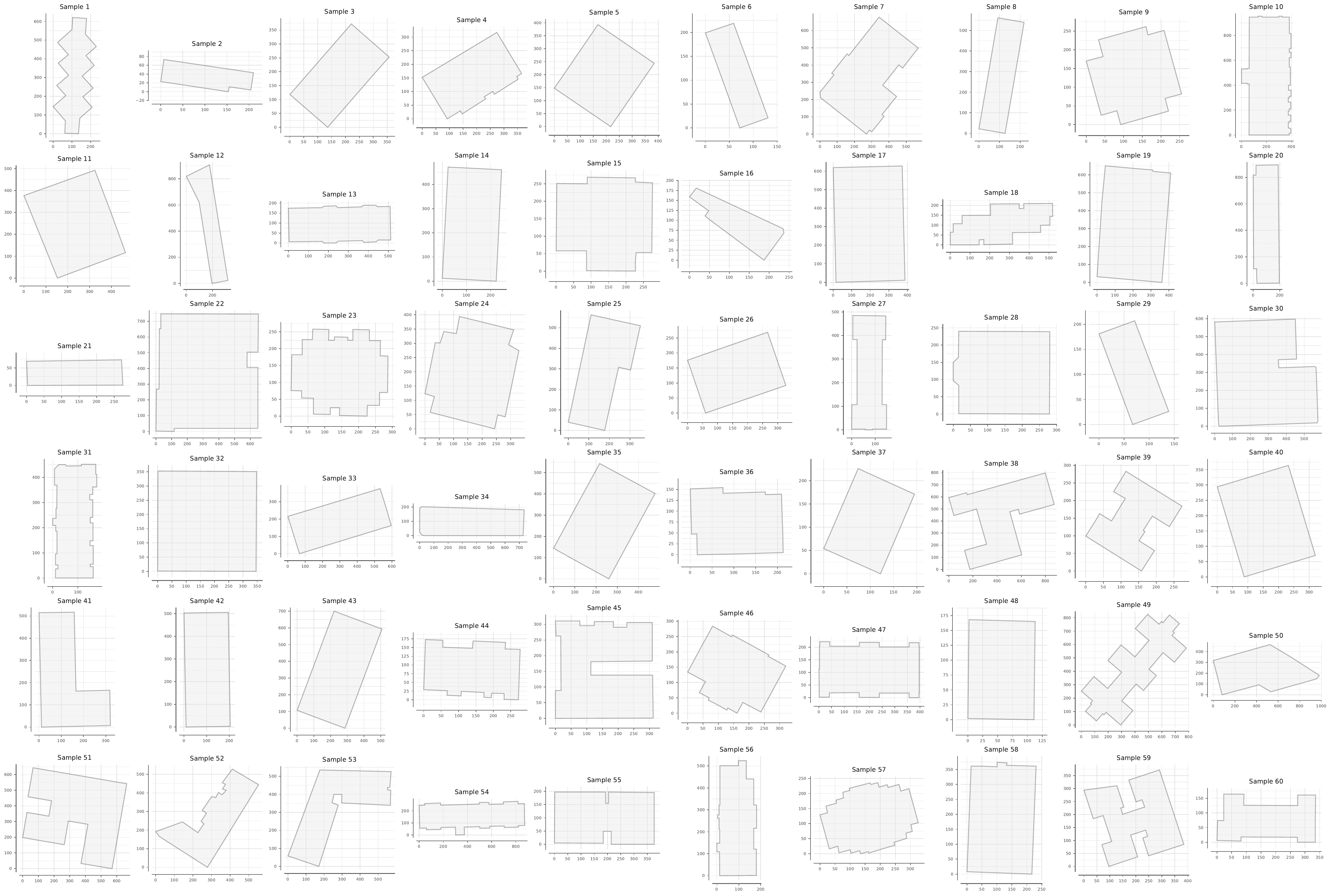}
    \caption{Samples from the \textbf{OSM} dataset. Each sample contains only the building massing and metadata, without ground-truth space layouts, and therefore cannot be used for pre-training.}
    \label{fig:osm_samples}
\end{figure}

\begin{figure}[b!]
    \centering
    \includegraphics[width=0.9\linewidth]{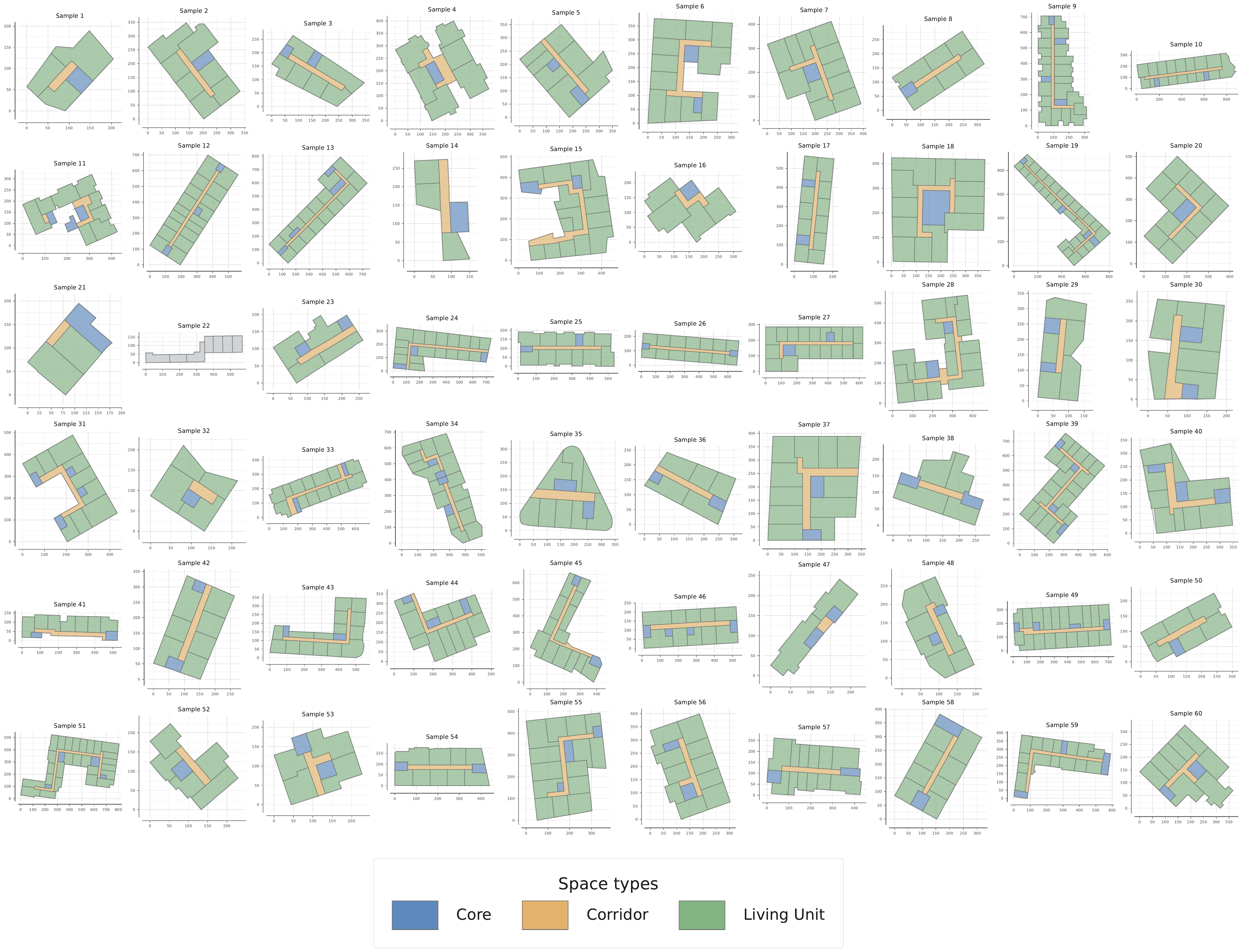}
    \caption{Samples from the \textbf{SFT} dataset. Each sample contains the building massing, metadata, and an architect-edited space layout consisting of living units and, when needed, cores and corridors.}
    \label{fig:sft_samples}
\end{figure}

\subsection{Data Samples}
\label[appendix]{app:data_samples}

\cref{fig:synthetic_samples,fig:osm_samples,fig:sft_samples} show representative samples from the OpenStreetMap (OSM), synthetic, and SFT datasets. Synthetic samples include procedurally generated space layouts, consisting of cores, corridors, and living units, in addition to the building massing and metadata. In contrast, OSM samples contain only the building massing and associated metadata, without ground-truth space layouts. The SFT samples, described in detail in \cref{app:human_feedback_collection}, are drawn from a curated set of building footprints and include architect-edited layouts. These edited layouts consist of living units and, depending on the massing scale and design requirements, may also include cores and corridors.

%% file: appendix/human_feedback_collection.tex
\subsection{Labeler Information}
\label[appendix]{app:labeler_info}
The labelers were 10 practicing architects hired through a BIM consultancy contractor. They were instructed to evaluate and edit model-generated outputs according to architectural conventions commonly observed in North America.

\subsection{Labeling Instructions and Interface}
\begin{figure}[h!]
    \centering
    \includegraphics[width=0.82\textwidth]{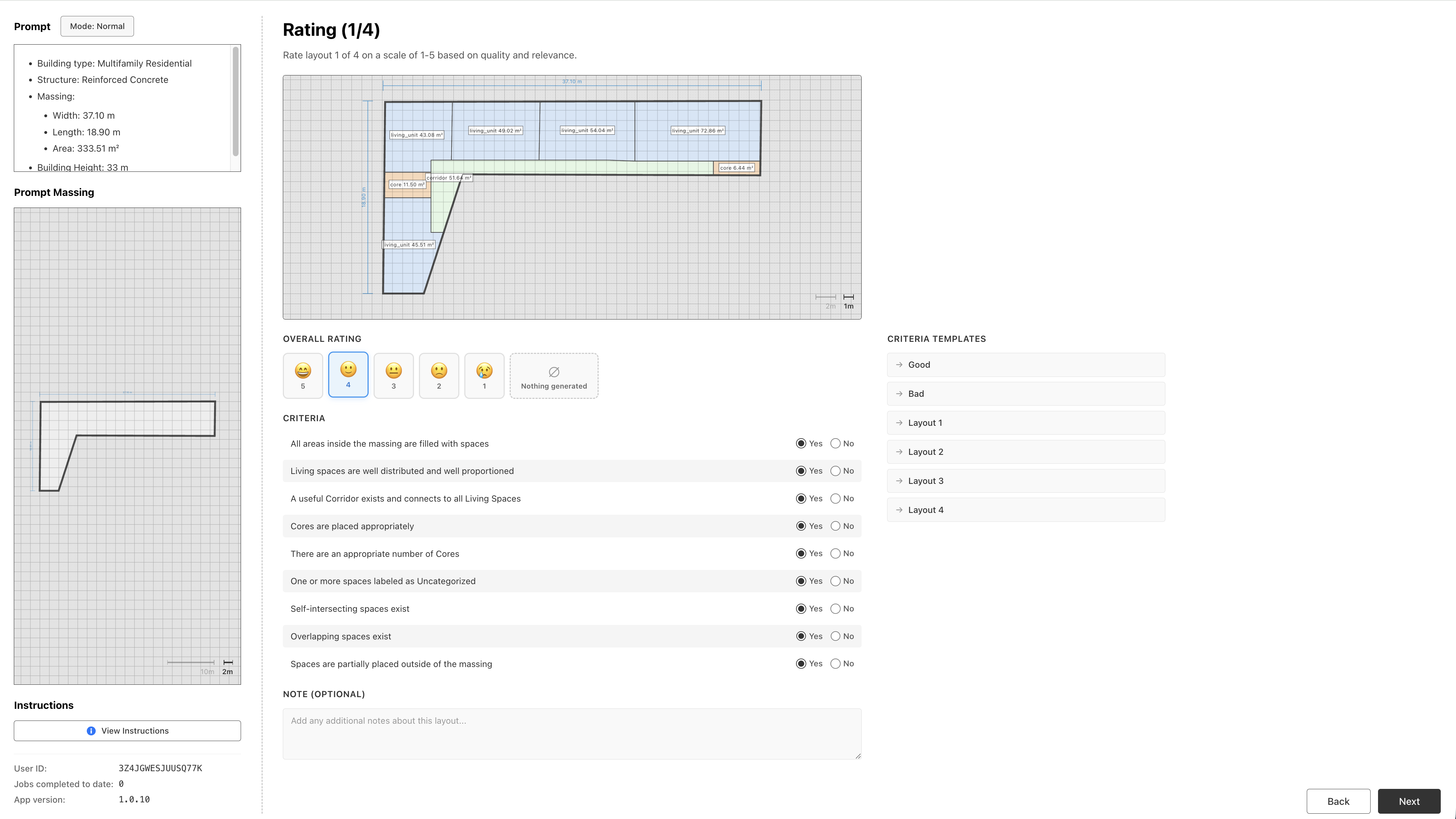}
    \caption{Labelers independently evaluate each model generation according to the criteria.}
    \label{fig:labeling_ui_rating}
\end{figure}

\begin{figure}[h!]
    \centering
    \includegraphics[width=0.82\textwidth]{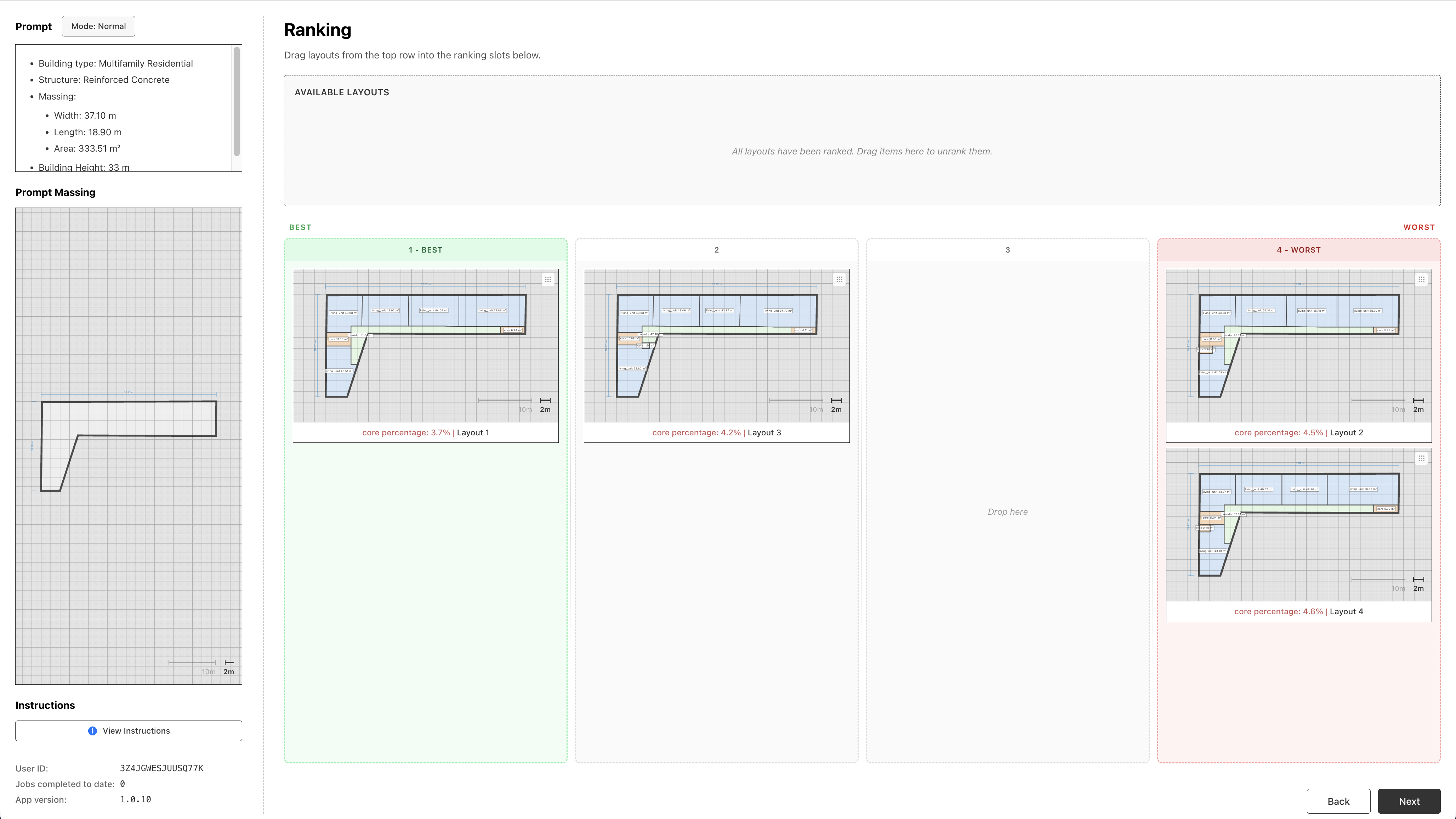}
    \caption{Model-generated outputs are ranked relative to one another, with ties permitted.}
    \label{fig:labeling_ui_ranking}
\end{figure}

\begin{figure}[h!]
    \centering
    \includegraphics[width=0.82\textwidth]{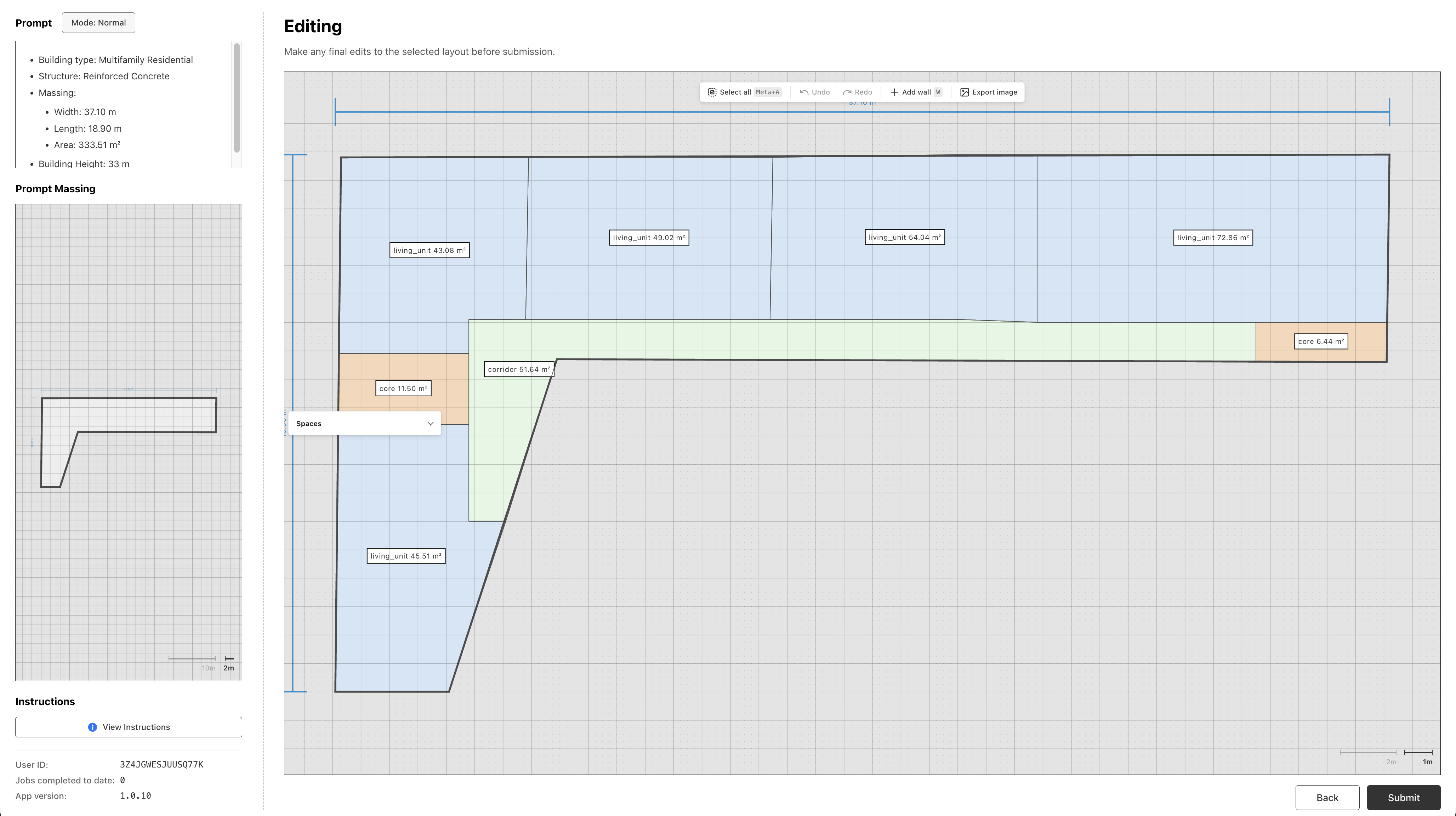}
    \caption{The highest-ranked model-generated output is subsequently edited by the labeler.}
    \label{fig:labeling_ui_editing}
\end{figure}

\label[appendix]{app:labeling_info}
The instructions provided to the labelers are summarized below. In short, a massing is first sampled from a curated dataset, after which four model-generated layouts are produced. Each layout is independently evaluated by the labeler according to a predefined set of criteria. The layouts are subsequently ranked relative to one another, with ties allowed. Finally, the highest-ranked layout is edited by the labeler to correct any remaining issues. \cref{fig:labeling_ui_editing,fig:labeling_ui_rating,fig:labeling_ui_ranking} show screenshots of our labeling interface. 

\begin{tcolorbox}[
    title=Instructions Provided to Labelers,
    colback=gray!5,
    colframe=black,
    fonttitle=\bfseries,
    breakable
]
\vspace{0.5em}

\textbf{Introduction}

You are given a randomly generated prompt that includes the following inputs:

\begin{itemize}[noitemsep, topsep=2pt, leftmargin=12pt]
    \item Building Type: Multifamily residential
    \item Structure: Either wood frame or concrete frame
    \item Massing: A length and width to define a massing envelope
\end{itemize}

\vspace{0.5em}

The LLM will generate four floor plan layouts based on this prompt. It can generate and arrange the following types of spaces to fit within the mass envelope:

\begin{itemize}[noitemsep, topsep=2pt, leftmargin=12pt]
    \item Cores: The model does not currently differentiate between elevators and stairs. Use your judgement to evaluate and edit the cores assuming they contain either stairs alone, elevators alone, or a combination of both depending on the layout.
    \item Corridors: Corridors should be between 1m (3ft) and 2m (6ft)
    \item Living Units: Use this as a guide for a "good" layout of living units:
    \begin{itemize}[noitemsep, topsep=2pt, leftmargin=12pt]
        \item Studio/1-bedroom: 30–60 m² (320–650 ft²)
        \item 2-bedroom: 60–90 m² (650–970 ft²)
        \item 3-bedroom: 90–120 m² (970–1,300 ft²)
    \end{itemize}
\end{itemize}

\vspace{1em}

\textbf{Step 1: Rating Each of the Four Layouts}

For each of the four generated layouts, provide:
\begin{itemize}[noitemsep, topsep=2pt, leftmargin=12pt]
    \item An overall rating from 5 (highest) to 1 (lowest)
    \item Responses to a set of yes/no evaluation questions
\end{itemize}

\vspace{0.5em}

\textbf{Step 1a: Rate Each Layout (Scale 1--5)}

Rate each layout using the following scale:

\begin{itemize}[noitemsep, topsep=2pt, leftmargin=12pt]
    \item 5: Architecturally appropriate, geometrically perfect, fully labeled, ready to use
    \item 4: Good quality with minor issues that are easily correctable (e.g., fixable in approximately 1 minute)
    \item 3: Moderate issues requiring some corrections, but salvageable (e.g., fixable in 2--3 minutes)
    \item 2: Many issues requiring significant corrections, but salvageable (e.g., fixable in 4--5 minutes)
    \item 1: Major architectural or geometric flaws; may not be worth correcting (e.g., would require more than 5 minutes to fix)
    \item Nothing Generated: The layout is completely empty
\end{itemize}

\vspace{0.5em}

\textbf{Criteria for Rating}

\begin{itemize}[noitemsep, topsep=2pt, leftmargin=12pt]
    \item Architectural Correctness: The floor plan should support well-proportioned living spaces and sensible circulation.
    \item Usability and Clarity: Layouts should be practical and clearly labeled.
    \item Completeness: All required elements should be present (cores, corridors, living spaces).
    \item Core Placement and Size: Cores should be appropriately placed and sized, without redundancy.
    \item Corridor Placement: Corridors should provide proper connectivity to all spaces.
    \item Geometric Correctness: No overlapping spaces or empty areas within the massing. Spaces highlighted in red indicate geometric errors and should negatively impact the rating.
\end{itemize}

\vspace{0.5em}

\textbf{Step 1b: Individual Criteria (Yes/No Questions)}

Answer the following questions to the best of your ability:

\begin{itemize}[noitemsep, topsep=2pt, leftmargin=12pt]
    \item All areas inside the massing are filled with spaces.
    \item Living spaces are well distributed and well proportioned.
    \item A useful corridor exists and connects to all living spaces.
    \item Cores are placed appropriately.
    \item There are an appropriate number of cores.
    \item One or more spaces are labeled as ``Uncategorized''.
    \item Self-intersecting spaces exist.
    \item Overlapping spaces exist.
    \item Spaces are partially placed outside of the massing.
\end{itemize}

\vspace{0.5em}

Add any additional notes that may help explain your evaluation decisions (optional).

\vspace{1em}

\textbf{Step 2: Rank Layouts}

Rank all layouts from best to worst based on your evaluation. Multiple layouts may be assigned the same rank, and some ranking slots may remain empty.

\vspace{0.5em}

\textbf{Ranking Guidelines}

\begin{itemize}[noitemsep, topsep=2pt, leftmargin=12pt]
    \item Prioritize geometric and architectural correctness over minor labeling issues.
    \item A layout with one major but fixable issue may rank higher than one with many minor unfixable issues.
    \item A layout with correct geometry but incorrect labels is more valuable than one with correct labels but overlapping rooms.
    \item Consider which layout would be most useful to an architect beginning the design process with the fewest required edits.
    \item Incorrect layouts can lead to construction errors, wasted materials, or safety concerns; prioritize correctness over creativity.
    \item When uncertain, ask: ``Which layout would I rather receive if I needed to create a working floor plan today?''
\end{itemize}

\vspace{1em}

\textbf{Step 3: Edit the Best Layout}

You will be presented with the highest-rated layout based on the ratings and rankings you provided. Using the simplified editor, make corrections according to the following guidelines.

\vspace{0.5em}

\textbf{Editor Capabilities}

\begin{itemize}[noitemsep, topsep=2pt, leftmargin=12pt]
    \item Draw a wall to separate one space into two
    \item Select and edit a wall
    \item Delete a wall
    \item Drag wall endpoints
    \item Modify space labels
\end{itemize}

\vspace{0.5em}

\textbf{Editing Principles}

\begin{itemize}[noitemsep, topsep=2pt, leftmargin=12pt]
    \item Be diligent and accurate; floor plan errors can have real-world consequences.
    \item Make precise modifications; small dimensional changes may be important.
    \item Prioritize corrections that affect structural integrity and usability.
\end{itemize}

\pagebreak
\vspace{0.5em}

\textbf{High-Priority Editing Objectives (in descending priority order)}

\vspace{0.5em}

\textbf{1. Fix Geometric Issues (Highest Priority)}

\begin{itemize}[noitemsep, topsep=2pt, leftmargin=12pt]
    \item Fix spaces that extend outside the massing.
    \item Ensure all spaces completely fill the massing.
    \item Eliminate overlapping polygons.
    \item Remove or merge tiny polygons.
    \item Resolve self-intersecting polygons.
\end{itemize}

\vspace{0.5em}

\textbf{2. Fix Critical Architectural Issues}

\begin{itemize}[noitemsep, topsep=2pt, leftmargin=12pt]
    \item Correct core placement and sizing.
    \item Ensure corridor connectivity to all living spaces.
    \item Adjust living space dimensions if severely incorrect.
\end{itemize}

\vspace{0.5em}

\textbf{3. Fix Room Labels}

\begin{itemize}[noitemsep, topsep=2pt, leftmargin=12pt]
    \item Ensure all spaces are correctly labeled.
    \item Verify labels match the intended space function.
\end{itemize}
\end{tcolorbox}

\subsection{Human Feedback Data Processing}
\label[appendix]{app:feedback_processing}

Human feedback data is collected through annotation sessions in which labelers evaluate 4 model-generated building layouts for a given prompt. For each candidate layout $k$, the labeler provides:
\begin{enumerate}[noitemsep,topsep=0pt]
    \item An ordinal rank \(r_k \in \{1,\ldots,4\}\), where lower rank is better.
    \item A scalar rating \(s_k \in [1,5]\).
    \item A binary checklist of design and geometry criteria.
    \item A human-edited layout.
\end{enumerate}

\paragraph{Basic Quality Filtering.}
Before constructing any training dataset, we apply several deterministic filters. First, we remove skipped annotation sessions and any candidate layout with missing or sentinel labels, such as a rank or rating of \(-1\) or \texttt{null}. Second, each candidate layout is converted from its structured JSON representation into our compact DSL representation using the DSL encoder. Candidates whose DSL serialization fails or exceeds a certain wall-clock timeout are removed. Any preference pair or training example depending on a failed serialization is also removed.

Finally, we remove duplicate preference pairs. In particular, if the chosen and rejected
layouts in a pair produce identical DSL strings after serialization, the pair is discarded, since it provides no useful preference signal.

\paragraph{Rank and Rating Consistency.} Each annotation contains both a rank and a rating. These two signals should generally agree, but noisy annotations may contain contradictions. For example, a layout may be ranked above another layout while receiving a lower rating. To reduce such noise, we keep only pairs whose rank order is consistent with their ratings.

For two candidate layouts \(a\) and \(b\), with ranks \(r_a, r_b\) and ratings \(s_a, s_b\), we
define
\[
\operatorname{consistent}(a,b)=
\begin{cases}
s_a \geq s_b, & \text{if } r_a < r_b, \\
s_b \geq s_a, & \text{if } r_a > r_b, \\
\top, & \text{if } r_a = r_b.
\end{cases}
\]
Thus, if layout \(a\) is ranked better than layout \(b\), then its rating must be at least as high as the rating of \(b\). Ties in rank are always treated as
consistent. This check is applied to all pairwise combinations within each annotation session. Pairs that fail the check are excluded from preference training.

\paragraph{Geometric Filtering of Edited Completions.}
Human-edited layouts are used as SFT targets, but they may still contain geometric errors. We therefore filter edited completions using an automated test based on the geometric correctness verifier, described in \cref{sec:verifiable_rewards} and \cref{app:verifiable_rewards}.

\paragraph{Rotation Augmentation.} Architectural layouts are invariant to planar rotation; i.e., rotating a valid layout usually preserves its design structure and geometric validity. We use this property to augment the data utilizing the same approach used for the pre-training dataset, described in \cref{sec:data}.

For each training example, we generate rotated variants by sampling an angle \(\theta\) from three regimes:
$$
\{\theta_i\}_{i=1}^{12}
\sim
4\,\operatorname{Uniform}(0^\circ,10^\circ)
+4\,\operatorname{Uniform}(10^\circ,350^\circ)
+4\,\operatorname{Uniform}(350^\circ,360^\circ),
$$
This produces both small near-identity perturbations and large rotations. If any rotated coordinate falls outside the tokenizer coordinate range of 1024 grid units, the augmented example is discarded. Therefore, each training example produces at most \(3N_{\mathrm{aug}}\) rotated variants, corresponding to a nominal
training-set expansion factor of $3N_{\mathrm{aug}} + 1$, where $N_{\mathrm{aug}}$ was set to 4.

\Cref{tab:feedback_filtering} summarizes the number of datapoints removed during human-feedback data filtering. After this filtering step, \cref{tab:feedback_info} summarizes the size of the remaining human-feedback datasets before and after rotation augmentation. 

We reserve 10\% of the filtered data for validation. To prevent information leakage, the split is performed at the annotation-session level, so that all candidate layouts and pairwise preferences associated with the same prompt are assigned entirely to either the training or validation split. Training and validation examples are augmented separately so that rotated variants of the same example do not appear across splits.

\begin{table}[h!]
\centering
\caption{Number of human feedback datapoints removed during data filtering and processing.}
\label{tab:feedback_filtering}
\footnotesize
\begin{tabular}{llr}
    \toprule
    \textbf{Category} & \textbf{Reason} & \textbf{Count} \\
    \midrule
    Skipped & Rating $-1$ & 61 \\
    Skipped & Inconsistent pairs & 575 \\
    Skipped & Failed geometric checks & 287 \\
    Errors & Failed DSL encodings & 227 \\
    \midrule
    \multicolumn{2}{l}{\textbf{Total}} & \textbf{1,150} \\
    \bottomrule
\end{tabular}
\end{table}

\begin{table}[h!]
\centering
\caption{Number of datapoints in the human-feedback dataset after filtering, before and after rotation augmentation.}
\label{tab:feedback_info}
\footnotesize
\begin{tabular}{lcc}
    \toprule
    \textbf{Dataset} 
    & \textbf{Before Aug.} 
    & \textbf{After Aug.} \\
    \midrule
    SFT completions 
    & 10.4k 
    & 135.4k \\
    Pairwise preferences 
    & 90.6k 
    & 1.2M \\
    \bottomrule
\end{tabular}
\end{table}

%% file: appendix/training_hparams.tex
\subsection{Pre-training Settings and Hyperparameters}
\label[appendix]{app:pretraining_hparams}

\paragraph{Tokenization Efficiency.}
To improve token efficiency, we use a tokenizer specialized for the DSL representation. Its vocabulary is composed primarily of DSL keywords, delimiters, and quantized numeric tokens, which better match the structure of the generated architectural sequences. As shown in \cref{fig:seq_lengths}, the DSL tokenizer substantially shortens the sequence-length distribution compared with the original Qwen3 tokenizer. The statistics in \cref{tab:tokenizer_stats} quantify this effect, showing reductions in the mean, median, 95th-percentile, and total token counts. This reduction is especially important after augmentation, where the total token budget decreases from 38.5B tokens with the Qwen3 tokenizer to 17.6B tokens with the DSL tokenizer.

\begin{table}[h!]
\centering
\caption{Tokenization statistics by tokenizer. ``Before Aug.'' and ``After Aug.'' denote total token counts before and after augmentation.}
\footnotesize
\begin{tabular}{lcccccc}
    \toprule
    \textbf{Tokenizer} & \textbf{Mean} & \textbf{Median} & \textbf{P95} & \textbf{Max} & \textbf{Before Aug.} & \textbf{After Aug.} \\
    \midrule
    DSL Tokenizer   & 235.9 & 219.0 & 373.0 & 785   & 880M & 17.6B \\
    Qwen3 Tokenizer & 516.6 & 475.0 & 844.0 & 1,902 & 1.9B & 38.5B \\
    \bottomrule
\end{tabular}
\label{tab:tokenizer_stats}
\end{table}

\paragraph{Effective Parameter Counts.}
\cref{tab:pretraining_hparams} reports the effective parameter counts after resizing the embedding layers to match the DSL tokenizer vocabulary. The tokenizer contains 1367 tokens and is resized to 1408, the nearest multiple of 64, to improve embedding efficiency on modern hardware.

\paragraph{Pre-Training.}
All models are optimized using AdamW \citep{kingma2014adam, loshchilovdecoupled} with a weight decay of 0.01 and a cosine learning rate schedule with 4000 warmup steps and a minimum learning rate of 1e-6. We maintain a fixed effective batch size of 1024 sequences across all experiments, while tuning the learning rate separately for each model size, as shown in \cref{tab:pretraining_hparams}. Sequence lengths are dynamically varied during training and padded to the maximum sequence length within each batch. All models are trained for a maximum of 200k optimizer steps, with the best checkpoint selected based on validation loss. Pretraining takes approximately three epochs with this effective batch size.

Training is performed with PyTorch \citep{li13pytorch} in bfloat16 mixed precision using FlashAttention-2 \citep{dao2024flashattention}, gradient checkpointing \citep{chen2016training}, and Liger kernels \citep{hsu2024liger} to improve memory efficiency and throughput. All experiments were conducted on NVIDIA H100 GPUs using either Distributed Data Parallel (DDP) \citep{li13pytorch} or DeepSpeed \citep{rajbhandari2020zero, rasley2020deepspeed}, depending on model scale, with the number of GPUs adjusted to maintain a consistent effective batch size across experiments. 

\begin{table}[h!]
\centering
\caption{Effective parameter counts after resizing the embedding layers to match the DSL tokenizer vocabulary, and peak learning rates used during pre-training.}
\label{tab:pretraining_hparams}
\footnotesize
\begin{tabular}{lcc}
    \toprule
    \textbf{Base Model} & \textbf{Effective Params} & \textbf{Learning Rate} \\
    \midrule
    Qwen3-0.6B & 440M & 3e-4 \\
    Qwen3-1.7B & 1.4B & 4e-4 \\
    \midrule
    Pythia-70M      & 20M  & 2e-4 \\
    Pythia-160M & 90M & 6e-5\footnotemark \\
    Pythia-410M     & 310M & 3e-4 \\
    Pythia-1.4B     & 1.2B & 3e-4 \\
    \bottomrule
\end{tabular}
\end{table}
\footnotetext{A lower learning rate was used because training was unstable at higher learning rates.}

\subsection{Post-training Settings and Hyperparameters}
\label[appendix]{app:post_hparams}

Our post-training setup is based on TRL \citep{vonwerra2020trl} and PyTorch \citep{paszke2019pytorch}. Across all post-training stages, we use full-parameter fine-tuning with AdamW optimization \citep{kingma2014adam, loshchilovdecoupled} with a cosine learning rate schedule, full bfloat16 precision, and gradient checkpointing \citep{chen2016training}. All experiments were conducted on NVIDIA H100 GPUs using Distributed Data Parallel (DDP) \citep{li13pytorch}. Within each post-training stage, the effective batch size is held constant across model scales, while learning rates are tuned separately for each model and stage. The resulting hyperparameters are reported in \cref{tab:post_hparams}.

\paragraph{Supervised Fine-Tuning.}
We perform supervised fine-tuning (SFT) by fine-tuning all model parameters from the pretrained base checkpoint. SFT uses a 10\% warmup ratio and no weight decay. We train for 2k optimization steps with an effective batch size of 256.

\paragraph{Reward Modeling.}
We train the reward model (RM) by fine-tuning all parameters of the SFT checkpoint on pairwise preference data. The RM warmup ratio is 10\%, and weight decay is set to 0.01. We train for 5k optimization steps with an effective batch size of 256.

\paragraph{GRPO.}
We fully fine-tune the SFT model using GRPO with verifiable DSL rewards and the RM. GRPO uses 500 warmup steps, no weight decay, and a maximum gradient norm of 1.0. We train for 5k optimization steps with an effective batch size of 256. The coefficients for the reward model and verifiable rewards are set to $\lambda_{\text{RM}} = \lambda_i = 1$, and the reward model normalization scale in \cref{eq:rm_normalization} is set to $\alpha = 3$.

For each prompt, we sample 8 completions with temperature 1.0, top-$p$ 0.95, and a maximum completion length of 2048 tokens. The KL coefficient is set to 0; in our experiments, we observed better stability without the KL term. This also removes the need to load a reference model, enabling a larger effective batch size. Generation is accelerated with collocated vLLM \citep{kwon2023efficient}.

\begin{table}[h!]
\centering
\footnotesize
\caption{Peak learning rates used during post-training.}
\label{tab:post_hparams}
\begin{tabular}{lcccc}
\toprule
\textbf{Model} & \textbf{SFT} & \textbf{RM} & \textbf{GRPO} \\
\midrule
    Qwen3-0.6B & 2e-4 & 8e-5 & 6e-5 \\
    Qwen3-1.7B & 2e-4 & 3e-4\footnotemark & 8e-5 \\
    \midrule
    Pythia-410M & 2e-4 & 8e-5 & 4e-5 \\
    Pythia-1.4B & 2e-4 & 8e-5 & 4e-5 \\
\bottomrule
\end{tabular}
\end{table}
\footnotetext{A higher learning rate was used because the model was under-trained at lower learning rates.}

%% file: appendix/verifiable_rewards.tex
As outlined in \cref{sec:verifiable_rewards}, human feedback alone is insufficient to guarantee adherence to specific design requirements. We therefore employ verifiable rewards to automatically evaluate and enforce geometric and functional constraints. Here, we describe the verifiable rewards in more detail.

Unless otherwise stated, continuous constraint violations are converted to rewards using a smooth-step function. For a normalized error $e \in [0,1]$ and tolerance $\tau$, we define
\begin{equation}
    S_{\tau}(e) =
    \begin{cases}
        1, & e \leq \tau, \\[2pt]
        \exp\left(-\frac{e-\tau}{\tau}\right), & e > \tau.
    \end{cases}
    \label{eq:smooth_reward}
\end{equation}
This mapping provides full reward within a small tolerance around the desired constraint and decays exponentially once the tolerance is exceeded. All individual reward components are bounded to $[0,1]$.

\subsection{Geometric Correctness}
Geometric correctness evaluates whether the generated space polygons form a valid partition of the input building massing. We compute three spatial consistency signals from the parsed polygon geometries:

\begin{enumerate}[noitemsep,topsep=0pt]
    \item \textbf{Containment.} Generated spaces should lie within the building massing. The verifier computes the normalized fraction of space geometry outside the massing, denoted $e_{\mathrm{out}}$.
    \item \textbf{Non-overlap.} Generated spaces should not overlap one another. The verifier computes the normalized overlap violation $e_{\mathrm{ov}}$ across the generated spaces.
    \item \textbf{Massing coverage.} The generated spaces should collectively occupy the complete building footprint. Let $c_{\mathrm{mass}} \in [0,1]$ denote the fraction of the massing covered by the union of the generated spaces. We define the coverage error as $e_{\mathrm{cov}} = 1 - c_{\mathrm{mass}}$. Coverage is rewarded only when the containment constraint is satisfied within tolerance. This prevents a layout from obtaining a high coverage score by extending spaces beyond the massing boundary.
\end{enumerate}

For geometric correctness, we use a tolerance of $\tau_{\mathrm{geom}}=0.01$. The three component rewards are therefore
\begin{align}
    r_{\mathrm{contain}} &= S_{\tau_{\mathrm{geom}}}(e_{\mathrm{out}}), \\
    r_{\mathrm{overlap}} &= S_{\tau_{\mathrm{geom}}}(e_{\mathrm{ov}}), \\
    r_{\mathrm{coverage}} &=
    \begin{cases}
        S_{\tau_{\mathrm{geom}}}(e_{\mathrm{cov}}),
        & e_{\mathrm{out}} \leq \tau_{\mathrm{geom}}, \\[2pt]
        0,
        & \text{otherwise}.
    \end{cases}
\end{align}

The final geometry reward is computed as the geometric mean of the three component scores,
\begin{equation}
    R_{\mathrm{geom}}
    =
    \operatorname{GM}
    \left(
        r_{\mathrm{contain}},
        r_{\mathrm{overlap}},
        r_{\mathrm{coverage}}
    \right),
    \label{eq:geom_reward}
\end{equation}
where
\begin{equation}
    \operatorname{GM}(r_1,\ldots,r_n)
    =
    \left(
        \prod_{i=1}^{n}
        \max(r_i,\varepsilon)
    \right)^{1/n}.
    \label{eq:geom_mean}
\end{equation}
In practice, the geometric mean is computed in log space using
$\exp(\frac{1}{n}\sum_i \log(\max(r_i,\varepsilon)))$, with
$\varepsilon=10^{-8}$ for numerical stability. Unlike an arithmetic mean, this aggregation prevents a high score on one constraint from compensating for a severe violation of another.

Outputs that cannot be successfully parsed, geometrically evaluated, or that contain invalid polygons such as self-intersecting or degenerate geometries receive zero reward.

\subsection{Functional Compliance}
Functional compliance evaluates residential design requirements using the parsed space labels, counts, and polygon areas. We consider four criteria:

\begin{enumerate}[noitemsep,topsep=0pt]
    \item \textbf{Core count.} The number of vertical circulation cores must fall between one and four.
    \item \textbf{Corridor count.} The number of corridors must fall between one and three.
    \item \textbf{Unit size.} At least 95\% of living units must satisfy the minimum area requirement of $A_{\min}=30$ square meters.
    \item \textbf{Unit labeling.} At least 95\% of living units must satisfy the expected labeling requirement.
\end{enumerate}

As with geometric correctness, each criterion is mapped to a continuous score in $[0,1]$. We use a smooth tolerance of $\tau_{\mathrm{func}}=0.05$, providing full reward when a requirement is satisfied and exponentially decreasing reward as the violation increases.

For count-based constraints with an acceptable interval $[\ell,u]$, we define the normalized violation as
\begin{equation}
    e_{\mathrm{range}}(x;\ell,u)
    =
    \begin{cases}
        \dfrac{\ell-x}{\max(\ell,1)}, & x < \ell, \\[6pt]
        0, & \ell \leq x \leq u, \\[6pt]
        \dfrac{x-u}{\max(u,1)}, & x > u.
    \end{cases}
    \label{eq:functional_range_error}
\end{equation}
The corresponding core and corridor rewards are
\begin{align}
    r_{\mathrm{core}}
    &=
    S_{\tau_{\mathrm{func}}}
    \left(
        e_{\mathrm{range}}(n_{\mathrm{core}};1,4)
    \right), \\
    r_{\mathrm{corridor}}
    &=
    S_{\tau_{\mathrm{func}}}
    \left(
        e_{\mathrm{range}}(n_{\mathrm{corridor}};1,3)
    \right).
\end{align}

For the unit-size and labeling requirements, let $p_{\mathrm{size}}$ denote the fraction of living units satisfying the minimum area requirement and $p_{\mathrm{label}}$ the fraction satisfying the labeling requirement. Their normalized violations are
\begin{align}
    e_{\mathrm{size}} &= \max\left(0,0.95-p_{\mathrm{size}}\right), \\
    e_{\mathrm{label}} &= \max\left(0,0.95-p_{\mathrm{label}}\right),
\end{align}
with corresponding rewards
\begin{align}
    r_{\mathrm{size}} &= S_{\tau_{\mathrm{func}}}(e_{\mathrm{size}}), \\
    r_{\mathrm{label}} &= S_{\tau_{\mathrm{func}}}(e_{\mathrm{label}}).
\end{align}

The final functional compliance reward is computed as the geometric mean of the four components:
\begin{equation}
    R_{\mathrm{func}}
    =
    \operatorname{GM}
    \left(
        r_{\mathrm{core}},
        r_{\mathrm{corridor}},
        r_{\mathrm{size}},
        r_{\mathrm{label}}
    \right).
    \label{eq:functional_reward}
\end{equation}
This aggregation ensures that deficiencies in any individual requirement substantially reduce the overall functional reward. Outputs that cannot be successfully parsed, evaluated, or that contain invalid polygon geometries receive zero reward.


\subsection{Soft Overlong Penalty}
To discourage excessively long completions and improve the stability of training, we apply a length-based penalty adapted from DAPO \citep{yu2026dapo} and implemented in TRL \citep{vonwerra2020trl}. Let $|y|$ denote the completion length, $L_{\max}$ the maximum allowed completion length, and $L_{\mathrm{cache}}$ a soft penalty window. The reward is defined as:
\begin{equation}
R_{\mathrm{length}}(y)=
\begin{cases}
0, & |y| \le L_{\max}-L_{\mathrm{cache}}, \\
\dfrac{(L_{\max}-L_{\mathrm{cache}})-|y|}{L_{\mathrm{cache}}}, & L_{\max}-L_{\mathrm{cache}} < |y| \le L_{\max}, \\
-1, & |y| > L_{\max}.
\end{cases}    
\end{equation}
This reward does not incentivize shorter completions. Instead, it imposes a progressively stronger penalty as the completion length approaches the maximum budget and assigns the maximum penalty once the budget is exceeded.

%% file: appendix/vlm_judge.tex
We evaluate perceptual and architectural layout quality using a
pairwise VLM judge protocol. For each evaluation prompt, two candidate
completions are generated from the same input: one from model $A$ and
one from model $B$. The judge compares the corresponding rendered floor plans
and assigns one of three outcomes: $A$ wins, $B$ wins, or tie. Outcomes are then
aggregated across prompts to obtain strict win, loss, and tie rates.

\subsection{VLM Judge Implementation}
\paragraph{Pair Construction.}
For each prompt, we select one completion from each model and render both
outputs as $784 \times 784$ pixel floor-plan images using the same deterministic
DSL parser and renderer used during training and evaluation. This ensures that the VLM judge observes only the geometric and semantic content induced by the generated
DSL.

\paragraph{Judgment Protocol.}
The two rendered floor plans are provided to a vision-language model in a
single comparison prompt. The images are labeled as \emph{Floorplan A} and
\emph{Floorplan B}, giving the judge unambiguous references that are independent
of their visual position in the prompt. The judge is instructed to evaluate the
layouts using the structured prompt shown below. The prompt asks the judge to
first record observations for each floor plan independently, then compare the two
layouts along six architectural criteria, and finally synthesize a concise
comparative assessment. The final verdict is required to appear inside
\texttt{<answer></answer>} tags as a JSON object with fields
\texttt{reasoning} and \texttt{winner}, where
\[
\texttt{winner} \in \{\texttt{"A"}, \texttt{"B"}, \texttt{"tie"}\}.
\]

\begin{tcolorbox}[
    title=System Prompt for Pairwise VLM Judge,
    colback=gray!5,
    colframe=black,
    fonttitle=\bfseries,
    breakable
]
\small
\begin{verbatim}
system_prompt: |-
  You are an expert architectural floorplan evaluator. You will be shown
  TWO rendered 2D floorplan images generated for the same design brief:
  Floorplan A (the first image) and Floorplan B (the second image).
  Your task is to decide which floorplan is the better architectural layout.

  The floorplans use the following visual conventions:
  - Light gray polygons with black edges: building massing (outer boundary)
  - Blue polygons: cores (stairs, elevators)
  - Orange polygons: corridors
  - Green polygons: living units (residential)  
  - Gray polygons: undefined spaces
  - Labels at polygon centroids indicate space types

  ## Evaluation criteria

  1. Spatial Organization (most important): Are spaces logically arranged?
     Do living units fill the massing efficiently? Are cores and
     corridors placed to provide good access to all units?

  2. Space Proportions: Are individual rooms reasonably shaped, not too
     narrow and not excessively elongated? Do they have practical aspect
     ratios?

  3. Coverage & Utilization: Does the layout fill the massing boundary well?
     Are there large gaps or undefined areas? Good layouts use most of the
     available floor area.

  4. Circulation Quality: Are corridors and cores positioned to allow
     efficient movement through the building? Is there clear access from
     corridors to units?

  5. Structural Logic: If columns/gridlines are present, are they placed in a
     regular grid pattern? Do they align with the building structure?

  6. Overlap & Validity: Do spaces avoid overlapping each other? Are all
     spaces contained within the massing boundary?

  First, summarize what you observe in each floorplan in a few short
  bullet points, listing Floorplan A and Floorplan B separately.

  Next, for each of the six criteria above, write one sentence comparing the
  two floorplans grounded in your observations, and state which is stronger on
  that criterion:
  - A = Floorplan A is stronger
  - B = Floorplan B is stronger
  - tie = the two are comparable on this criterion

  Based on the above, give a short summary assessment comparing the two
  layouts overall. Weight Spatial Organization most heavily, and treat validity
  failures, such as overlaps or spaces outside the massing, as strong evidence
  against a floorplan.

  After your synthesis, output your final verdict inside <answer></answer>
  tags as a JSON object with exactly two fields:
  - "reasoning": a brief 1-3 sentence comparative summary of the verdict
  - "winner": "A" if Floorplan A is the better layout, "B" if Floorplan B is
    the better layout, or "tie" if the two are of essentially equal quality

  Judge purely on the layouts shown; do not assume the image order implies
  quality.

  Example response structure:
  Observations:
  - Floorplan A: square footprint; 1 core, 1 corridor, 4 living units;
    no overlaps; near-full coverage
  - Floorplan B: same footprint; 2 living units, no core or corridor;
    large empty areas
  ...

  Criterion comparison:
  1. Spatial Organization [A]: A has a corner core with a corridor serving
     every unit, while B has no circulation infrastructure.
  2. Space Proportions [tie]: Both use rectangular units with practical
     proportions.
  ...

  Assessment:
  Floorplan A provides efficient circulation and near-complete coverage, while
  Floorplan B lacks any core/corridor and wastes most of the floor plate.
  Floorplan A is clearly the better layout.

  <answer>
  {
    "reasoning": "Floorplan A offers logical circulation and near-complete
    coverage, whereas B has no circulation infrastructure and large unused
    areas.",
    "winner": "A"
  }
  </answer>

  Always end your response with the <answer></answer> block. Do not put
  anything after the closing </answer> tag.

user_query: |-
  Compare these two floorplans. Floorplan A is the first image, and Floorplan B
  is the second. Follow the three-step protocol from the system instructions:
  (1) Observations for Floorplan A and Floorplan B, (2) Criterion comparison
  across all six criteria, (3) Synthesis, then the <answer> JSON block with
  the winner.
\end{verbatim}
\end{tcolorbox}

\paragraph{Order-bias Mitigation.}
To reduce sensitivity to presentation order, each pair is evaluated twice:
once in the original order $(A,B)$ and once in the swapped order $(B,A)$.
The verdict from the swapped comparison is mapped back to the original model
labels before aggregation. If both judgments agree after canonicalization, the
corresponding win direction is accepted. If the two judgments disagree, the
comparison is conservatively recorded as a tie.

\paragraph{Aggregation and Uncertainty.}
Let $n$ denote the number of evaluated prompts, and let $w_A$, $w_B$, and $t$
denote the number of strict wins for model $A$, strict wins for model $B$, and
ties, respectively. We report
\[
p_A = \frac{w_A}{n}, \qquad
p_{\mathrm{tie}} = \frac{t}{n}, \qquad
p_B = \frac{w_B}{n},
\]
so that $p_A + p_{\mathrm{tie}} + p_B = 1$. Ties are therefore reported as a
separate outcome rather than being split between the two models.

For each strict win rate, we report a 95\% Wilson score confidence interval.
For a binomial proportion $\hat{p}=w/n$ and $z=1.96$, the interval is
\[
\mathrm{CI}_{95\%}(\hat{p})=\left[\frac{\hat{p}+\frac{z^2}{2n}-z\sqrt{\frac{\hat{p}(1-\hat{p})}{n}+\frac{z^2}{4n^2}}}{1+\frac{z^2}{n}},\frac{\hat{p}+\frac{z^2}{2n}+z\sqrt{\frac{\hat{p}(1-\hat{p})}{n}+\frac{z^2}{4n^2}}}{1+\frac{z^2}{n}}\right],\quad z=1.96.
\]
These intervals quantify uncertainty due to finite prompt sampling within a
single evaluation run and do not capture cross-seed variation.

\subsection{Validation Against Architect Feedback}
To evaluate whether the VLM judge reflects expert architectural preferences, we apply the same pairwise judging protocol to preference pairs from the reward-model dataset. We use a VLM judge rather than the learned reward model for evaluation because FLOORA is directly optimized using that reward model during post-training, which would bias the evaluation in favor of our models.

For each pair, we render both layouts and evaluate them using the same VLM judge protocol described above, including evaluation under both presentation orders. Across approximately 10k model--architect pairs, the VLM judge agrees with the architect-implied preference in \textbf{77.5\% }of cases. This provides direct evidence that the VLM judge is meaningfully aligned with expert architectural corrections, while also indicating that it does not perfectly reproduce human judgment.

\subsection{Examples of the VLM Judge Outcomes}
\begin{figure}[h!]
    \centering
    \begin{subfigure}[b]{0.75\textwidth}
        \centering
        \includegraphics[width=\textwidth]{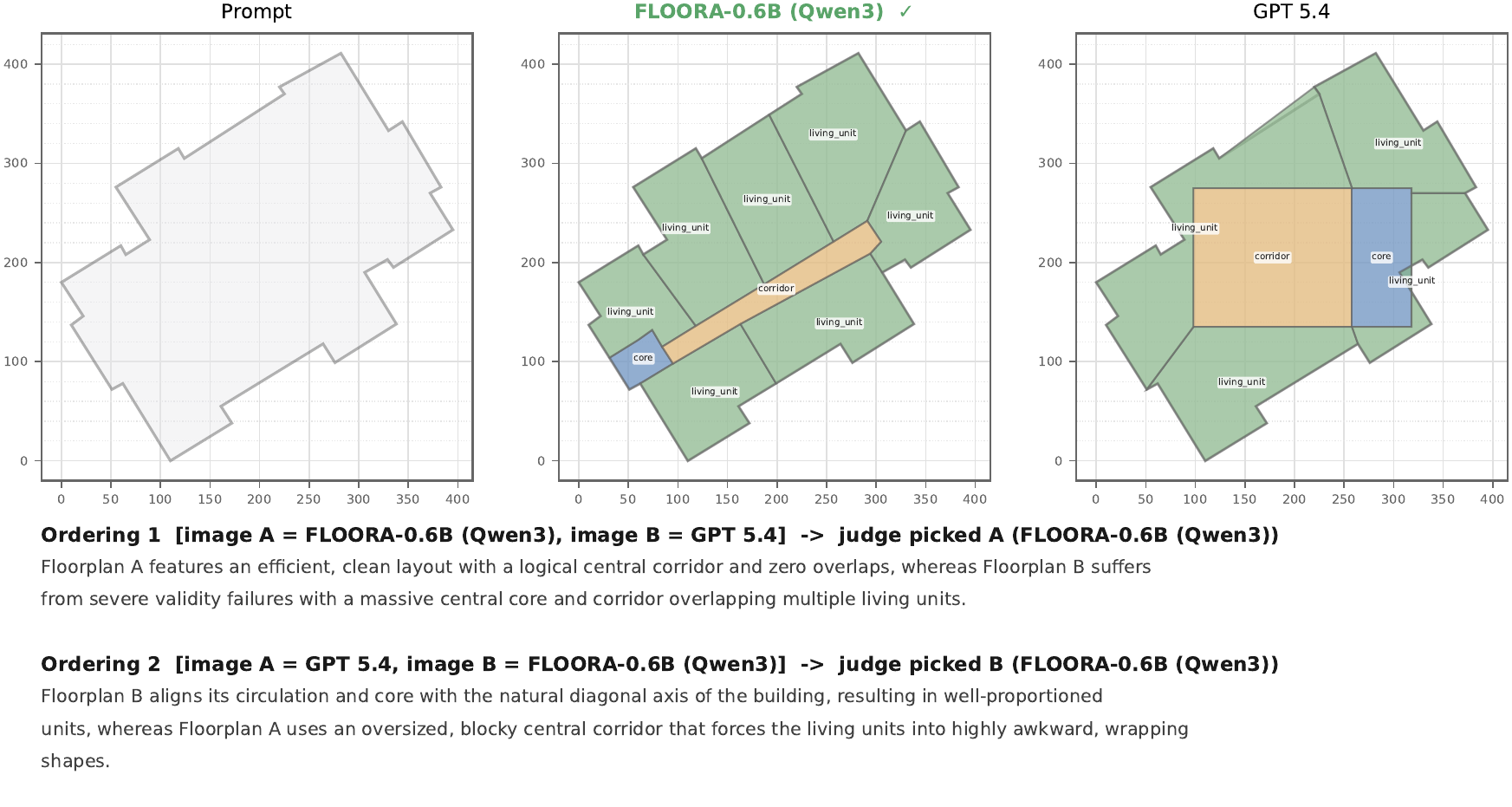}
        \caption{Example VLM judge reasoning for a comparison in which the FLOORA model is preferred.}
    \end{subfigure}
    \vspace{2em}

    \begin{subfigure}[b]{0.75\textwidth}
        \centering
        \includegraphics[width=\textwidth]{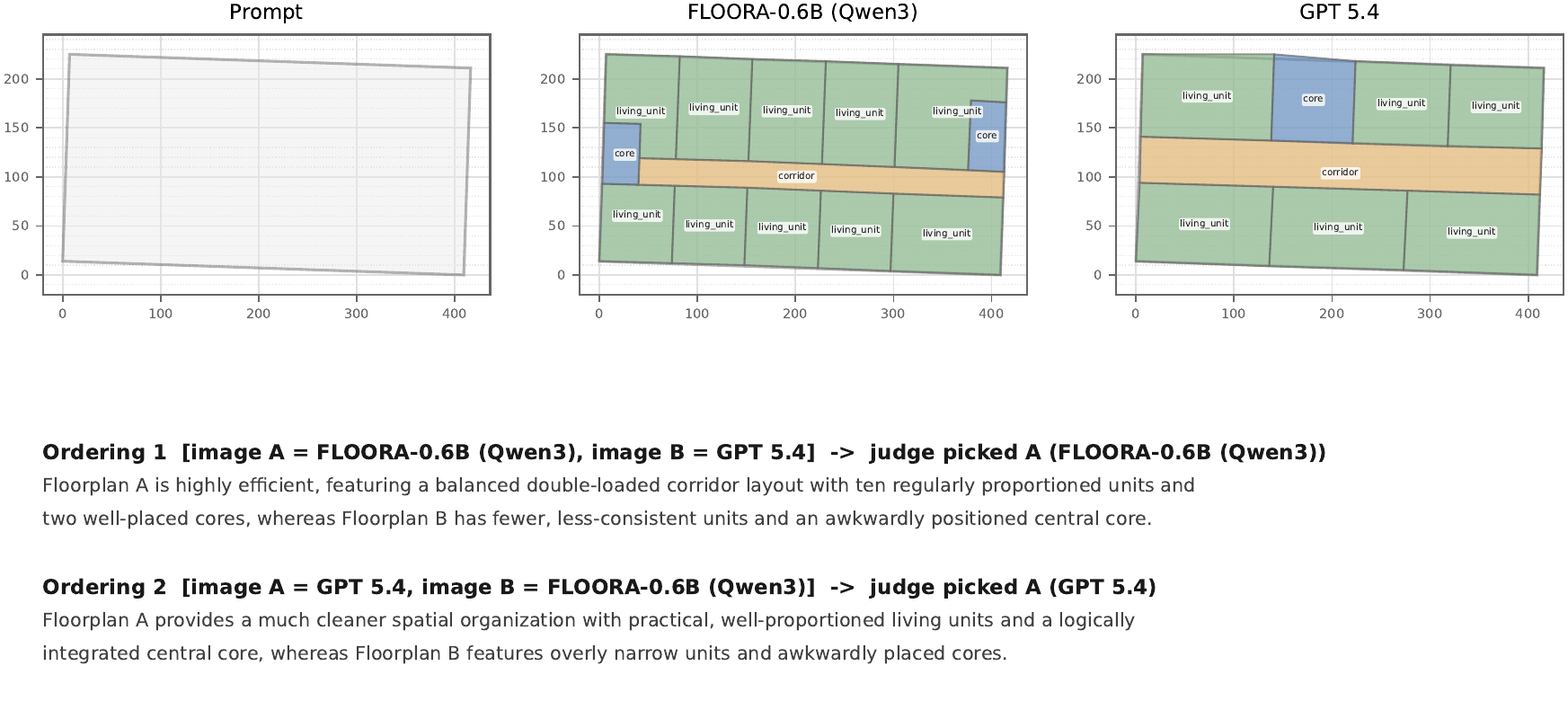}
        \caption{Example VLM judge reasoning for a comparison judged as a tie.}
    \end{subfigure}
    \vspace{2em}

    \begin{subfigure}[b]{0.75\textwidth}
        \centering
        \includegraphics[width=\textwidth]{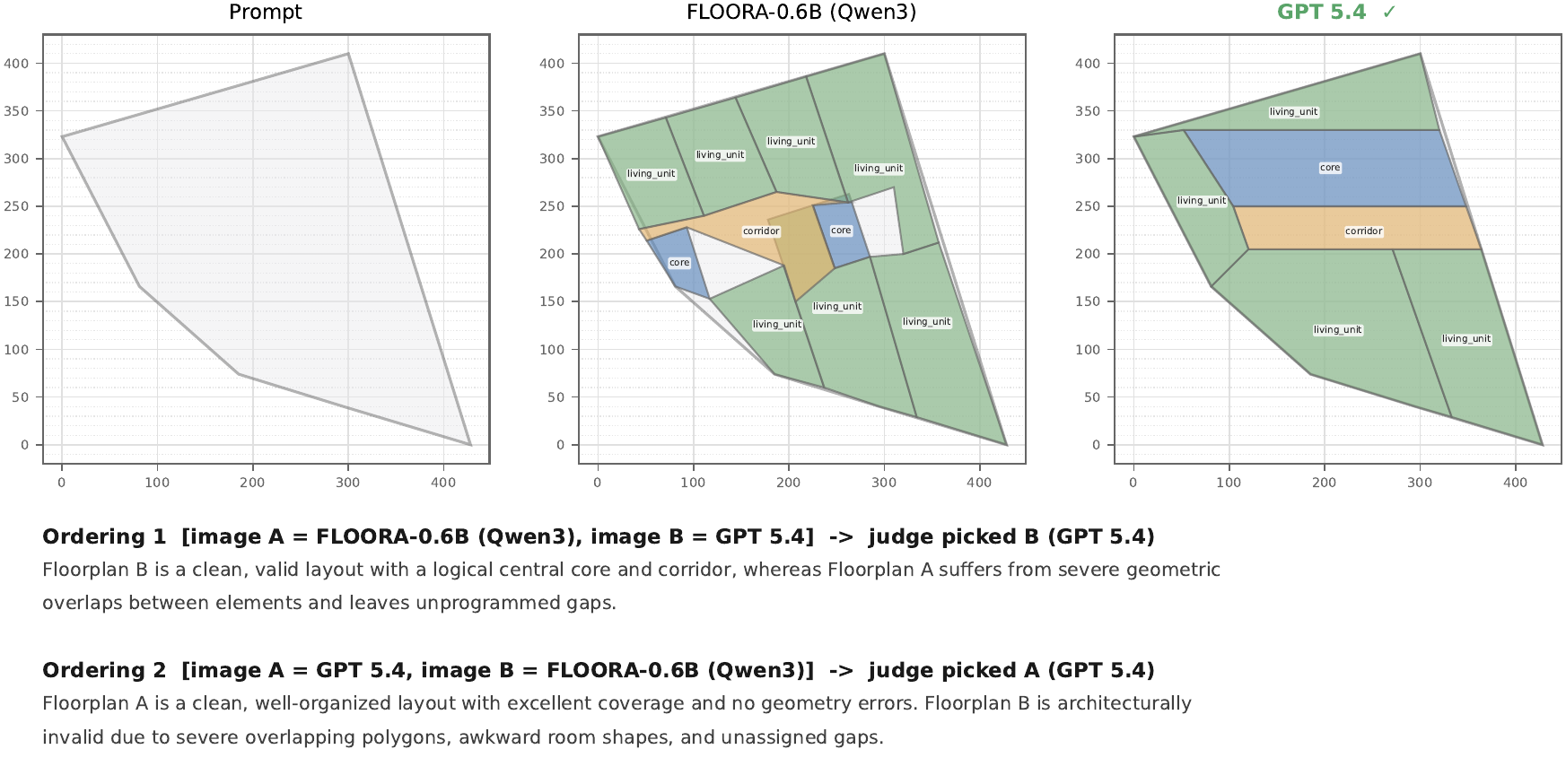}
        \caption{Example VLM judge reasoning for a comparison in which the frontier model is preferred.}
    \end{subfigure}

    \caption{Example VLM judge reasoning for representative pairwise outcomes.}
    
    \label{fig:judge_examples}
\end{figure}

\cref{fig:judge_examples} shows representative VLM judge reasoning for a FLOORA win, a tie, and a baseline win. The examples show that decisions reflect architectural criteria, including spatial organization, circulation, coverage, proportions, and validity, rather than visual appearance alone. The tie case also demonstrates how comparable strengths and weaknesses are handled without forcing a preference.

%% file: appendix/frontier_model_inference.tex
To establish a strong external baseline, we evaluate a set of frontier LLMs on the same floor-plan generation task used for our domain-specific models. Unlike the models trained in this work, these frontier models are accessed only through inference APIs and are not fine-tuned on the DSL or on the architectural feedback dataset. This evaluation therefore measures the few-shot ability of general-purpose frontier models to produce syntactically valid and architecturally plausible floor-plan layouts from structured DSL context.

All frontier models are evaluated on the same prompt distributions used in the main text, including synthetic test prompts and OpenStreetMap-derived real-world massing prompts. Their completions are scored with the same verifiable reward functions described in \cref{sec:verifiable_rewards}, including geometric correctness, functional compliance, and aggregate total reward. As in the main evaluation, a completion passes the total reward criterion only when all required checks pass. This shared protocol enables direct comparison between frontier model completions and completions from our trained DSL models. We also compute the win-rate of our models against frontier baselines using a VLM judge.

\subsection{Models Evaluated}
We evaluate Claude Sonnet 4.6, Claude Opus 4.8, OpenAI GPT-5.4, Gemini 2.5 Flash Image, and Gemini 3.5 Flash through inference APIs, all with a maximum of 8,192 tokens and provider-default temperature. All models receive identical prompts with one illustrative example of the expected DSL syntax.

\subsection{Prompt Format}
Each query consists of a system message that defines the floor-plan generation task and a user message containing the full DSL context for the evaluated sample. The user message includes all available input modalities, such as the \texttt{building}, \texttt{structure}, \texttt{massing} blocks. The model is instructed to return only a complete \texttt{spaces} block delimited by sentinel tokens.

\begin{tcolorbox}[
    title=System Prompt for Frontier Model Baselines,
    colback=gray!5,
    colframe=black,
    fonttitle=\bfseries,
    breakable
]
\small
\begin{verbatim}
You are an expert architectural design assistant generating floor plan layouts.

You will receive a building description in DSL format. The DSL uses named
modality blocks:
- `building { ... }`: building metadata, including occupancy type and floor levels
- `structure { ... }`: structural grid and material information
- `massing { polygon <type> x,y x,y ... }`: the building massing footprint polygon
- `spaces { polygon <type> x,y x,y ... }`: individual spaces, when partially provided

All coordinates are quantized integers in the range 0 to 1023. Each
`polygon <type>` is followed by space-separated `x,y` coordinate pairs defining
a closed polygon. Do NOT repeat the first vertex at the end.

Space types: `core`, `corridor`, `living_unit`.

Cores are spaces intended to contain stairs, elevators, or other building
services. Corridors are spaces intended for circulation. Living units are spaces
intended for residential use.

Your task is to generate a complete `spaces` layout for the given massing.
Produce a DSL `spaces { ... }` block with exactly the following syntax:

spaces {
  polygon <type> x,y x,y x,y ...
  polygon <type> x,y x,y x,y ...
  ...
}

Rules:
- All coordinate values must be integers in the range 0 to 1023.
- Each polygon must have at least 3 vertices.
- Do NOT repeat the first vertex at the end.
- Spaces must tile the massing polygon completely. Every point inside the
  massing must belong to exactly one space.
- There must be no uncovered area within the massing boundary.
- Every `living_unit` must share at least one edge with the exterior facade,
  defined by the massing boundary.
- Every `living_unit` must share at least one edge with a `corridor`.
- Space sizes must be architecturally plausible. For example, each core must
  be large enough to realistically contain stairs, elevators, or building
  services.

Example

Input:
building { occupancy_type multifamily_residential storeys 4 level 1 elevation 0 }
structure { material reinforced_concrete }
massing { polygon mass 0,0 400,0 400,200 0,200 }

Output:
STARTING_GENERATION
spaces {
  polygon living_unit 0,0 160,0 160,80 0,80
  polygon core 160,0 240,0 240,80 160,80
  polygon living_unit 240,0 400,0 400,80 240,80
  polygon corridor 0,80 400,80 400,120 0,120
  polygon living_unit 0,120 200,120 200,200 0,200
  polygon living_unit 200,120 400,120 400,200 200,200
}
END_GENERATION

Output your response in this exact format:
STARTING_GENERATION
spaces {
  ...
}
END_GENERATION

No markdown, no explanations, and no extra text outside the sentinels.
\end{verbatim}
\end{tcolorbox}

\subsection{Output Parsing}

Frontier models are instructed to delimit their generations using the sentinel tokens \texttt{STARTING\_GENERATION} and \texttt{END\_GENERATION}. When both sentinels are present, we extract the text between them and strip surrounding whitespace. When one or both sentinels are absent, which can occur when a model emits preamble text or otherwise deviates from the requested format, the full model response is retained and passed to the parser.

The extracted text is validated using the same DSL parser used during data preprocessing and evaluation, described in \cref{app:dsl}. This parser checks whether the generated text can be interpreted as a valid \texttt{spaces \{ ... \}} block and converted into the internal geometric representation. Responses that fail parsing are not discarded. Instead, they are passed to the reward functions, which assign zero reward to syntactically invalid DSL. This preserves comparability with trained model completions, where invalid generations are evaluated under the same scoring pipeline rather than filtered out before evaluation.

%% file: appendix/results.tex
\subsection{Pre-Training Results}
\label[appendix]{app:results_pt}

\begin{figure}[h!]
    \centering
    \begin{subfigure}[b]{0.48\textwidth}
        \centering
        \includegraphics[width=\textwidth]{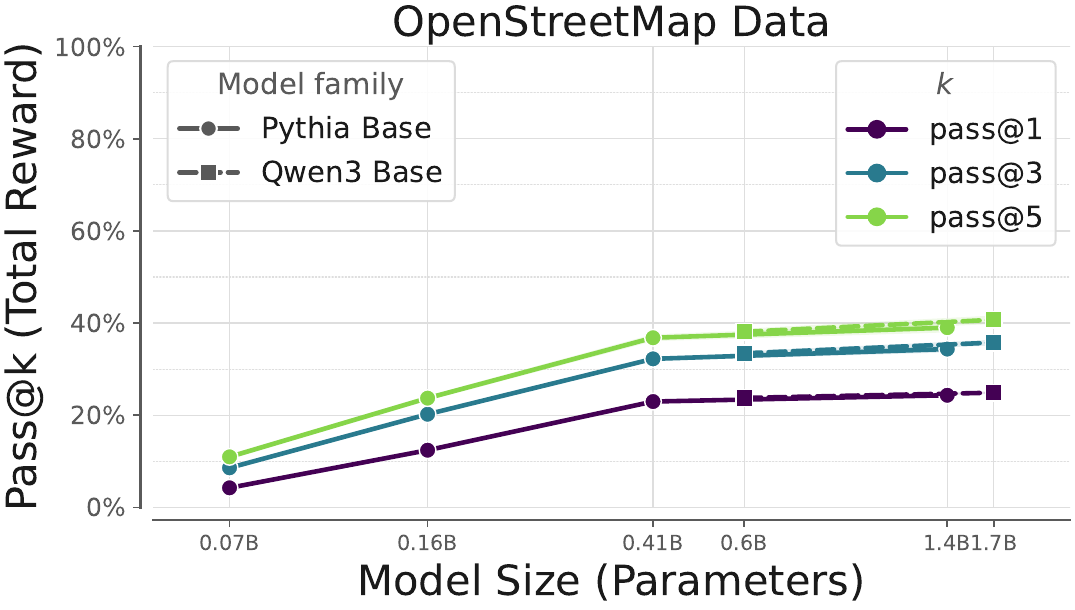}
        \caption{Pass@k on the OSM test set.}
    \end{subfigure}
    \hfill
    \begin{subfigure}[b]{0.48\textwidth}
        \centering
        \includegraphics[width=\textwidth]{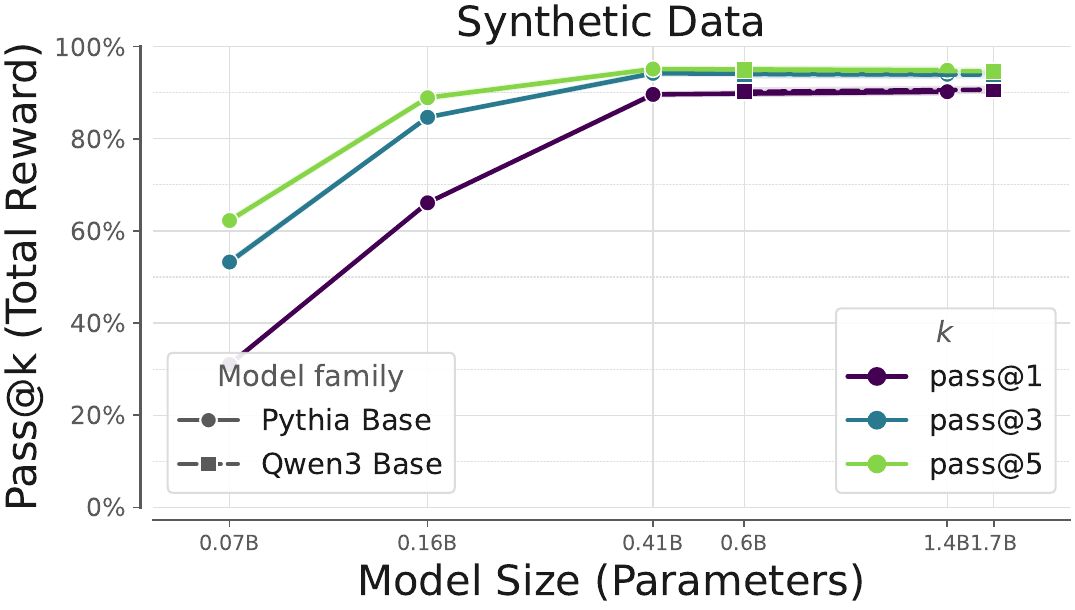}
        \caption{Pass@k on the synthetic test set.}
    \end{subfigure}
    \caption{Pass@1/3/5 performance of the base models on OSM and synthetic test sets. Pass@k is considered achieved only when both the geometric correctness and functional compliance checks pass. Shaded regions represent 95\% CIs across 5 evaluation seeds.}
    \label{fig:results_pt_eval}
\end{figure}

\cref{fig:results_pt_eval} and \cref{tab:pt_osm,tab:pt_synthetic} present pass@1/3/5 evaluation results for the base models across different model scales on the synthetic and OpenStreetMap (OSM) test sets. Pass@k for the total reward is considered achieved only when both the geometric correctness and functional compliance checks pass. Models with fewer than 0.4B parameters struggle to learn the underlying building patterns, resulting in poor performance even on synthetic data. While larger models largely saturate on the synthetic benchmark, their generalization continues to improve on OSM, which represents a held-out evaluation setting. 

\begin{figure}[b!]
    \centering
    \begin{subfigure}[b]{0.48\textwidth}
        \centering
        \includegraphics[width=\textwidth]{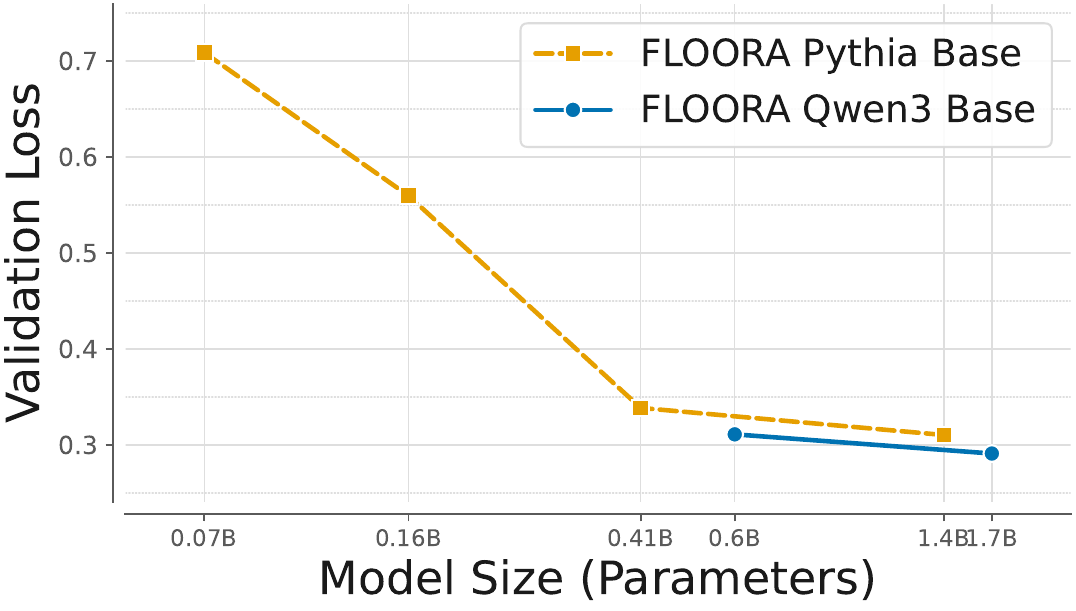}
        \caption{Final validation loss.}
    \end{subfigure}
    \hfill
    \begin{subfigure}[b]{0.48\textwidth}
        \centering
        \includegraphics[width=\textwidth]{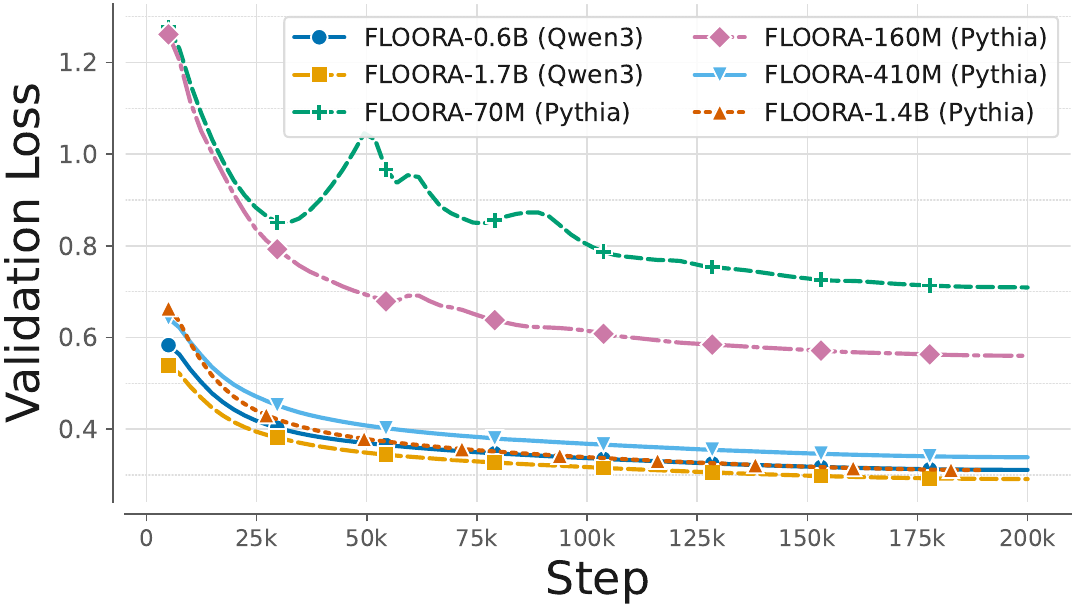}
        \caption{Validation loss curves.}
    \end{subfigure}
    \caption{Pre-training validation across model scales. \textbf{(a)} Final validation loss achieved by each model. \textbf{(b)} Validation loss throughout pre-training.}
    \label{fig:results_pt_loss}
\end{figure}

\cref{fig:results_pt_loss} shows pre-training validation loss, measured as next-token cross-entropy, across model scales throughout training. Consistent with the downstream results, models below 0.4B parameters struggle to capture the underlying building patterns, obtaining higher validation cross-entropy at the end of the pre-training stage.


\subsection{Supervised Fine-Tuning}
\label[appendix]{app:results_sft}

\cref{fig:results_main_app} compares the pass@1/3/5 results for the base and SFT checkpoints across model families and scales. SFT improves performance on the OSM test set across the evaluated settings, indicating that architect-edited completions provide an effective alignment signal for real-world building footprints. Since OSM is out-of-distribution relative to the synthetic pre-training data and SFT data (see \cref{fig:tsne_data}), these gains suggest improved transfer beyond the procedural data distribution.

On the synthetic test set, we observe a modest decrease in pass@k performance after SFT. We attribute this to the distribution shift between the procedural synthetic layouts used during pre-training and the architect-corrected completions used for SFT. In this sense, SFT shifts the model away from reproducing the synthetic generator distribution and toward layouts that better reflect expert architectural judgment.

\begin{figure}[t!]
    \centering
    \begin{subfigure}[b]{0.48\textwidth}
        \centering
        \includegraphics[width=\textwidth]{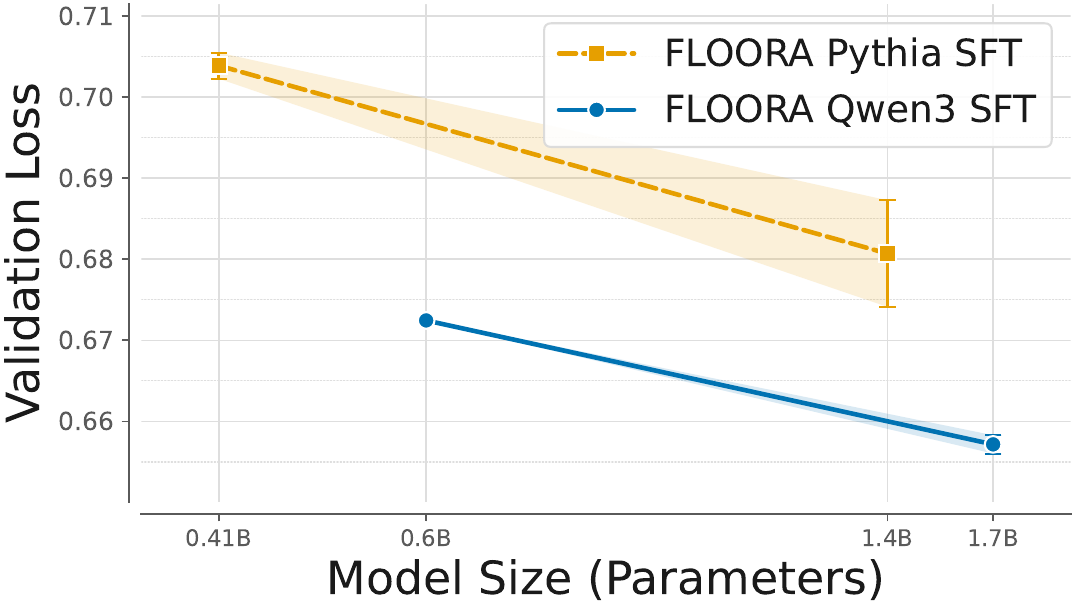}
        \caption{Final validation loss.}
    \end{subfigure}
    \hfill
    \begin{subfigure}[b]{0.48\textwidth}
        \centering
        \includegraphics[width=\textwidth]{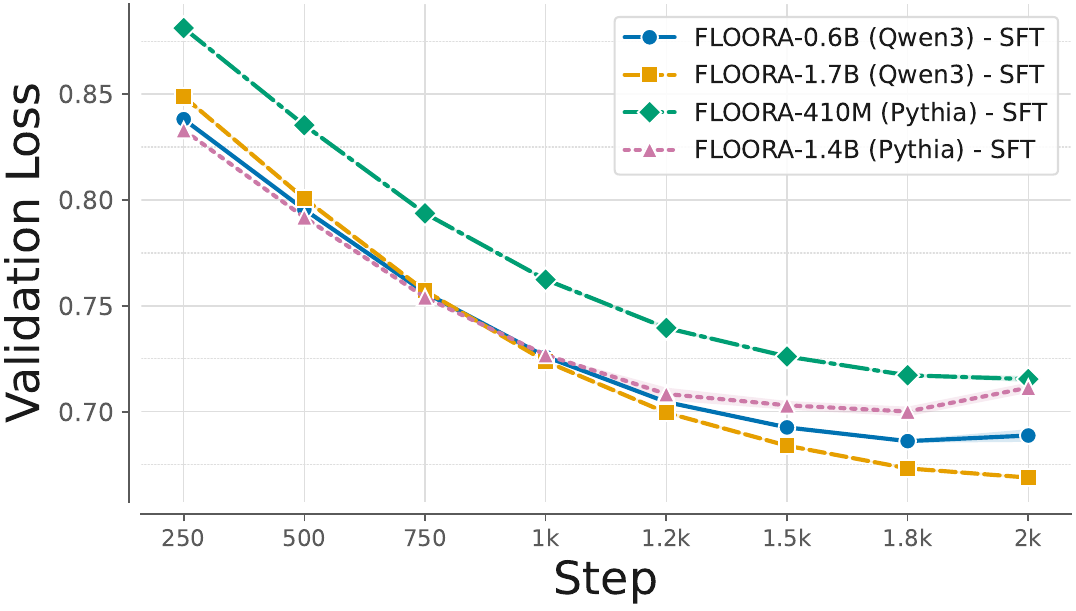}
        \caption{Validation loss curves.}
    \end{subfigure}
    \caption{SFT validation loss across model scales. \textbf{(a)} Final validation loss achieved by each model. \textbf{(b)} Validation loss throughout training. Shaded regions represent 95\% CIs across 3 training seeds.}
    \label{fig:sft_results}
\end{figure}

\cref{fig:sft_results} reports SFT validation cross-entropy. These metrics provide a training diagnostic for the supervised objective, but they should be interpreted alongside the pass@k results rather than as direct measures of architectural quality. Overall, the results suggest that SFT functions primarily as a domain-alignment stage, trading some in-distribution synthetic performance for improved generalization to real-world footprints.

\subsection{Reward Model Training}
\begin{figure}[h!]
    \centering
    \begin{subfigure}[b]{0.48\textwidth}
        \centering
        \includegraphics[width=\textwidth]{figures/results_rm/best_eval_accuracy_vs_scale.pdf}
        \caption{Final validation accuracy.}
    \end{subfigure}
    \hfill
    \begin{subfigure}[b]{0.48\textwidth}
        \centering
        \includegraphics[width=\textwidth]{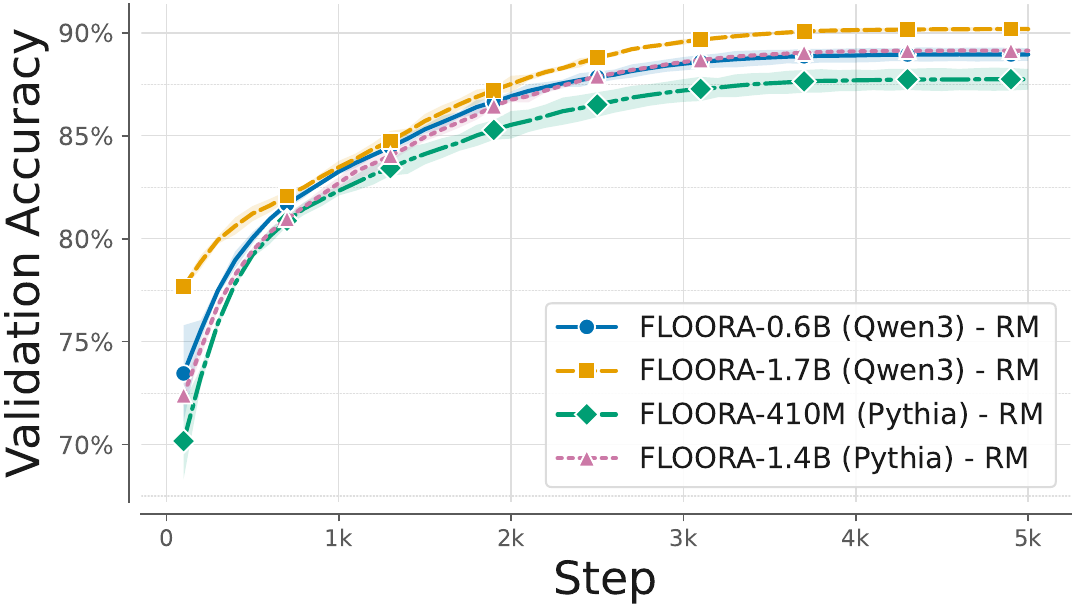}
        \caption{Validation accuracy curves.}
    \end{subfigure}
    \caption{Reward model validation accuracy across model scales. \textbf{(a)} Final validation accuracy achieved by each model. \textbf{(b)} Validation accuracy throughout reward model training. Shaded regions represent 95\% CIs across 3 training seeds.}
    \label{fig:rm_results_accuracy}
\end{figure}

\label[appendix]{app:results_rm}
\cref{fig:rm_results_accuracy} presents reward model validation accuracy. Accuracy improves rapidly during the early stages of optimization and then plateaus, indicating that the models learn most of the pairwise preference signal within the first few thousand training steps.

As expected, final validation accuracy increases with model size, suggesting that larger reward models better capture the architectural preferences expressed in the feedback data.

\clearpage
\subsection{Reinforcement Learning}
\label[appendix]{app:results_rl}

\begin{figure}[h!]
    \centering
    \begin{subfigure}[b]{0.48\textwidth}
        \centering
        \includegraphics[width=\textwidth]{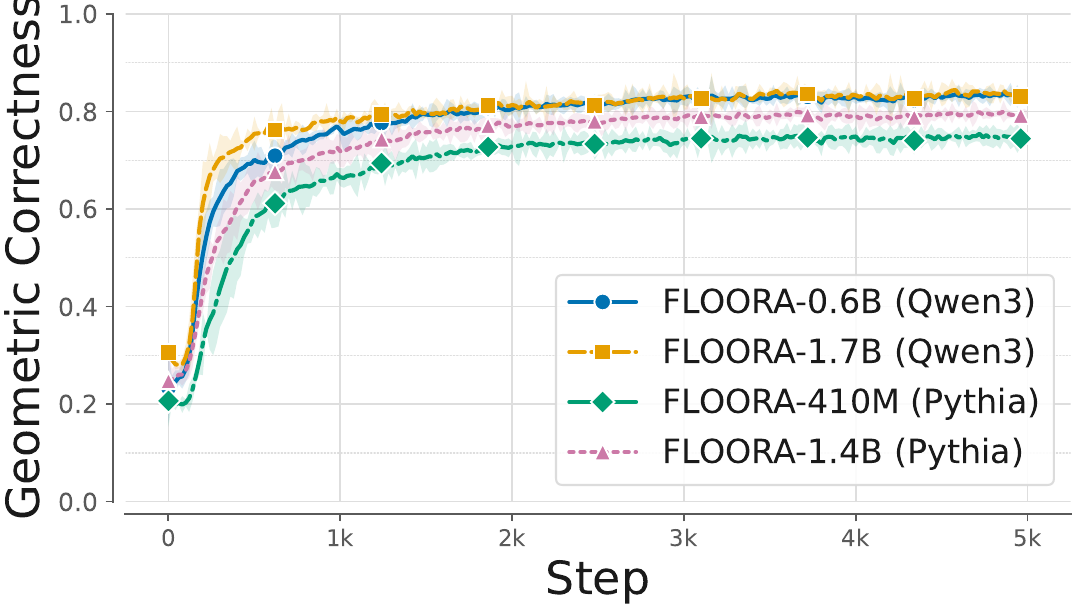}
        \caption{Geometric correctness reward curves.}
    \end{subfigure}
    \hfill
    \begin{subfigure}[b]{0.48\textwidth}
        \centering
        \includegraphics[width=\textwidth]{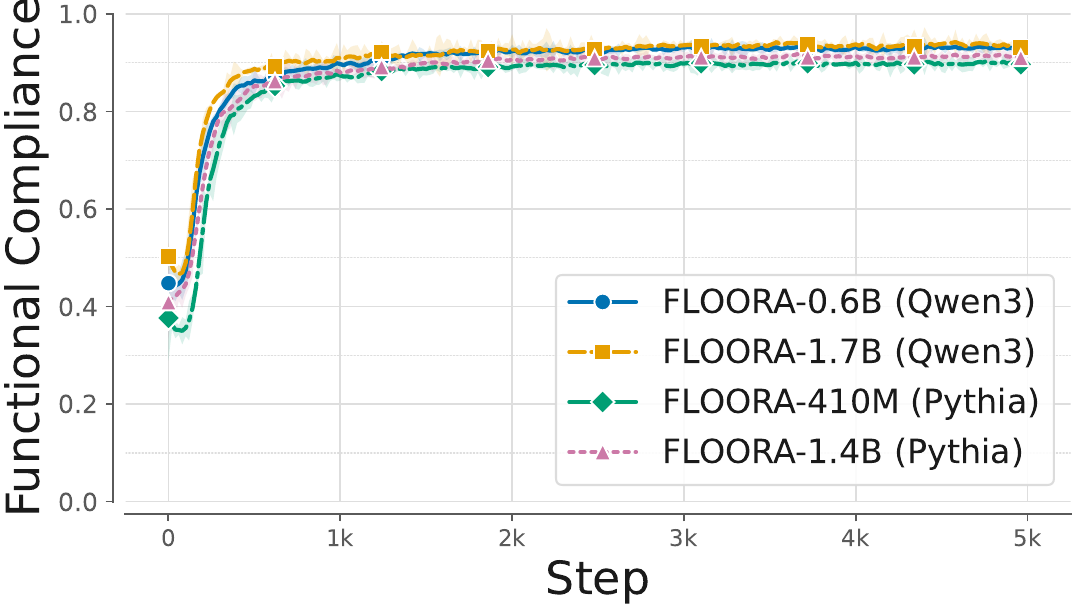}
        \caption{Functional compliance reward curves.}
    \end{subfigure}
    \vspace{1em}

    \centering
    \begin{subfigure}[b]{0.48\textwidth}
        \centering
        \includegraphics[width=\textwidth]{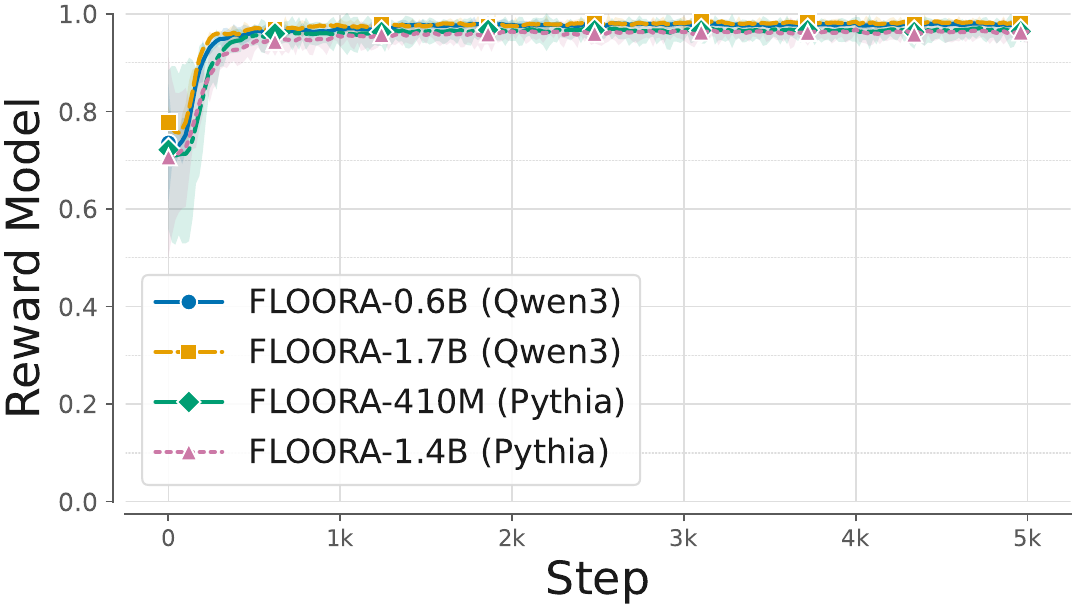}
        \caption{Reward model curves.}
    \end{subfigure}
    \centering
    \begin{subfigure}[b]{0.48\textwidth}
        \centering
        \includegraphics[width=\textwidth]{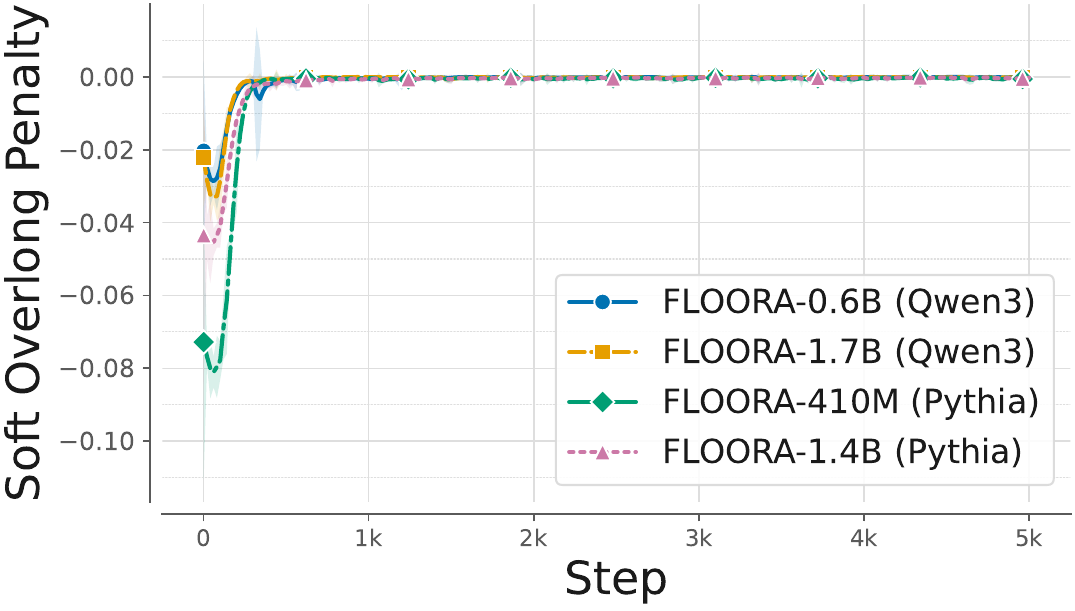}
        \caption{Soft overlong penalty curves.}
    \end{subfigure}
    \caption{Learning curves for GRPO reward components across model scales, including geometric correctness, functional compliance, the normalized reward model score, and the soft overlong penalty. Shaded regions show 95\% CIs across 3 training seeds.}
    \label{fig:rlhf_learning_curves}
\end{figure}

\cref{fig:rlhf_learning_curves} shows the GRPO learning curves for the reward components. Across model scales, training is stable; geometric correctness, functional compliance, and the reward-model score improve quickly and then plateau, while the soft overlong penalty remains near zero.

\begin{figure}[t!]
    \centering
    \begin{subfigure}[b]{0.48\textwidth}
        \centering
        \includegraphics[width=\textwidth]{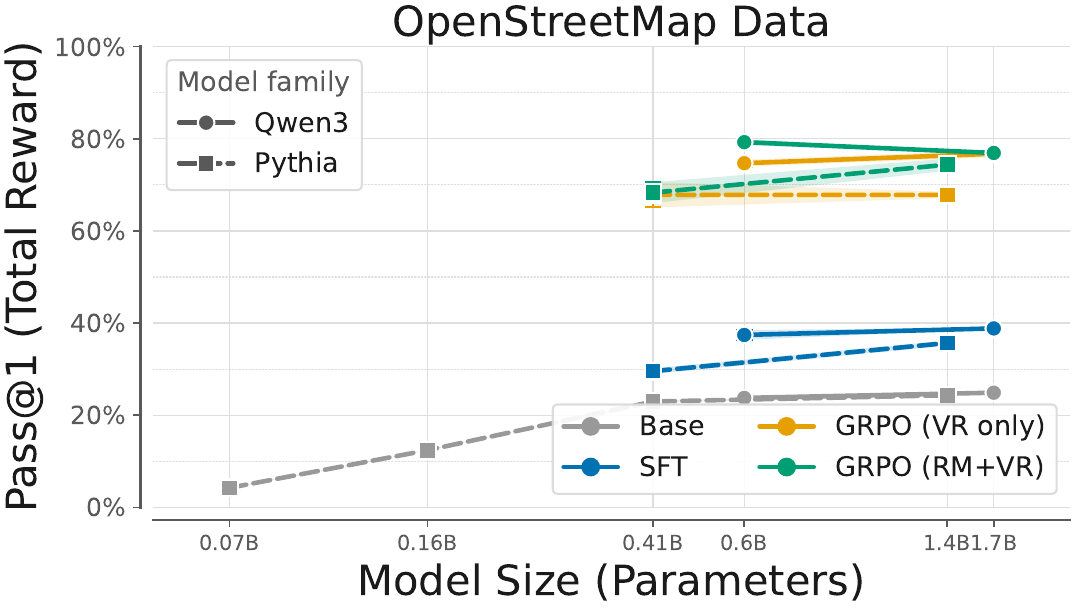}
        \caption{Pass@1 on the OSM test set.}
    \end{subfigure}
    \hfill
    \begin{subfigure}[b]{0.48\textwidth}
        \centering
        \includegraphics[width=\textwidth]{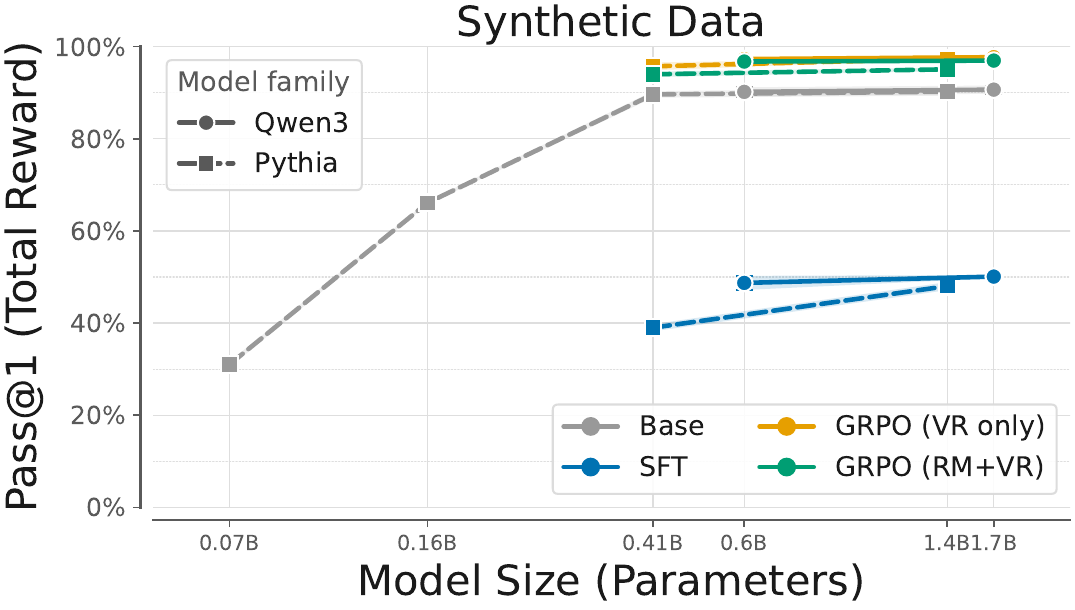}
        \caption{Pass@1 on the synthetic test set.}
    \end{subfigure}
    
    \vspace{2em}
    \centering
    \begin{subfigure}[b]{0.48\textwidth}
        \centering
        \includegraphics[width=\textwidth]{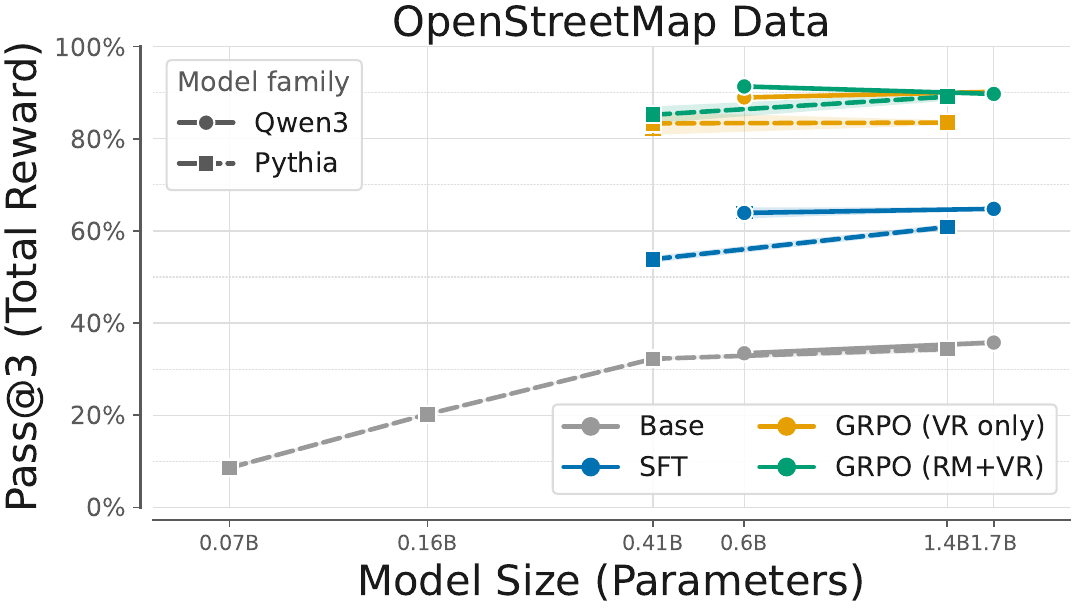}
        \caption{Pass@3 on the OSM test set.}
    \end{subfigure}
    \hfill
    \begin{subfigure}[b]{0.48\textwidth}
        \centering
        \includegraphics[width=\textwidth]{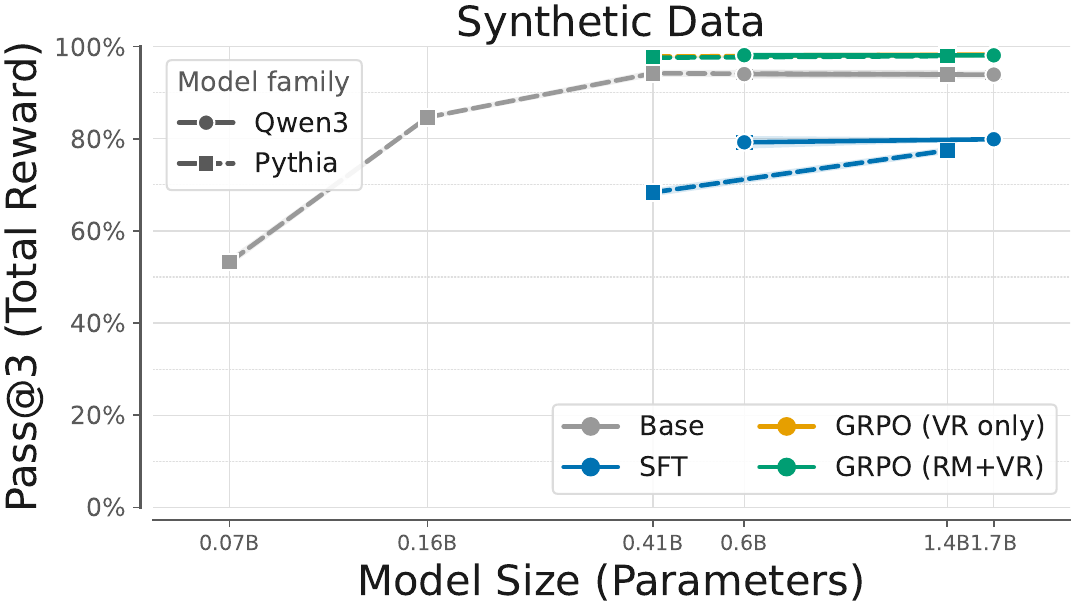}
        \caption{Pass@3 on the synthetic test set.}
    \end{subfigure}
    
    \vspace{2em}
    \centering
    \begin{subfigure}[b]{0.48\textwidth}
        \centering
        \includegraphics[width=\textwidth]{figures/results_main/osm_pass5_vs_scale_total_reward.pdf}
        \caption{Pass@5 on the OSM test set.}
    \end{subfigure}
    \hfill
    \begin{subfigure}[b]{0.48\textwidth}
        \centering
        \includegraphics[width=\textwidth]{figures/results_main/synthetic_pass5_vs_scale_total_reward.pdf}
        \caption{Pass@5 on the synthetic test set.}
    \end{subfigure}
    
    \caption{Pass@1/3/5 performance of the base and fine-tuned models on OSM and synthetic test sets. Pass@k is considered achieved only when both the geometric correctness and functional compliance checks pass. Shaded regions represent 95\% CIs across 3 training and 5 evaluation seeds.}
    \label{fig:results_main_app}
\end{figure}

The main post-training results are shown in \cref{fig:results_main_app} and \cref{tab:rl_osm,tab:rl_synthetic}. Both GRPO variants substantially improve pass@1/3/5 on OSM and synthetic data, indicating that RL fixes many of the validity and constraint-satisfaction issues left by the base and SFT models.
GRPO produces consistent positive improvements across datasets and model scales. The only exception is SFT on the synthetic test set, where performance drops due to the distribution shift discussed in \cref{app:results_sft}. Overall, GRPO recovers this loss while preserving the OSM gains from the SFT stage.

\begin{figure}[b!]
    \centering
    \begin{subfigure}[b]{0.48\textwidth}
        \centering
        \includegraphics[width=\textwidth]{figures/results_rlhf/osm_rlhf_stacked_k1.pdf}
        \caption{Win rates on the OSM test set.}
    \end{subfigure}
    \hfill
    \begin{subfigure}[b]{0.48\textwidth}
        \centering
        \includegraphics[width=\textwidth]{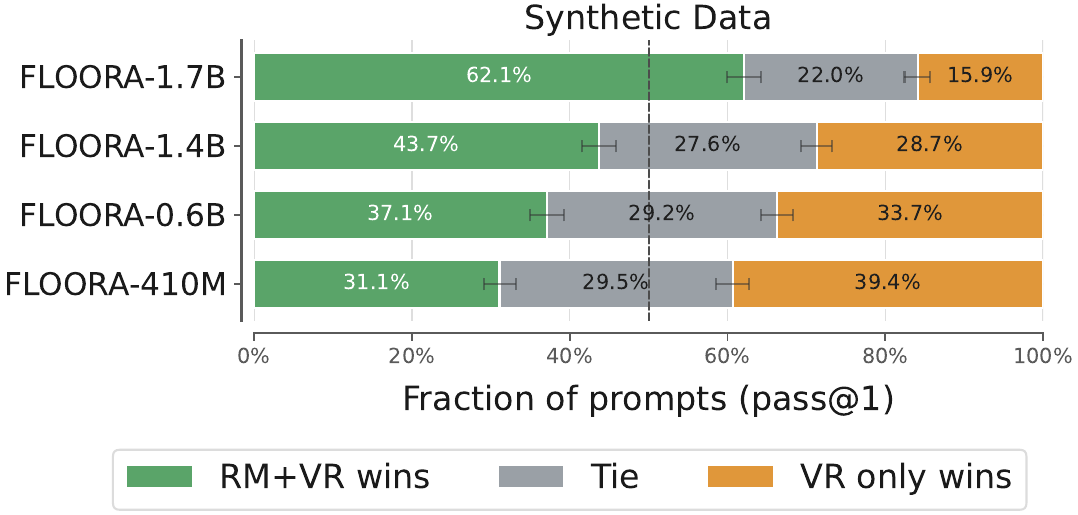}
        \caption{Win rates on the synthetic test set.}
    \end{subfigure}
    \caption{Pairwise VLM judge outcomes for each FLOORA GRPO (RM+VR) model against its matched GRPO (VR only) counterpart. Bars show the fraction of prompts where RM+VR wins, ties, or loses. As model size increases, the RM+VR variant is increasingly preferred. Error bars indicate 95\% Wilson intervals over the evaluation prompts.}
    \label{fig:winrates_rlhf_app}
\end{figure}

Finally, we evaluate the effect of the learned reward model during post-training by comparing models trained with GRPO (RM+VR) against matched GRPO (VR only) models using the VLM judge described in \cref{app:vlm_judge}. \cref{fig:winrates_rlhf_app} reports pairwise outcomes supplementing \cref{fig:winrates_rlhf}, where a win indicates that the RM+VR model is preferred over its VR only counterpart. The results show that the benefit of including the reward model grows with model scale. In particular, RM+VR underperforms VR only for the smallest model, FLOORA-410M (Pythia), but becomes increasingly preferred as model capacity increases, reaching the strongest margins for FLOORA-1.7B (Qwen3).

This trend suggests that the learned reward model can improve qualitative architectural preferences beyond VR-only training when both the policy and reward model have sufficient capacity to capture and exploit architectural-quality judgments, but that its benefits are capacity-dependent and may be limited or even harmful for smaller models.

\begin{landscape}
\input{content/table_pt_results}
\input{content/table_rl_results}

\begin{table}[h!]
\centering
\footnotesize
\caption{Pass@1/3/5 comparison of the FLOORA-0.6B (Qwen3) -- GRPO (RM+VR) and frontier baselines on the \textbf{OSM} test set. Best absolute value is shown in bold. Results are obtained on a single seed.}
\label{tab:frontier_osm}
\begin{tabular}{l ccc ccc ccc}
\toprule
 & \multicolumn{3}{c}{\textbf{Functional Compliance}} & \multicolumn{3}{c}{\textbf{Geometric Correctness}} & \multicolumn{3}{c}{\textbf{Total Reward}} \\
\cmidrule(lr){2-4}\cmidrule(lr){5-7}\cmidrule(lr){8-10}
\textbf{Model} & @1 & @3 & @5 & @1 & @3 & @5 & @1 & @3 & @5 \\
\midrule
FLOORA-0.6B (Qwen3) -- GRPO (RM+VR) & \textbf{84.8} & \textbf{95.3} & \textbf{97.2} & \textbf{79.5} & \textbf{91.6} & \textbf{94.0} & \textbf{78.1} & \textbf{90.8} & \textbf{93.5} \\
\midrule
Claude Opus 4.8 & 46.1 & 73.6 & 83.0 & 36.4 & 60.8 & 72.0 & 24.7 & 45.0 & 55.6 \\
Claude Sonnet 4.6 & 19.3 & 38.4 & 49.4 & 10.0 & 20.2 & 26.8 & 5.1 & 10.7 & 14.5 \\
Gemini 2.5 Flash Image & 37.6 & 62.0 & 71.7 & 25.5 & 44.3 & 53.8 & 11.1 & 20.3 & 26.0 \\
Gemini 3.5 Flash & {52.6} & {76.6} & {83.9} & {42.0} & {69.1} & {79.7} & {33.3} & {57.4} & {68.2} \\
OpenAI GPT 5.4 & 46.6 & 71.6 & 79.7 & 30.3 & 50.4 & 60.4 & 19.9 & 36.0 & 45.0 \\
\bottomrule
\end{tabular}
\end{table}

\begin{table}[h!]
\centering
\footnotesize
\caption{Pass@1/3/5 comparison of the FLOORA-0.6B (Qwen3) -- GRPO (RM+VR) model and frontier baselines on the \textbf{synthetic} test set. Best absolute value is shown in bold. Results are obtained on a single seed.}
\label{tab:frontier_synthetic}
\begin{tabular}{l ccc ccc ccc}
\toprule
& \multicolumn{3}{c}{\textbf{Functional Compliance}} & \multicolumn{3}{c}{\textbf{Geometric Correctness}} & \multicolumn{3}{c}{\textbf{Total Reward}} \\
\cmidrule(lr){2-4}\cmidrule(lr){5-7}\cmidrule(lr){8-10}
\textbf{Model} & @1 & @3 & @5 & @1 & @3 & @5 & @1 & @3 & @5 \\
\midrule
FLOORA-0.6B (Qwen3) -- GRPO (RM+VR) & \textbf{97.4} & \textbf{98.3} & \textbf{98.4} & \textbf{96.5} & \textbf{97.8} & \textbf{98.0} & \textbf{96.5} & \textbf{97.7} & \textbf{97.9} \\
\midrule
Claude Opus 4.8 & 66.5 & 92.2 & 97.1 & 38.6 & 62.9 & 72.9 & 30.9 & 54.2 & 64.5 \\
Claude Sonnet 4.6 & 39.0 & 70.2 & 82.5 & 8.1 & 15.5 & 20.4 & 6.9 & 13.5 & 17.9 \\
Gemini 2.5 Flash Image & 64.1 & 90.7 & 96.7 & 15.6 & 29.7 & 38.0 & 12.9 & 24.0 & 30.6 \\
Gemini 3.5 Flash & {75.3} & {96.4} & {99.2} & {39.9} & {66.2} & {77.0} & {39.1} & {65.2} & {76.0} \\
OpenAI GPT 5.4 & 69.8 & 93.8 & 98.1 & 27.4 & 44.1 & 52.2 & 23.9 & 40.5 & 48.5 \\
\bottomrule
\end{tabular}
\end{table}

\end{landscape}

\clearpage
\subsection{Comparison with Frontier Models}
\begin{figure}[b!]
    \centering
    \begin{subfigure}[b]{0.45\textwidth}
        \centering
        \includegraphics[width=\textwidth]{figures/results_frontier/osm_passk_curves_total_reward.pdf}
        \caption{Pass@k on the OSM test set.}
    \end{subfigure}
    \hfill
    \begin{subfigure}[b]{0.45\textwidth}
        \centering
        \includegraphics[width=\textwidth]{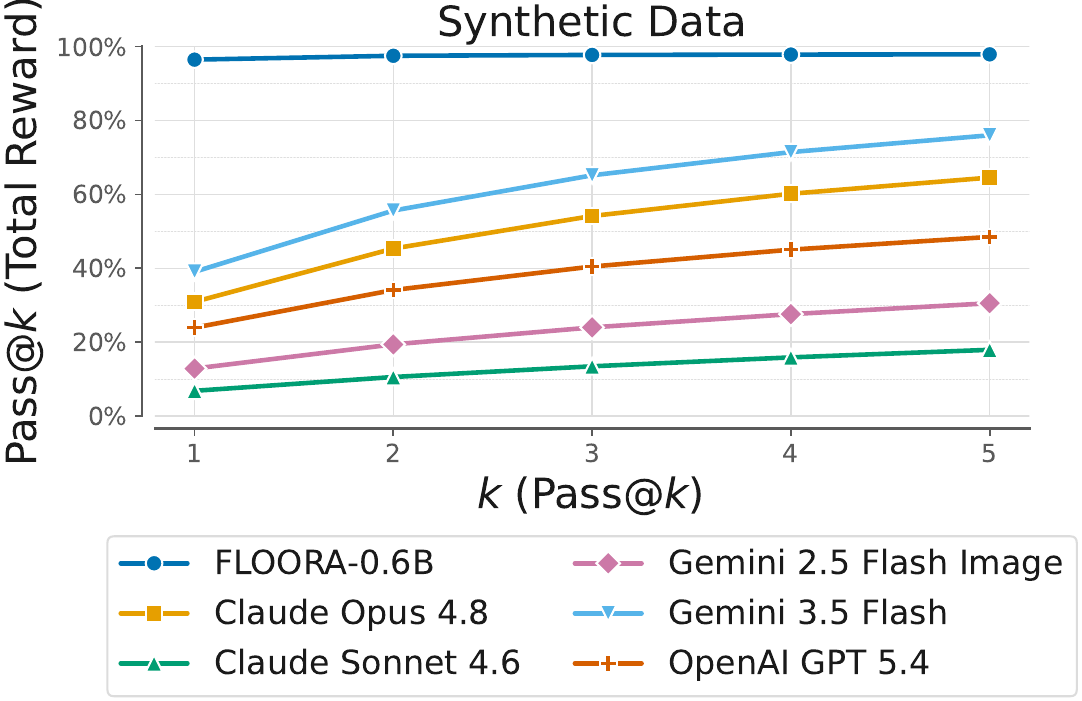}
        \caption{Pass@k on the synthetic test set.}
    \end{subfigure}

    \caption{Pass@k comparison of the FLOORA-0.6B (Qwen3) GRPO (RM+VR) model and frontier baselines, where success requires passing both geometric and functional checks. Our model substantially outperforms all evaluated frontier baselines.}
    \label{fig:passk_frontier_app}
\end{figure}

\label[appendix]{app:results_frontier}
\paragraph{Verifier-Based Evaluations.} 
We provide additional results for the comparison between our FLOORA-0.6B (Qwen3) GRPO (RM+VR) model and the frontier-model baselines evaluated under the inference protocol in \cref{app:frontier_model_inference}. \cref{fig:passk_frontier_app} and \cref{tab:frontier_osm,tab:frontier_synthetic} supplement \cref{fig:passk_frontier} by reporting the corresponding pass@1, pass@3, and pass@5 results on the OSM and synthetic test sets. Across both benchmarks, our model achieves the highest total reward pass@k values.

\begin{figure}[b!]
    \centering
    \begin{subfigure}[b]{0.45\textwidth}
        \centering
        \includegraphics[width=\textwidth]{figures/results_frontier/osm_winrate_stacked_k1.pdf}
        \caption{Win rates on the OSM test set.}
    \end{subfigure}
    \hfill
    \begin{subfigure}[b]{0.45\textwidth}
        \centering
        \includegraphics[width=\textwidth]{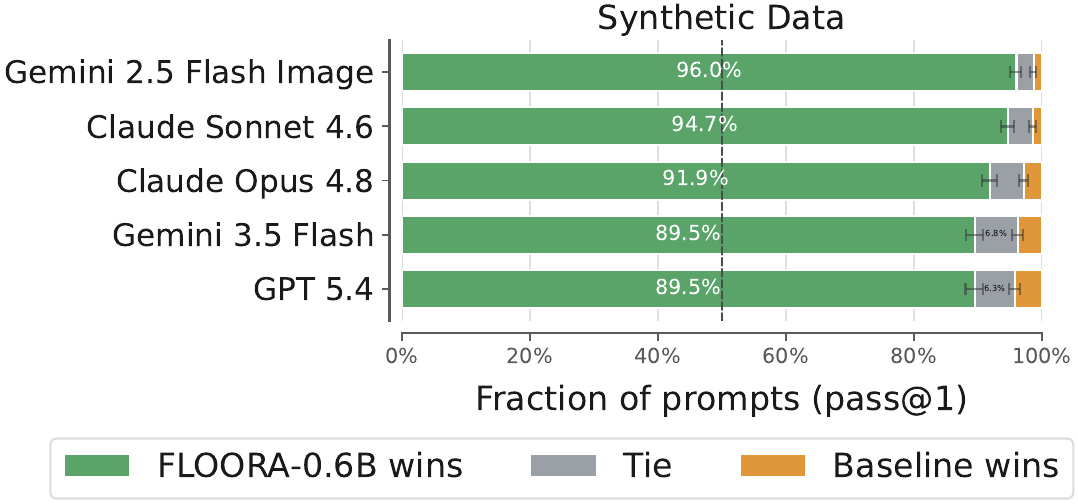}
        \caption{Win rates on the synthetic test set.}
    \end{subfigure}
    \caption{Pairwise VLM judge outcomes for the FLOORA-0.6B (Qwen3) GRPO (RM+VR) model against frontier baselines. Bars show the fraction of prompts for which the FLOORA model wins, ties, or loses. Our model is preferred over all evaluated frontier baselines. Error bars indicate 95\% Wilson intervals over the evaluation prompts.}    
    \label{fig:winrates_frontier_app}
\end{figure}

\paragraph{Pairwise Win Rates Using a VLM Judge.} 
We report the full pairwise VLM judge outcomes (\cref{app:vlm_judge}) in \cref{fig:winrates_frontier_app}, supplementing \cref{fig:winrates_frontier}. FLOORA-0.6B is preferred over every frontier baseline, with win rates of 73.8\%--96.0\% across OSM and synthetic prompts. These results show that the gains extend beyond verifier-defined checks to visual and architectural quality.

\paragraph{Human Evaluations.}
We conduct a human preference evaluation on 100 test prompts, comprising 50 OSM and 50 synthetic samples. For each prompt, 10 labelers view outputs from all six models and select the single best layout according to the architectural criteria in \cref{app:labeling_info}. The evaluation labelers are distinct from the architects who provided training feedback. Model identities are hidden, and output order is randomized for each prompt to mitigate position bias. 

We report each model's preference rate as its fraction of votes across all labelers and prompts. Since the labelers evaluate the same prompts, votes within a prompt are correlated and cannot be treated as independent. We therefore compute 95\% CIs using a cluster bootstrap over prompts. In each of 10,000 iterations, we sample 100 prompts with replacement, retain all 10 associated votes for each sampled prompt, and recompute each model's preference rate. The 2.5th and 97.5th percentiles of the bootstrap distribution define the confidence interval. This preserves the clustered evaluation structure and avoids understating uncertainty by treating all 1,000 votes as independent. 

Results are shown in \cref{fig:human_eval}. The FLOORA-0.6B output is selected as the best layout in 89.3\% of evaluations, substantially exceeding all frontier baselines.

\paragraph{Qualitative Comparisons.}
The comparisons in \cref{app:results_qualitative_frontier} illustrate this distinction. While some frontier-model outputs pass automatic checks, they often exhibit weak corridor connectivity, irregular unit subdivisions, or implausible proportions. In contrast, our model more consistently produces coherent circulation, regular unit organization, and architecturally plausible layouts, suggesting that domain-specific representation and post-training improve both validity and design quality.

\subsection{Qualitative Analysis}
\label[appendix]{app:results_qualitative}

\subsubsection{Qualitative Comparison with Frontier Models}
\label[appendix]{app:results_qualitative_frontier}

\begin{figure}[h!]
    \centering
    \begin{subfigure}[b]{\textwidth}
        \centering
        \includegraphics[width=0.92\textwidth]{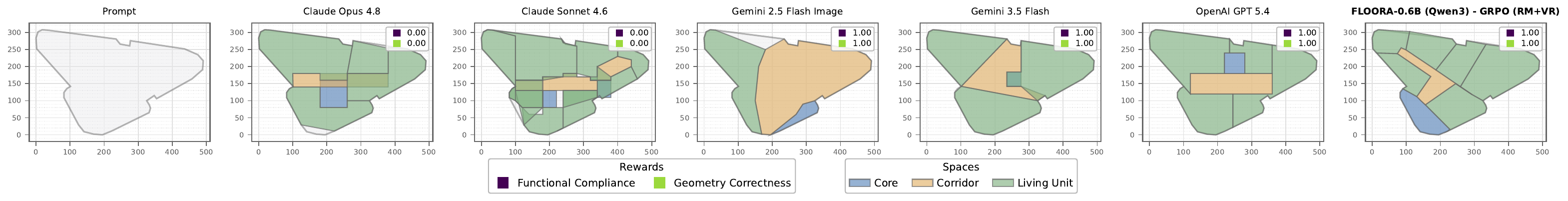}
    \end{subfigure}

    \centering
    \begin{subfigure}[b]{\textwidth}
        \centering
        \includegraphics[width=0.92\textwidth]{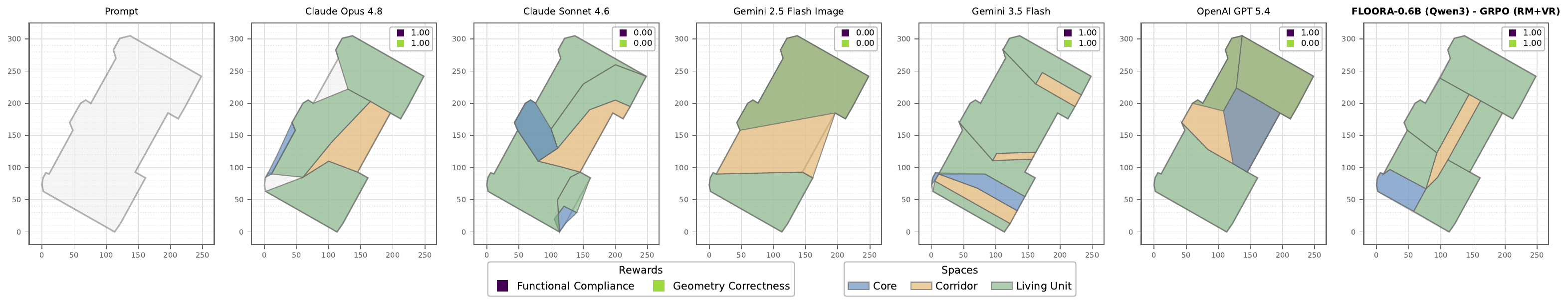}
    \end{subfigure}

    \centering
    \begin{subfigure}[b]{\textwidth}
        \centering
        \includegraphics[width=0.92\textwidth]{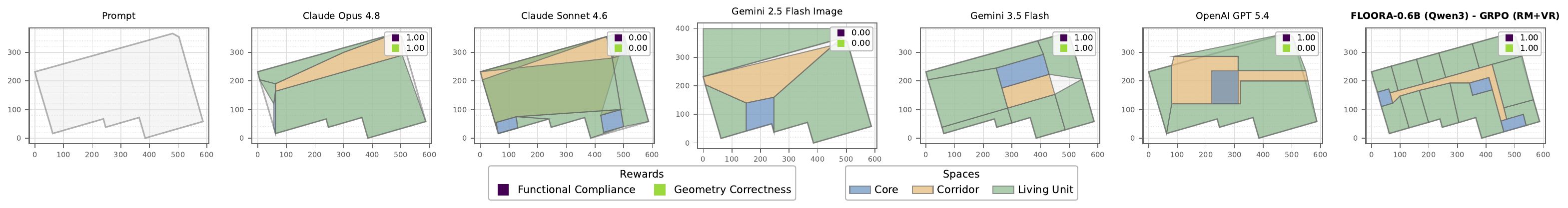}
    \end{subfigure}

    \centering
    \begin{subfigure}[b]{\textwidth}
        \centering
        \includegraphics[width=0.92\textwidth]{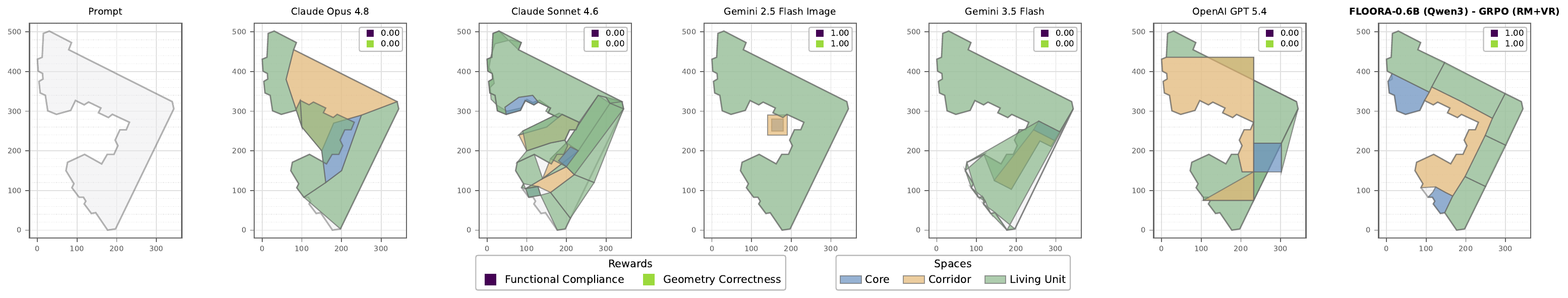}
    \end{subfigure}

    \centering
    \begin{subfigure}[b]{\textwidth}
        \centering
        \includegraphics[width=0.92\textwidth]{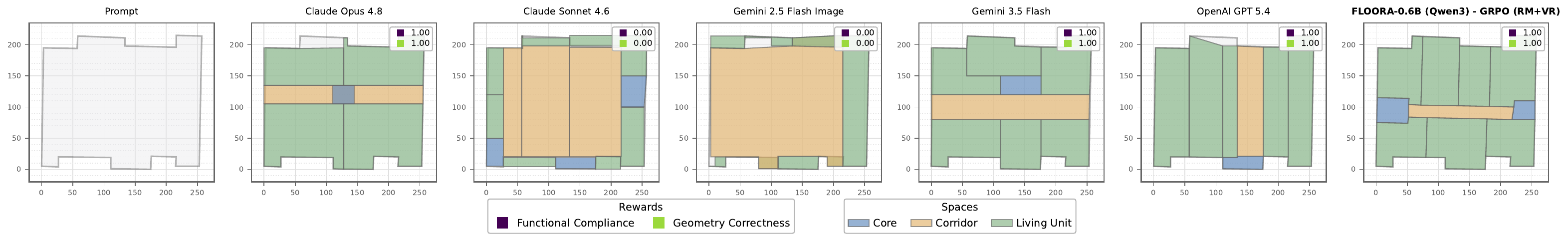}
    \end{subfigure}

    \centering
    \begin{subfigure}[b]{\textwidth}
        \centering
        \includegraphics[width=0.92\textwidth]{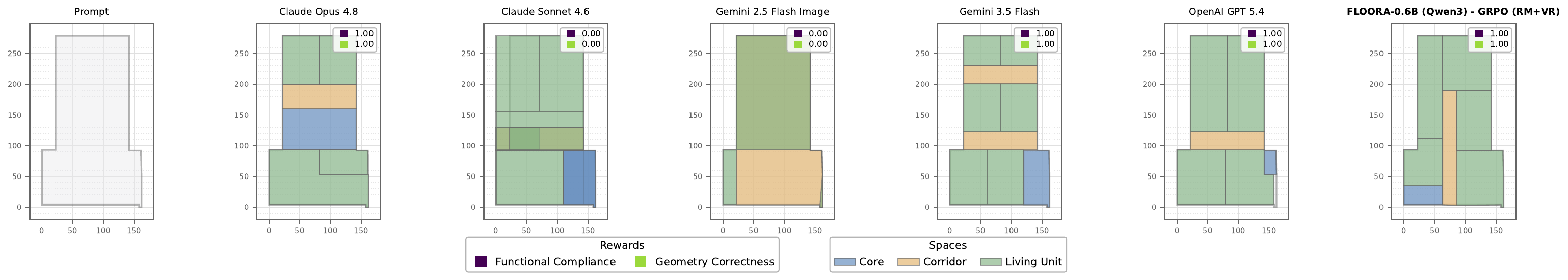}
    \end{subfigure}

    \centering
    \begin{subfigure}[b]{\textwidth}
        \centering
        \includegraphics[width=0.92\textwidth]{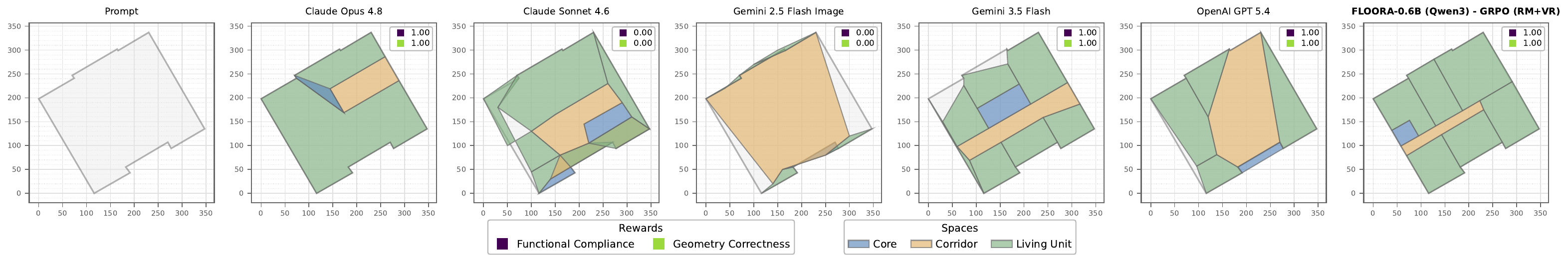}
    \end{subfigure}

    \centering
    \begin{subfigure}[b]{\textwidth}
        \centering
        \includegraphics[width=0.92\textwidth]{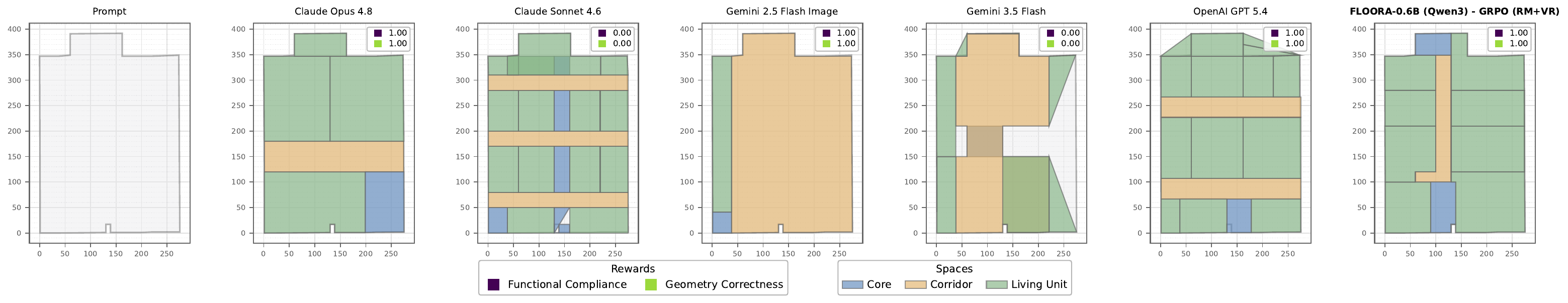}
    \end{subfigure}

    \centering
    \begin{subfigure}[b]{\textwidth}
        \centering
        \includegraphics[width=0.92\textwidth]{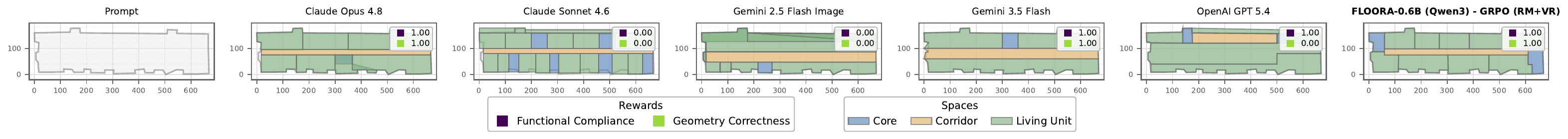}
    \end{subfigure}

    \centering
    \begin{subfigure}[b]{\textwidth}
        \centering
        \includegraphics[width=0.92\textwidth]{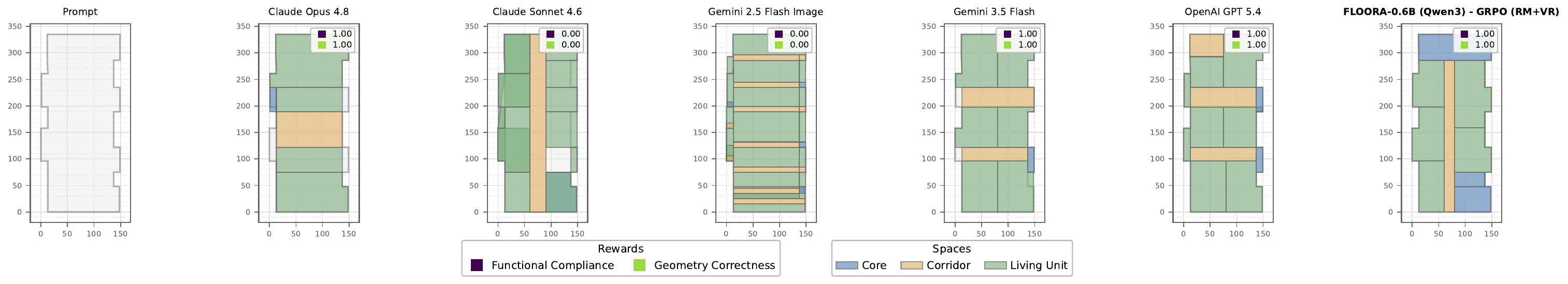}
    \end{subfigure}

    \caption{Qualitative comparison between the FLOORA-0.6B (Qwen3) GRPO (RM+VR) model and frontier models on the \textbf{OSM} test set.}
    \label{fig:qualitative_frontier_osm}
\end{figure}

\begin{figure}[h!]
    \centering
    \begin{subfigure}[b]{\textwidth}
        \centering
        \includegraphics[width=0.92\textwidth]{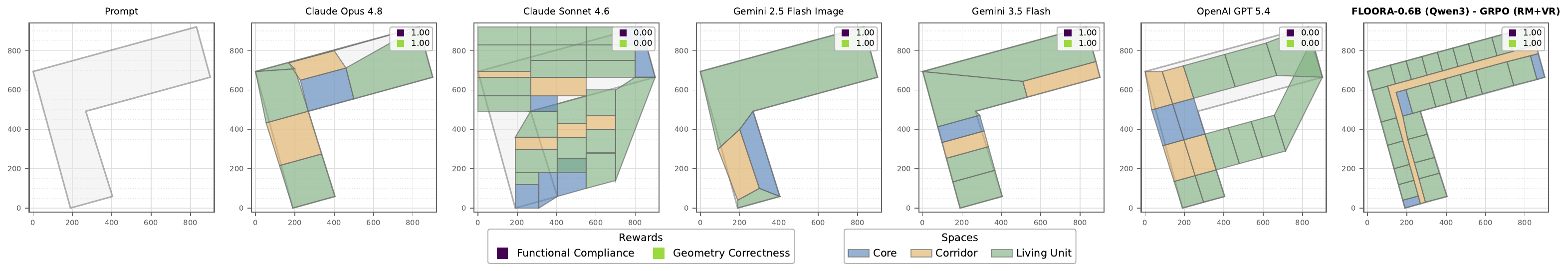}
    \end{subfigure}

    \centering
    \begin{subfigure}[b]{\textwidth}
        \centering
        \includegraphics[width=0.92\textwidth]{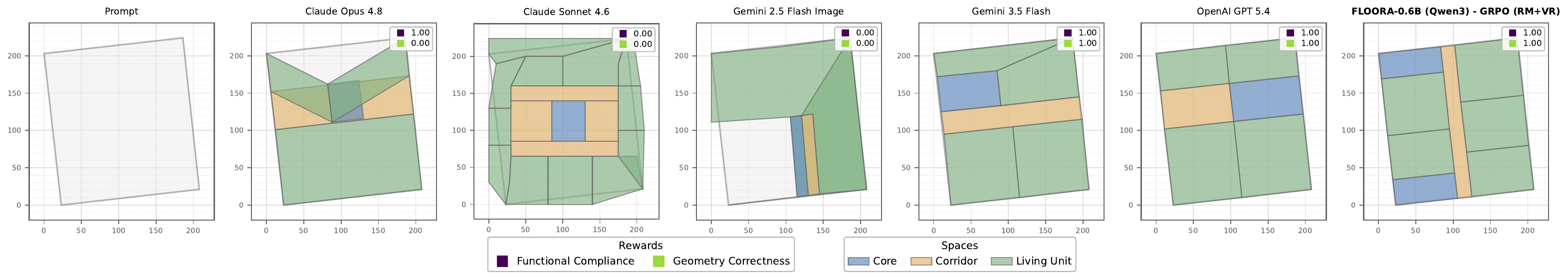}
    \end{subfigure}

    \centering
    \begin{subfigure}[b]{\textwidth}
        \centering
        \includegraphics[width=0.92\textwidth]{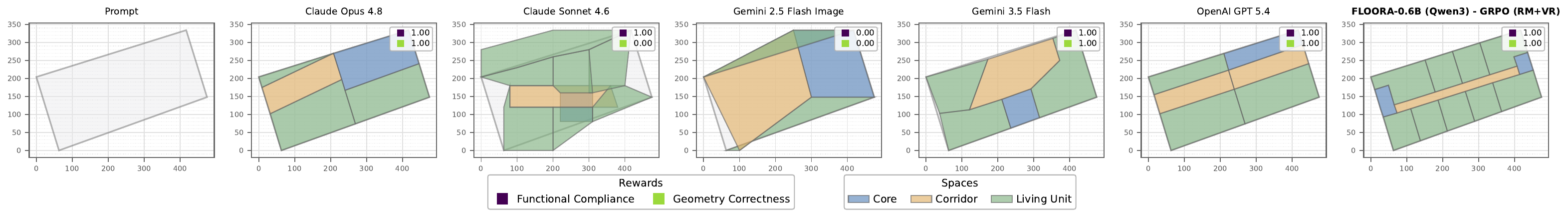}
    \end{subfigure}

    \centering
    \begin{subfigure}[b]{\textwidth}
        \centering
        \includegraphics[width=0.92\textwidth]{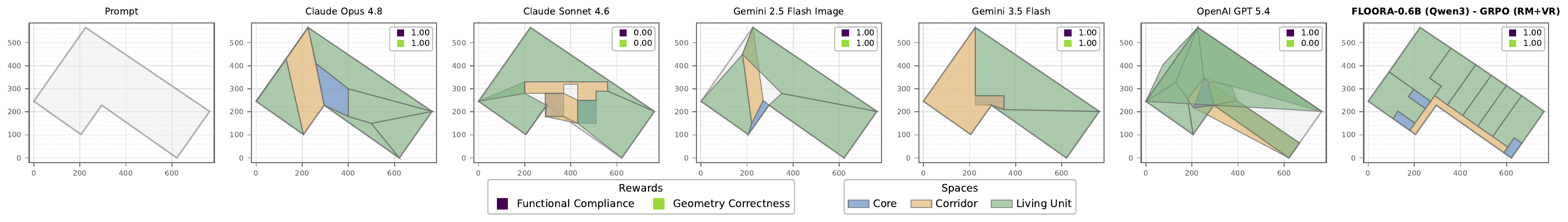}
    \end{subfigure}

    \centering
    \begin{subfigure}[b]{\textwidth}
        \centering
        \includegraphics[width=0.92\textwidth]{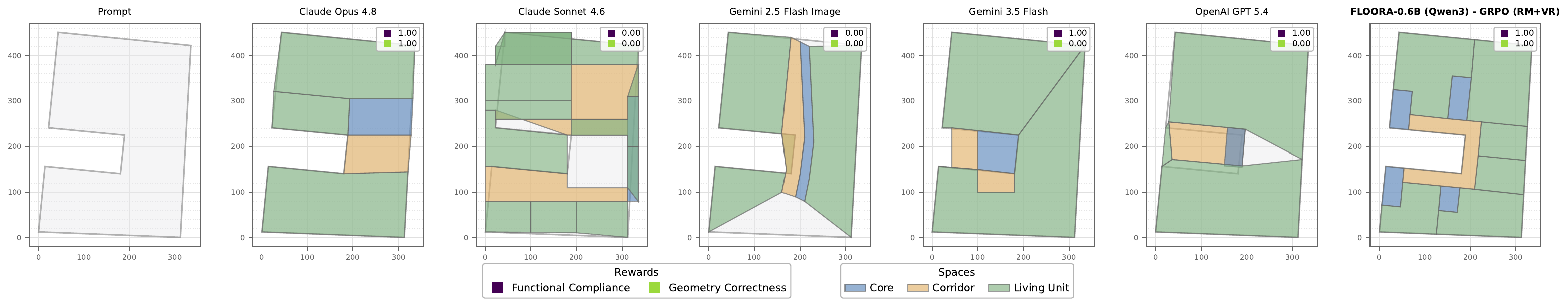}
    \end{subfigure}

    \centering
    \begin{subfigure}[b]{\textwidth}
        \centering
        \includegraphics[width=0.92\textwidth]{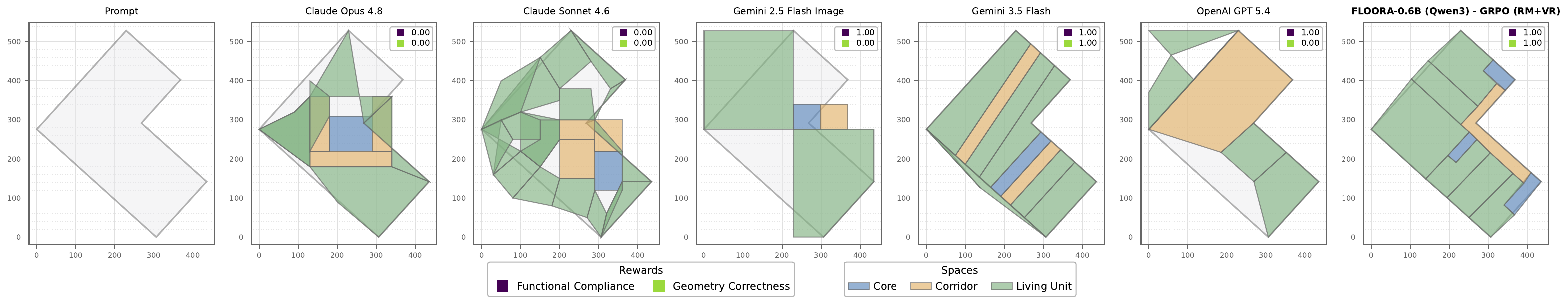}
    \end{subfigure}

    \centering
    \begin{subfigure}[b]{\textwidth}
        \centering
        \includegraphics[width=0.92\textwidth]{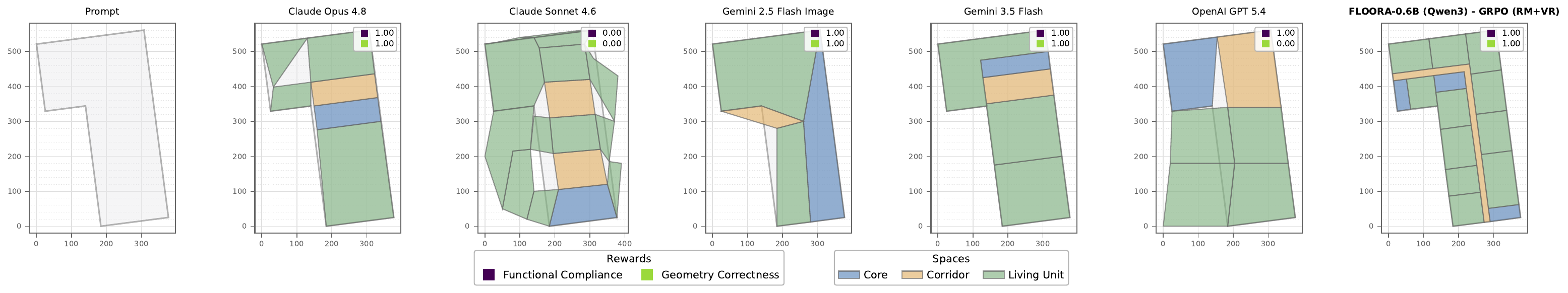}
    \end{subfigure}

    \centering
    \begin{subfigure}[b]{\textwidth}
        \centering
        \includegraphics[width=0.92\textwidth]{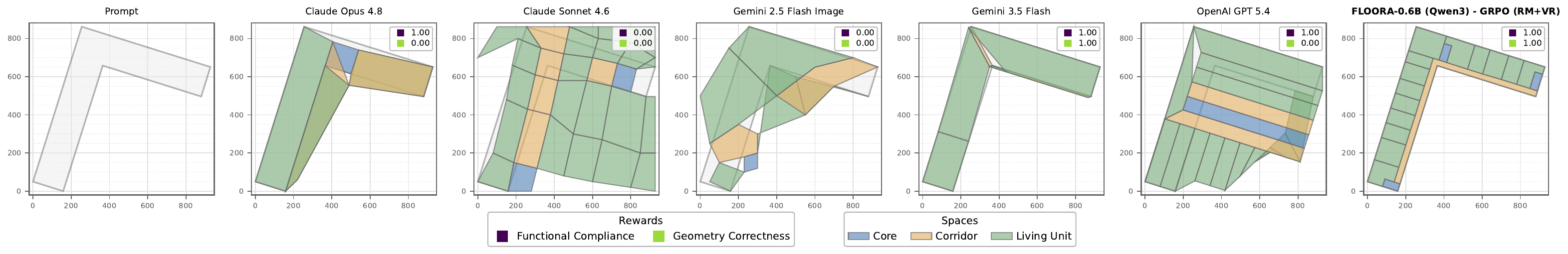}
    \end{subfigure}

    \centering
    \begin{subfigure}[b]{\textwidth}
        \centering
        \includegraphics[width=0.92\textwidth]{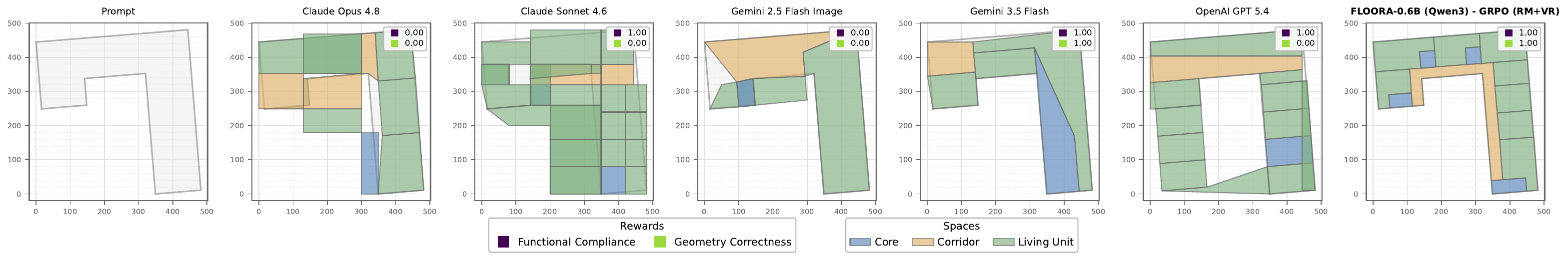}
    \end{subfigure}

    \centering
    \begin{subfigure}[b]{\textwidth}
        \centering
        \includegraphics[width=0.92\textwidth]{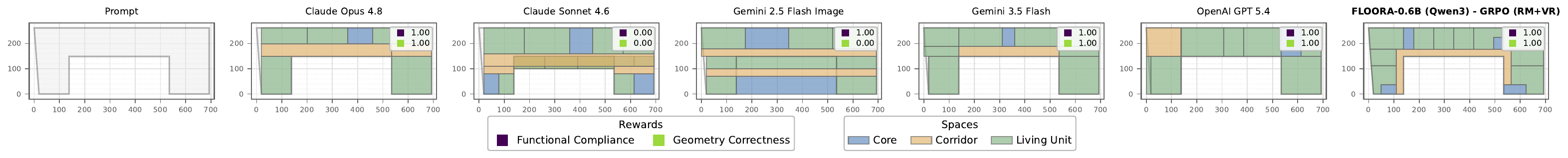}
    \end{subfigure}

    \caption{Qualitative comparison between the FLOORA-0.6B (Qwen3) GRPO (RM+VR) model and frontier models on the \textbf{synthetic} test set.}
    \label{fig:qualitative_frontier_synthetic}
\end{figure}

\cref{fig:qualitative_frontier_osm,fig:qualitative_frontier_synthetic} compare FLOORA-0.6B GRPO (RM+VR) with frontier baselines on OSM and synthetic samples. Even when frontier outputs pass geometric and functional checks, they may remain architecturally impractical, with poor connectivity, irregular units, or implausible proportions.


\clearpage
\subsubsection{Post-Training Improvements}
\label[appendix]{app:results_qualitative_progression}

\begin{figure}[h!]
    \centering
    \begin{subfigure}[b]{\textwidth}
        \centering
        \includegraphics[width=0.92\textwidth]{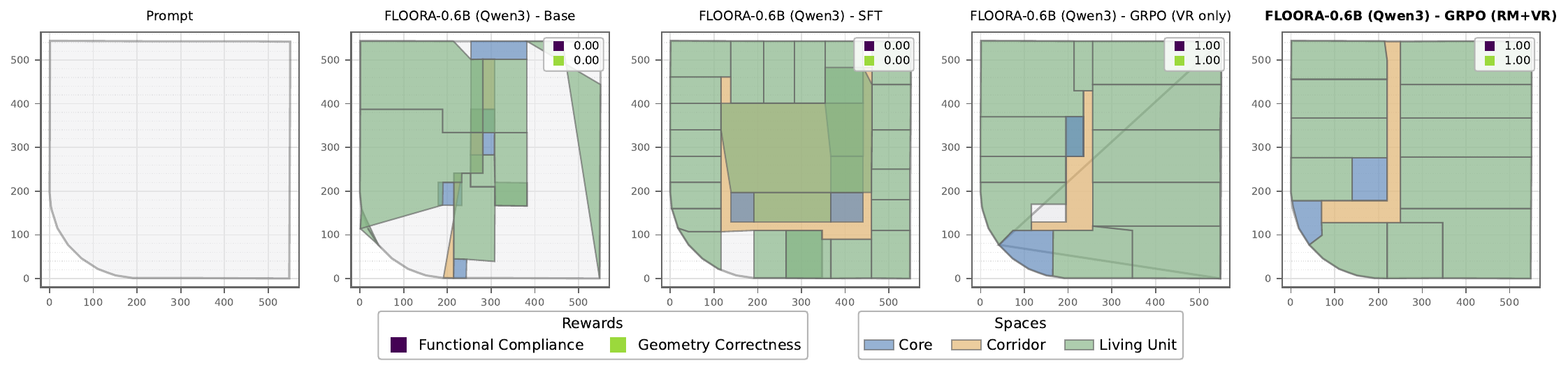}
    \end{subfigure}

    \centering
    \begin{subfigure}[b]{\textwidth}
        \centering
        \includegraphics[width=0.92\textwidth]{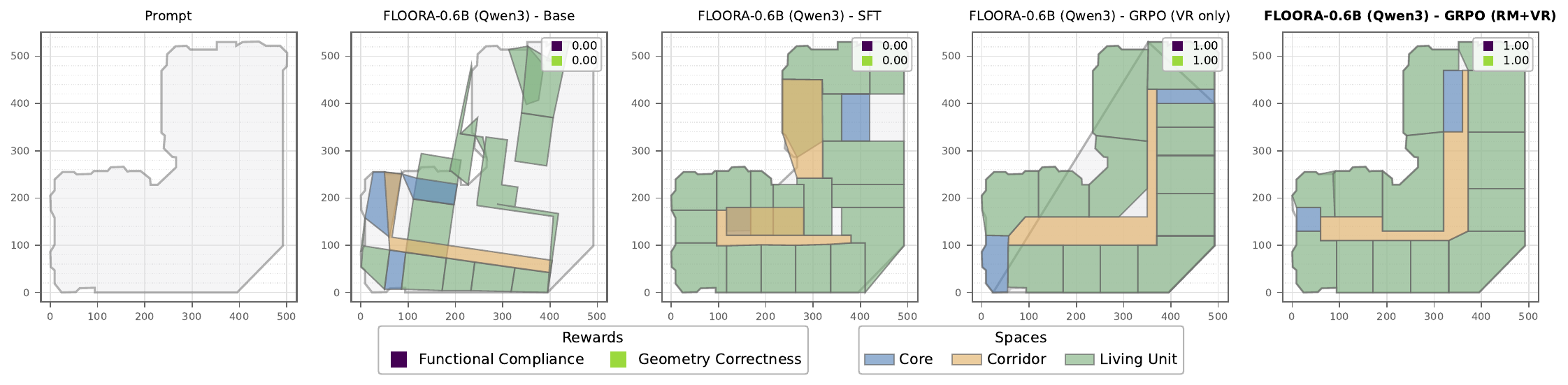}
    \end{subfigure}

    \centering
    \begin{subfigure}[b]{\textwidth}
        \centering
        \includegraphics[width=0.92\textwidth]{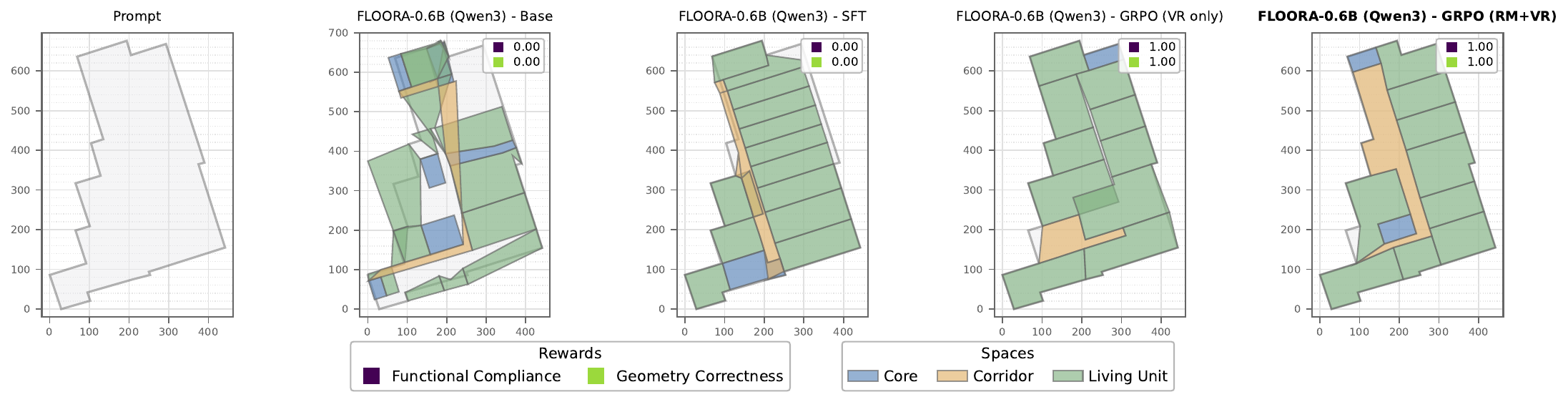}
    \end{subfigure}

    \centering
    \begin{subfigure}[b]{\textwidth}
        \centering
        \includegraphics[width=0.92\textwidth]{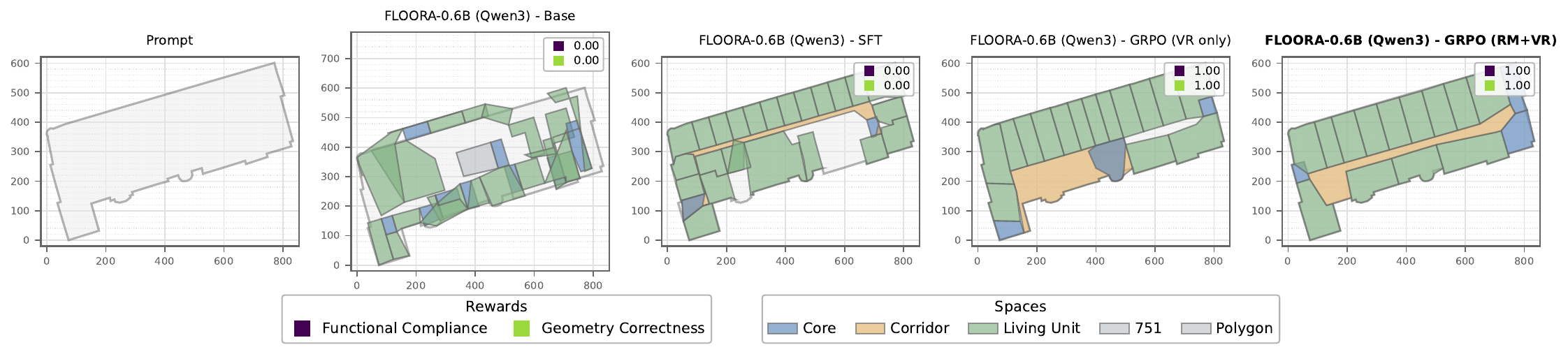}
    \end{subfigure}

    \centering
    \begin{subfigure}[b]{\textwidth}
        \centering
        \includegraphics[width=0.92\textwidth]{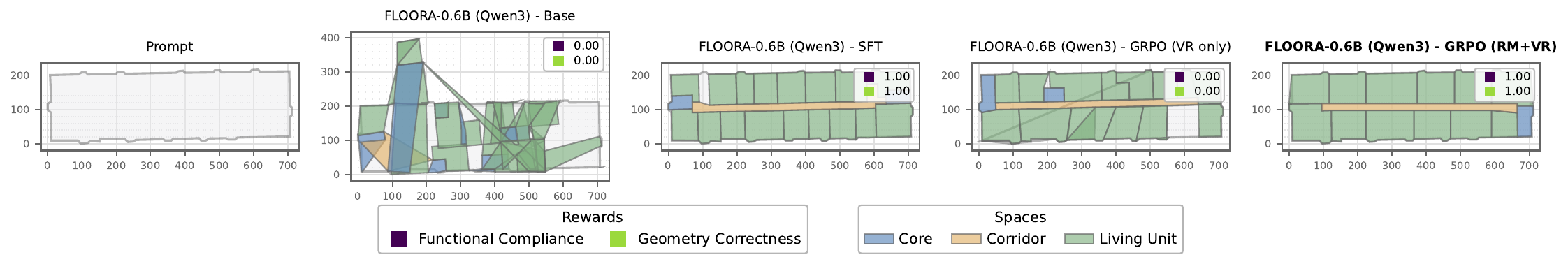}
    \end{subfigure}

    \caption{Qualitative comparison between the FLOORA-0.6B (Qwen3) Base, SFT, GRPO (RM+VR), and GRPO (VR only) models on the \textbf{OSM} test set.}
    \label{fig:qualitative_progression_osm}
\end{figure}

\begin{figure}[h!]
    \centering
    \begin{subfigure}[b]{\textwidth}
        \centering
        \includegraphics[width=0.92\textwidth]{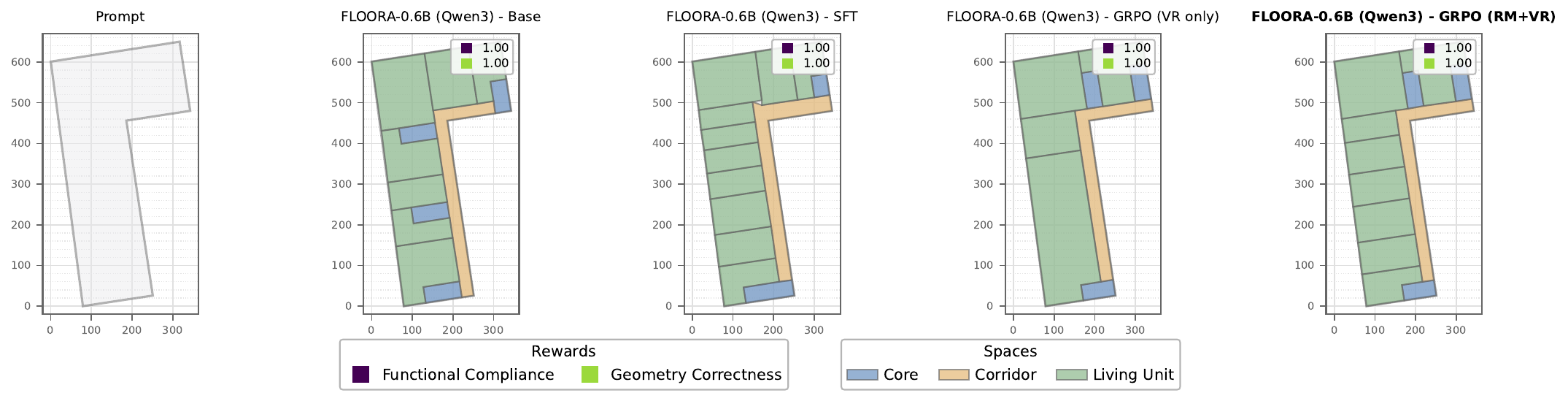}
    \end{subfigure}

    \centering
    \begin{subfigure}[b]{\textwidth}
        \centering
        \includegraphics[width=0.92\textwidth]{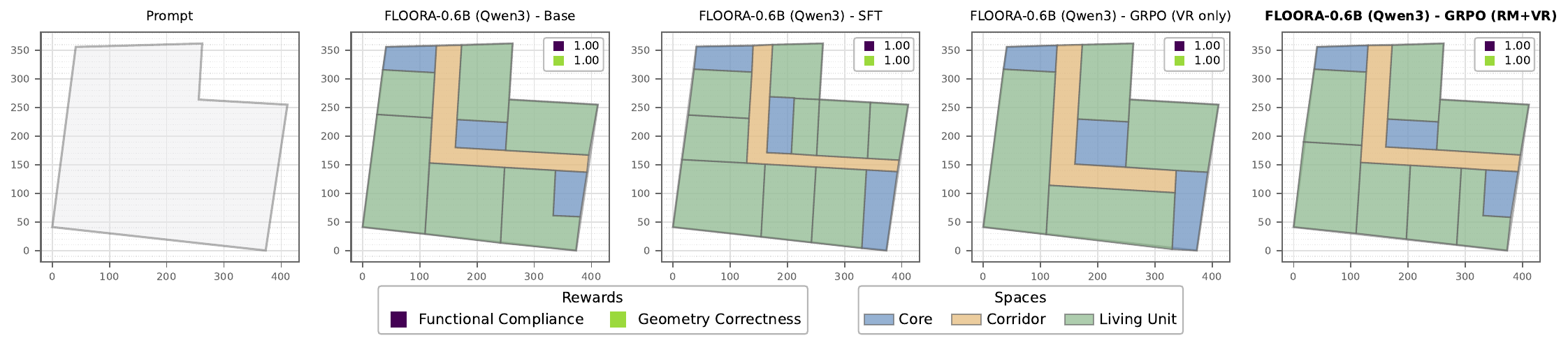}
    \end{subfigure}

    \centering
    \begin{subfigure}[b]{\textwidth}
        \centering
        \includegraphics[width=0.92\textwidth]{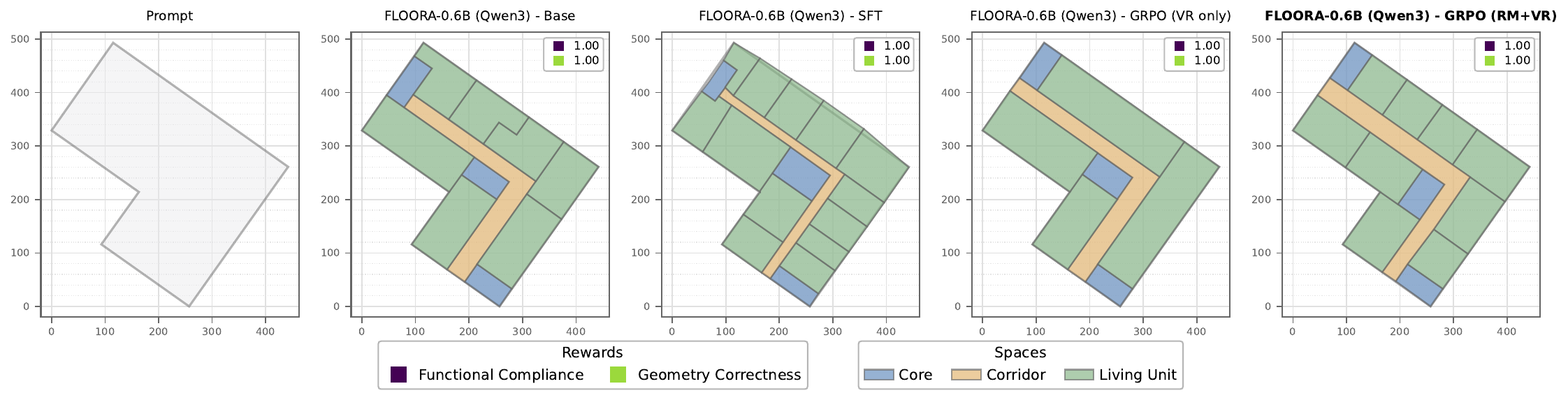}
    \end{subfigure}

    \centering
    \begin{subfigure}[b]{\textwidth}
        \centering
        \includegraphics[width=0.92\textwidth]{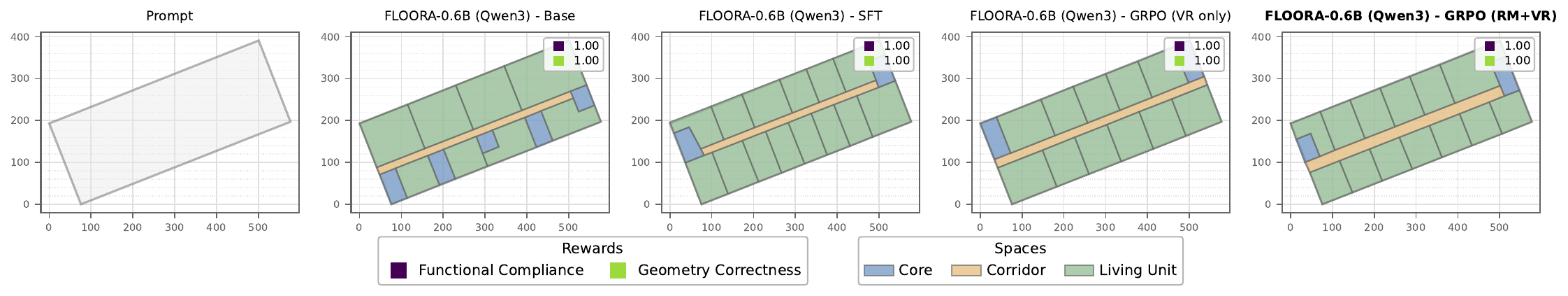}
    \end{subfigure}

    \centering
    \begin{subfigure}[b]{\textwidth}
        \centering
        \includegraphics[width=0.92\textwidth]{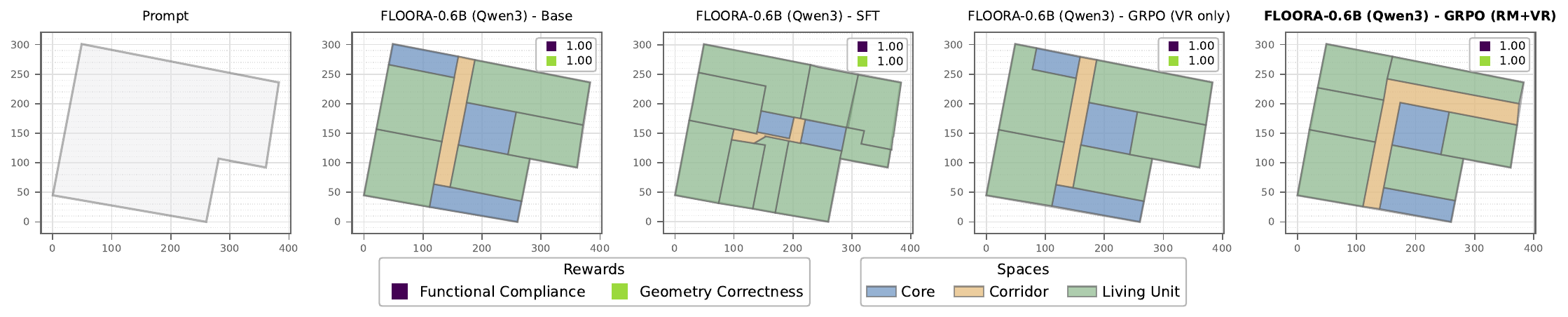}
    \end{subfigure}

    \centering
    \begin{subfigure}[b]{\textwidth}
        \centering
        \includegraphics[width=0.92\textwidth]{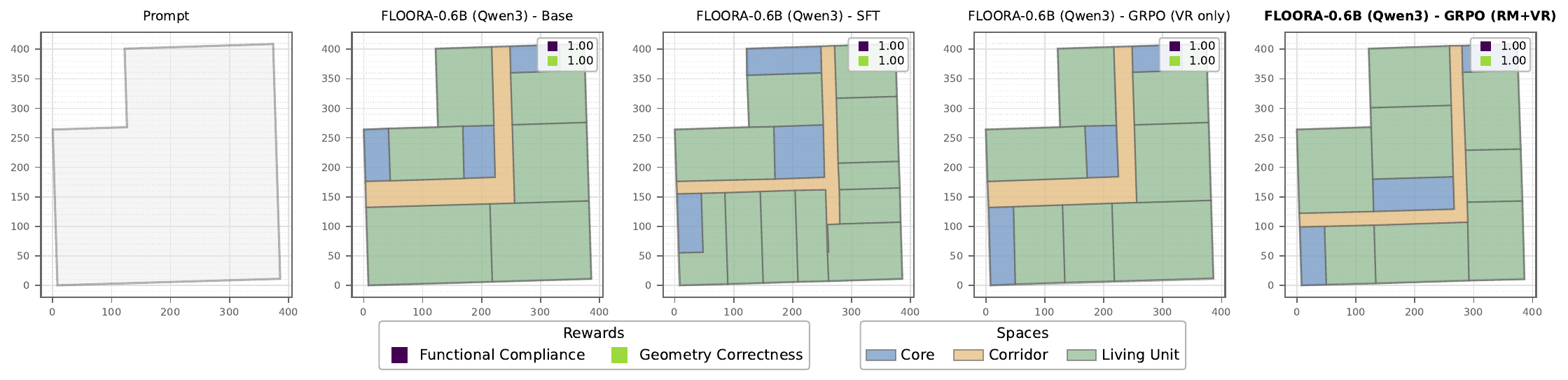}
    \end{subfigure}

    \centering
    \begin{subfigure}[b]{\textwidth}
        \centering
        \includegraphics[width=0.92\textwidth]{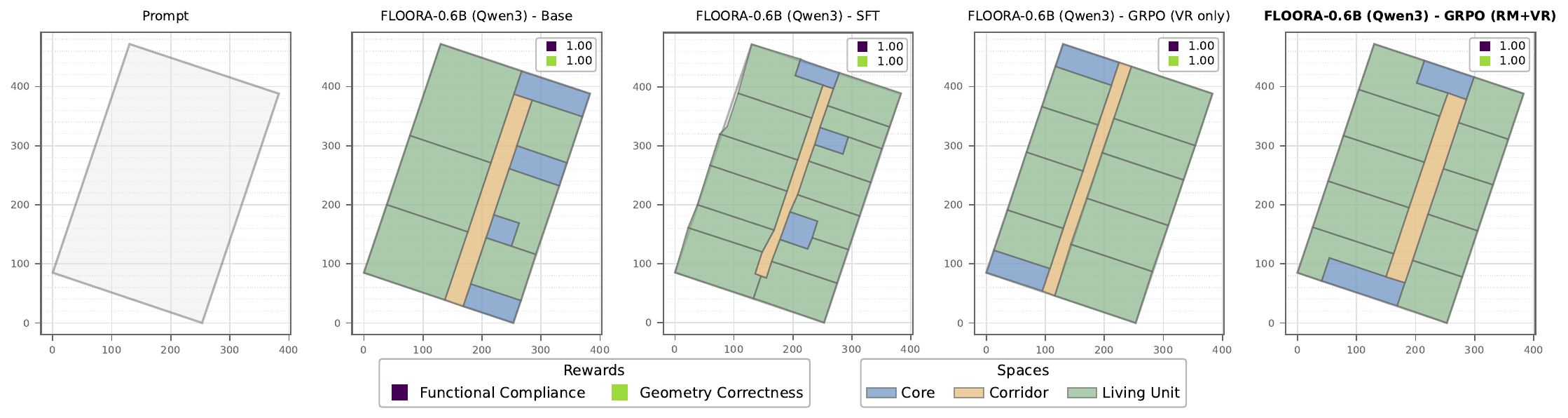}
    \end{subfigure}

    \caption{Qualitative comparison between the FLOORA-0.6B (Qwen3) Base, SFT, GRPO (RM+VR), and GRPO (VR only) models on the \textbf{synthetic} test set.}
    \label{fig:qualitative_progression_synthetic}
\end{figure}

\cref{fig:qualitative_progression_osm,fig:qualitative_progression_synthetic} show the qualitative improvements from SFT, GRPO (RM+VR), and GRPO (VR only) for the FLOORA-0.6B (Qwen3) model. Both GRPO variants substantially improve over the base and SFT models, producing more complete and geometrically regular layouts. However, GRPO (RM+VR) is preferable: it more consistently produces coherent space layouts, better unit sizes and proportions, better-placed cores with more appropriate counts, and more effective corridor positioning. This suggests that verifiable rewards enforce hard constraints, while the reward model adds an architectural-quality signal.

%% file: content/table_pt_results.tex
\begin{table}[t!]
\centering
\footnotesize
\caption{Pass@1/3/5 performance of the base models on the \textbf{OSM} test set, reported with 95\% CIs across 5 evaluation seeds. Best value within each model family is shown in bold.}
\label{tab:pt_osm}
\begin{tabular}{l ccc ccc ccc}
\toprule
 & \multicolumn{3}{c}{\textbf{Functional Compliance}} & \multicolumn{3}{c}{\textbf{Geometric Correctness}} & \multicolumn{3}{c}{\textbf{Total Reward}} \\
\cmidrule(lr){2-4}\cmidrule(lr){5-7}\cmidrule(lr){8-10}
\textbf{Model} & @1 & @3 & @5 & @1 & @3 & @5 & @1 & @3 & @5 \\
\midrule
FLOORA-0.6B (Qwen3) -- Base & 34.2 $\pm$ 0.4 & 51.2 $\pm$ 0.8 & 58.8 $\pm$ 1.2 & 27.8 $\pm$ 0.3 & 38.5 $\pm$ 0.6 & 43.2 $\pm$ 0.8 & 23.8 $\pm$ 0.1 & 33.4 $\pm$ 0.4 & 38.1 $\pm$ 0.7 \\
FLOORA-1.7B (Qwen3) -- Base & \textbf{35.6 $\pm$ 0.4} & \textbf{53.3 $\pm$ 0.5} & \textbf{61.2 $\pm$ 0.4} & \textbf{28.7 $\pm$ 0.1} & \textbf{40.5 $\pm$ 0.4} & \textbf{45.5 $\pm$ 0.6} & \textbf{24.9 $\pm$ 0.3} & \textbf{35.8 $\pm$ 0.6} & \textbf{40.7 $\pm$ 0.9} \\
\midrule
FLOORA-70M (Pythia) -- Base & 26.6 $\pm$ 0.4 & 47.6 $\pm$ 0.6 & 56.8 $\pm$ 0.8 & 4.3 $\pm$ 0.2 & 8.7 $\pm$ 0.5 & 11.1 $\pm$ 0.6 & 4.2 $\pm$ 0.2 & 8.6 $\pm$ 0.4 & 11.0 $\pm$ 0.6 \\
FLOORA-160M (Pythia) -- Base & 26.3 $\pm$ 0.3 & 43.8 $\pm$ 0.5 & 52.3 $\pm$ 0.8 & 13.1 $\pm$ 0.2 & 21.3 $\pm$ 0.2 & 24.9 $\pm$ 0.2 & 12.4 $\pm$ 0.1 & 20.2 $\pm$ 0.3 & 23.7 $\pm$ 0.5 \\
FLOORA-410M (Pythia) -- Base & 34.6 $\pm$ 0.4 & 52.3 $\pm$ 0.6 & 60.6 $\pm$ 0.7 & 26.7 $\pm$ 0.3 & 36.9 $\pm$ 0.5 & 41.5 $\pm$ 0.5 & 23.0 $\pm$ 0.1 & 32.2 $\pm$ 0.4 & 36.8 $\pm$ 0.7 \\
FLOORA-1.4B (Pythia) -- Base & \textbf{37.6 $\pm$ 0.3} & \textbf{56.8 $\pm$ 0.4} & \textbf{65.6 $\pm$ 0.7} & \textbf{28.0 $\pm$ 0.3} & \textbf{39.1 $\pm$ 0.9} & \textbf{44.0 $\pm$ 1.2} & \textbf{24.3 $\pm$ 0.4} & \textbf{34.4 $\pm$ 0.7} & \textbf{39.0 $\pm$ 1.0} \\
\bottomrule
\end{tabular}
\end{table}

\begin{table}[b!]
\centering
\footnotesize
\caption{Pass@1/3/5 performance of the base models on the \textbf{synthetic} test set, reported with 95\% CIs across 5 evaluation seeds. Best value within each model family is shown in bold.}
\label{tab:pt_synthetic}
\begin{tabular}{l ccc ccc ccc}
\toprule
 & \multicolumn{3}{c}{\textbf{Functional Compliance}} & \multicolumn{3}{c}{\textbf{Geometric Correctness}} & \multicolumn{3}{c}{\textbf{Total Reward}} \\
\cmidrule(lr){2-4}\cmidrule(lr){5-7}\cmidrule(lr){8-10}
\textbf{Model} & @1 & @3 & @5 & @1 & @3 & @5 & @1 & @3 & @5 \\
\midrule
FLOORA-0.6B (Qwen3) -- Base & 90.9 $\pm$ 0.9 & \textbf{94.3 $\pm$ 0.9} & \textbf{95.2 $\pm$ 0.9} & 96.9 $\pm$ 0.4 & 98.4 $\pm$ 0.3 & \textbf{98.5 $\pm$ 0.3} & 90.2 $\pm$ 1.0 & \textbf{94.0 $\pm$ 1.0} & \textbf{94.9 $\pm$ 1.0} \\
FLOORA-1.7B (Qwen3) -- Base & \textbf{91.3 $\pm$ 0.9} & 94.2 $\pm$ 0.8 & 95.0 $\pm$ 0.8 & \textbf{97.2 $\pm$ 0.4} & \textbf{98.4 $\pm$ 0.3} & \textbf{98.5 $\pm$ 0.3} & \textbf{90.7 $\pm$ 0.9} & 93.9 $\pm$ 0.8 & 94.7 $\pm$ 0.8 \\
\midrule
FLOORA-70M (Pythia) -- Base & 70.2 $\pm$ 1.0 & 90.6 $\pm$ 0.7 & 93.8 $\pm$ 0.6 & 32.0 $\pm$ 0.6 & 54.8 $\pm$ 0.8 & 64.0 $\pm$ 0.7 & 31.0 $\pm$ 0.6 & 53.2 $\pm$ 0.9 & 62.2 $\pm$ 0.9 \\
FLOORA-160M (Pythia) -- Base & 79.2 $\pm$ 0.6 & 91.9 $\pm$ 0.6 & 93.8 $\pm$ 0.5 & 69.6 $\pm$ 0.5 & 88.3 $\pm$ 0.6 & 92.3 $\pm$ 0.8 & 66.1 $\pm$ 0.3 & 84.7 $\pm$ 0.5 & 88.9 $\pm$ 0.9 \\
FLOORA-410M (Pythia) -- Base & 91.1 $\pm$ 0.7 & \textbf{94.7 $\pm$ 0.7} & \textbf{95.6 $\pm$ 0.7} & 96.0 $\pm$ 0.3 & 98.3 $\pm$ 0.3 & 98.4 $\pm$ 0.3 & 89.6 $\pm$ 0.7 & \textbf{94.2 $\pm$ 0.7} & \textbf{95.1 $\pm$ 0.8} \\
FLOORA-1.4B (Pythia) -- Base & \textbf{91.3 $\pm$ 0.9} & 94.4 $\pm$ 0.9 & 95.2 $\pm$ 0.9 & \textbf{96.7 $\pm$ 0.3} & \textbf{98.4 $\pm$ 0.3} & \textbf{98.5 $\pm$ 0.3} & \textbf{90.2 $\pm$ 0.9} & 94.0 $\pm$ 0.9 & 94.9 $\pm$ 1.0 \\
\bottomrule
\end{tabular}
\end{table}

%% file: content/table_rl_results.tex
\begin{table}[t!]
\centering
\scriptsize
\caption{Pass@1/3/5 performance of the post-trained models on the \textbf{OSM} test set, reported with 95\% CIs across 3 training and 5 evaluation seeds. The final column reports percentage-point changes in Total Reward relative to the base checkpoint. Best absolute value within each model family is shown in bold.}
\label{tab:rl_osm}
\begin{tabular}{l ccc ccc ccc ccc}
\toprule
 & \multicolumn{3}{c}{\textbf{Functional Compliance}} & \multicolumn{3}{c}{G\textbf{eometric Correctness}} & \multicolumn{3}{c}{\textbf{Total Reward}} & \multicolumn{3}{c}{\textbf{$\Delta$ Total Reward (vs Base)}} \\
\cmidrule(lr){2-4}\cmidrule(lr){5-7}\cmidrule(lr){8-10}\cmidrule(lr){11-13}
\textbf{Model} & @1 & @3 & @5 & @1 & @3 & @5 & @1 & @3 & @5 & @1 & @3 & @5 \\
\midrule
FLOORA-0.6B (Qwen3) -- SFT & 45.5 $\pm$ 1.0 & 73.4 $\pm$ 0.9 & 82.7 $\pm$ 0.8 & 45.1 $\pm$ 0.9 & 71.3 $\pm$ 1.0 & 79.6 $\pm$ 0.8 & 37.4 $\pm$ 1.0 & 63.9 $\pm$ 1.2 & 73.9 $\pm$ 1.1 & +13.7 & +30.5 & +35.8 \\
FLOORA-0.6B (Qwen3) -- GRPO (RM+VR) & 85.6 $\pm$ 0.3 & 95.6 $\pm$ 0.2 & 97.5 $\pm$ 0.2 & \textbf{80.9 $\pm$ 0.7} & \textbf{92.3 $\pm$ 0.5} & \textbf{94.8 $\pm$ 0.5} & \textbf{79.3 $\pm$ 0.7} & \textbf{91.3 $\pm$ 0.5} & \textbf{94.0 $\pm$ 0.5} & \textbf{+55.5} & \textbf{+57.9} & \textbf{+55.9} \\
FLOORA-0.6B (Qwen3) -- GRPO (VR only) & 84.0 $\pm$ 0.7 & 95.3 $\pm$ 0.4 & 97.2 $\pm$ 0.3 & 76.3 $\pm$ 0.5 & 90.1 $\pm$ 0.3 & 93.5 $\pm$ 0.3 & 74.7 $\pm$ 0.7 & 89.0 $\pm$ 0.5 & 92.5 $\pm$ 0.5 & +50.9 & +55.5 & +54.3 \\
FLOORA-1.7B (Qwen3) -- SFT & 46.1 $\pm$ 0.2 & 73.5 $\pm$ 0.3 & 82.5 $\pm$ 0.3 & 46.6 $\pm$ 0.2 & 72.1 $\pm$ 0.3 & 79.9 $\pm$ 0.3 & 38.8 $\pm$ 0.2 & 64.8 $\pm$ 0.3 & 74.2 $\pm$ 0.3 & +13.9 & +29.0 & +33.5 \\
FLOORA-1.7B (Qwen3) -- GRPO (RM+VR) & \textbf{85.8 $\pm$ 0.4} & 95.7 $\pm$ 0.2 & 97.4 $\pm$ 0.2 & 78.2 $\pm$ 0.1 & 90.5 $\pm$ 0.2 & 93.4 $\pm$ 0.2 & 76.9 $\pm$ 0.1 & 89.7 $\pm$ 0.2 & 92.8 $\pm$ 0.3 & +52.0 & +54.0 & +52.0 \\
FLOORA-1.7B (Qwen3) -- GRPO (VR only) & 85.0 $\pm$ 0.3 & \textbf{95.8 $\pm$ 0.1} & \textbf{97.7 $\pm$ 0.2} & 78.2 $\pm$ 0.8 & 90.9 $\pm$ 0.5 & 94.0 $\pm$ 0.4 & 76.7 $\pm$ 0.8 & 90.2 $\pm$ 0.4 & 93.5 $\pm$ 0.3 & +51.9 & +54.4 & +52.7 \\
\midrule
FLOORA-410M (Pythia) -- SFT & 36.7 $\pm$ 0.6 & 64.0 $\pm$ 0.7 & 74.5 $\pm$ 0.8 & 36.0 $\pm$ 0.7 & 61.4 $\pm$ 0.8 & 70.8 $\pm$ 0.7 & 29.6 $\pm$ 0.6 & 53.8 $\pm$ 0.7 & 64.3 $\pm$ 0.7 & +6.6 & +21.6 & +27.5 \\
FLOORA-410M (Pythia) -- GRPO (RM+VR) & 78.7 $\pm$ 1.5 & 93.3 $\pm$ 0.9 & 96.3 $\pm$ 0.5 & 69.9 $\pm$ 2.2 & 86.1 $\pm$ 1.8 & 90.5 $\pm$ 1.4 & 68.3 $\pm$ 2.3 & 85.2 $\pm$ 1.8 & 89.8 $\pm$ 1.4 & +45.3 & +53.0 & +53.0 \\
FLOORA-410M (Pythia) -- GRPO (VR only) & 79.3 $\pm$ 1.5 & 92.6 $\pm$ 1.0 & 95.5 $\pm$ 0.7 & 69.6 $\pm$ 2.7 & 84.5 $\pm$ 2.5 & 88.8 $\pm$ 2.0 & 67.8 $\pm$ 2.7 & 83.3 $\pm$ 2.5 & 87.9 $\pm$ 2.0 & +44.9 & +51.1 & +51.0 \\
FLOORA-1.4B (Pythia) -- SFT & 43.1 $\pm$ 0.5 & 70.8 $\pm$ 0.5 & 80.4 $\pm$ 0.4 & 43.1 $\pm$ 0.5 & 68.5 $\pm$ 0.5 & 76.8 $\pm$ 0.5 & 35.7 $\pm$ 0.6 & 60.8 $\pm$ 0.7 & 70.5 $\pm$ 0.7 & +11.3 & +26.5 & +31.5 \\
FLOORA-1.4B (Pythia) -- GRPO (RM+VR) & \textbf{82.4 $\pm$ 1.0} & \textbf{95.0 $\pm$ 0.5} & \textbf{97.3 $\pm$ 0.3} & \textbf{76.0 $\pm$ 1.4} & \textbf{90.1 $\pm$ 0.9} & \textbf{93.3 $\pm$ 0.7} & \textbf{74.4 $\pm$ 1.3} & \textbf{89.1 $\pm$ 0.9} & \textbf{92.6 $\pm$ 0.7} & \textbf{+50.0} & \textbf{+54.8} & \textbf{+53.6} \\
FLOORA-1.4B (Pythia) -- GRPO (VR only) & 79.1 $\pm$ 0.7 & 92.6 $\pm$ 0.4 & 95.6 $\pm$ 0.3 & 69.5 $\pm$ 0.7 & 84.4 $\pm$ 0.6 & 88.8 $\pm$ 0.5 & 67.8 $\pm$ 0.7 & 83.5 $\pm$ 0.5 & 88.1 $\pm$ 0.5 & +43.5 & +49.1 & +49.1 \\
\bottomrule
\end{tabular}
\end{table}

\begin{table}[b!]
\centering
\scriptsize
\caption{Pass@1/3/5 performance of the post-trained models on the \textbf{synthetic} test set, reported with 95\% CIs across 3 training and 5 evaluation seeds. The final column reports percentage-point changes in Total Reward relative to the base checkpoint. Best absolute value within each model family is shown in bold.}
\label{tab:rl_synthetic}
\begin{tabular}{l ccc ccc ccc ccc}
\toprule
 & \multicolumn{3}{c}{\textbf{Functional Compliance}} & \multicolumn{3}{c}{\textbf{Geometric Correctness}} & \multicolumn{3}{c}{\textbf{Total Reward}} & \multicolumn{3}{c}{\textbf{$\Delta$ Total Reward (vs Base)}} \\
\cmidrule(lr){2-4}\cmidrule(lr){5-7}\cmidrule(lr){8-10}\cmidrule(lr){11-13}
\textbf{Model} & @1 & @3 & @5 & @1 & @3 & @5 & @1 & @3 & @5 & @1 & @3 & @5 \\
\midrule
FLOORA-0.6B (Qwen3) -- SFT & 57.0 $\pm$ 1.3 & 86.5 $\pm$ 0.9 & 93.6 $\pm$ 0.5 & 49.3 $\pm$ 1.5 & 79.7 $\pm$ 1.4 & 88.8 $\pm$ 0.9 & 48.7 $\pm$ 1.5 & 79.2 $\pm$ 1.4 & 88.5 $\pm$ 0.9 & -41.5 & -14.8 & -6.4 \\
FLOORA-0.6B (Qwen3) -- GRPO (RM+VR) & 97.4 $\pm$ 0.2 & 98.5 $\pm$ 0.1 & 98.6 $\pm$ 0.1 & 96.8 $\pm$ 0.2 & 98.2 $\pm$ 0.2 & 98.3 $\pm$ 0.1 & 96.8 $\pm$ 0.2 & 98.1 $\pm$ 0.2 & 98.3 $\pm$ 0.2 & +6.6 & +4.1 & +3.3 \\
FLOORA-0.6B (Qwen3) -- GRPO (VR only) & 97.9 $\pm$ 0.2 & 98.4 $\pm$ 0.1 & 98.5 $\pm$ 0.1 & 97.5 $\pm$ 0.1 & 98.2 $\pm$ 0.1 & 98.3 $\pm$ 0.1 & 97.3 $\pm$ 0.1 & 98.1 $\pm$ 0.2 & 98.2 $\pm$ 0.2 & \textbf{+7.1} & +4.0 & +3.2 \\
FLOORA-1.7B (Qwen3) -- SFT & 57.1 $\pm$ 0.3 & 86.0 $\pm$ 0.3 & 92.9 $\pm$ 0.3 & 50.7 $\pm$ 0.3 & 80.3 $\pm$ 0.3 & 88.9 $\pm$ 0.2 & 50.1 $\pm$ 0.3 & 79.9 $\pm$ 0.3 & 88.7 $\pm$ 0.2 & -40.6 & -14.0 & -6.0 \\
FLOORA-1.7B (Qwen3) -- GRPO (RM+VR) & 97.6 $\pm$ 0.2 & 98.5 $\pm$ 0.1 & 98.6 $\pm$ 0.1 & 97.0 $\pm$ 0.2 & 98.2 $\pm$ 0.1 & 98.3 $\pm$ 0.1 & 97.0 $\pm$ 0.2 & 98.1 $\pm$ 0.1 & 98.2 $\pm$ 0.1 & +6.3 & +4.2 & +3.6 \\
FLOORA-1.7B (Qwen3) -- GRPO (VR only) & \textbf{98.2 $\pm$ 0.1} & \textbf{98.6 $\pm$ 0.1} & \textbf{98.7 $\pm$ 0.1} & \textbf{97.8 $\pm$ 0.2} & \textbf{98.3 $\pm$ 0.1} & \textbf{98.4 $\pm$ 0.1} & \textbf{97.7 $\pm$ 0.2} & \textbf{98.3 $\pm$ 0.1} & \textbf{98.3 $\pm$ 0.1} & +7.0 & \textbf{+4.3} & \textbf{+3.7} \\
\midrule
FLOORA-410M (Pythia) -- SFT & 46.2 $\pm$ 0.7 & 76.8 $\pm$ 0.6 & 86.8 $\pm$ 0.5 & 39.5 $\pm$ 0.8 & 68.8 $\pm$ 0.9 & 79.8 $\pm$ 0.7 & 39.0 $\pm$ 0.8 & 68.4 $\pm$ 0.9 & 79.5 $\pm$ 0.7 & -50.6 & -25.8 & -15.6 \\
FLOORA-410M (Pythia) -- GRPO (RM+VR) & 95.8 $\pm$ 0.4 & 98.4 $\pm$ 0.1 & 98.6 $\pm$ 0.1 & 94.0 $\pm$ 0.6 & 97.6 $\pm$ 0.1 & 98.0 $\pm$ 0.1 & 94.0 $\pm$ 0.6 & 97.6 $\pm$ 0.1 & 98.0 $\pm$ 0.1 & +4.4 & +3.4 & +2.8 \\
FLOORA-410M (Pythia) -- GRPO (VR only) & 97.0 $\pm$ 0.6 & 98.5 $\pm$ 0.1 & 98.7 $\pm$ 0.1 & 95.8 $\pm$ 0.9 & 97.9 $\pm$ 0.2 & 98.1 $\pm$ 0.2 & 95.7 $\pm$ 0.9 & 97.8 $\pm$ 0.2 & 98.1 $\pm$ 0.2 & +6.1 & +3.6 & +3.0 \\
FLOORA-1.4B (Pythia) -- SFT & 54.8 $\pm$ 0.8 & 84.2 $\pm$ 0.6 & 91.9 $\pm$ 0.4 & 48.6 $\pm$ 0.9 & 77.9 $\pm$ 0.7 & 87.0 $\pm$ 0.5 & 48.0 $\pm$ 0.9 & 77.5 $\pm$ 0.7 & 86.6 $\pm$ 0.5 & -42.2 & -16.5 & -8.2 \\
FLOORA-1.4B (Pythia) -- GRPO (RM+VR) & 96.2 $\pm$ 0.3 & 98.5 $\pm$ 0.1 & \textbf{98.7 $\pm$ 0.1} & 95.1 $\pm$ 0.3 & 97.9 $\pm$ 0.2 & 98.2 $\pm$ 0.1 & 95.1 $\pm$ 0.3 & 97.9 $\pm$ 0.2 & 98.2 $\pm$ 0.2 & +4.9 & +3.9 & +3.3 \\
FLOORA-1.4B (Pythia) -- GRPO (VR only) & \textbf{98.0 $\pm$ 0.1} & \textbf{98.5 $\pm$ 0.1} & 98.6 $\pm$ 0.1 & \textbf{97.4 $\pm$ 0.1} & \textbf{98.2 $\pm$ 0.1} & \textbf{98.3 $\pm$ 0.1} & \textbf{97.4 $\pm$ 0.1} & \textbf{98.1 $\pm$ 0.1} & \textbf{98.3 $\pm$ 0.1} & \textbf{+7.1} & \textbf{+4.2} & \textbf{+3.4} \\
\bottomrule
\end{tabular}
\end{table}

%% file: appendix/ablations.tex
\subsection{Ablation on Reward Model Normalization}
\label[appendix]{app:ablation_reward_model}
As discussed in \cref{sec:post_setup}, the reward model produces an unconstrained scalar score, whereas the verifiable rewards are bounded to $[0, 1]$ by construction. Without normalization, the RM can therefore dominate the combined reward, effectively changing the relative importance of the learned preference signal and the verifiable geometric and functional constraints. We ablate the normalization scale $\alpha$ in \cref{eq:rm_normalization} to study this effect during GRPO training. 

\begin{figure}[h!]
    \centering
    \begin{subfigure}[b]{0.45\textwidth}
        \centering
        \includegraphics[width=\textwidth]{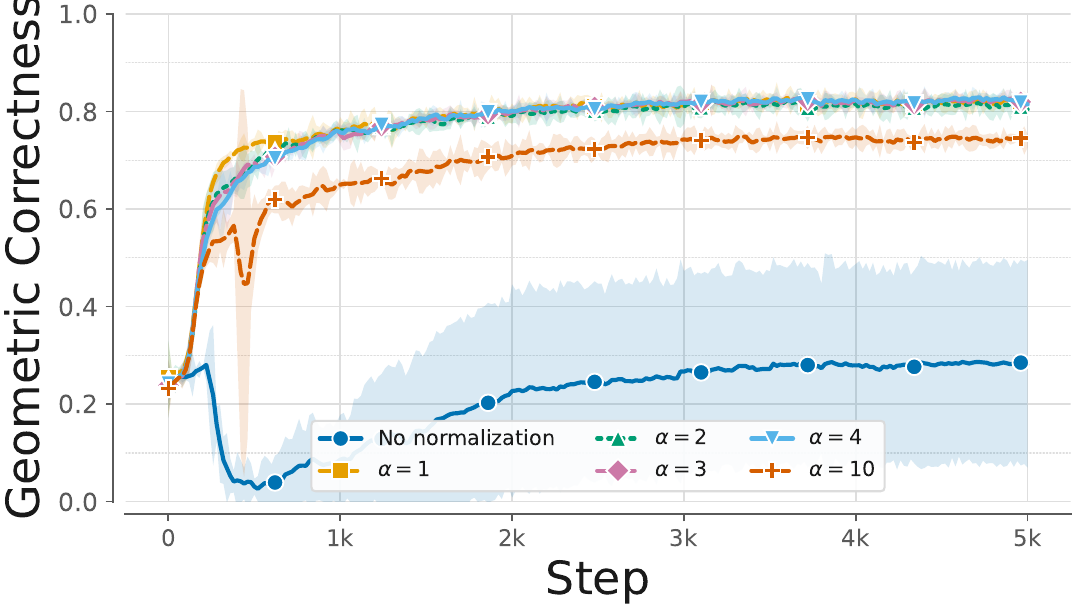}
        \caption{Geometric correctness reward curves.}
    \end{subfigure}
    \hfill
    \begin{subfigure}[b]{0.45\textwidth}
        \centering
        \includegraphics[width=\textwidth]{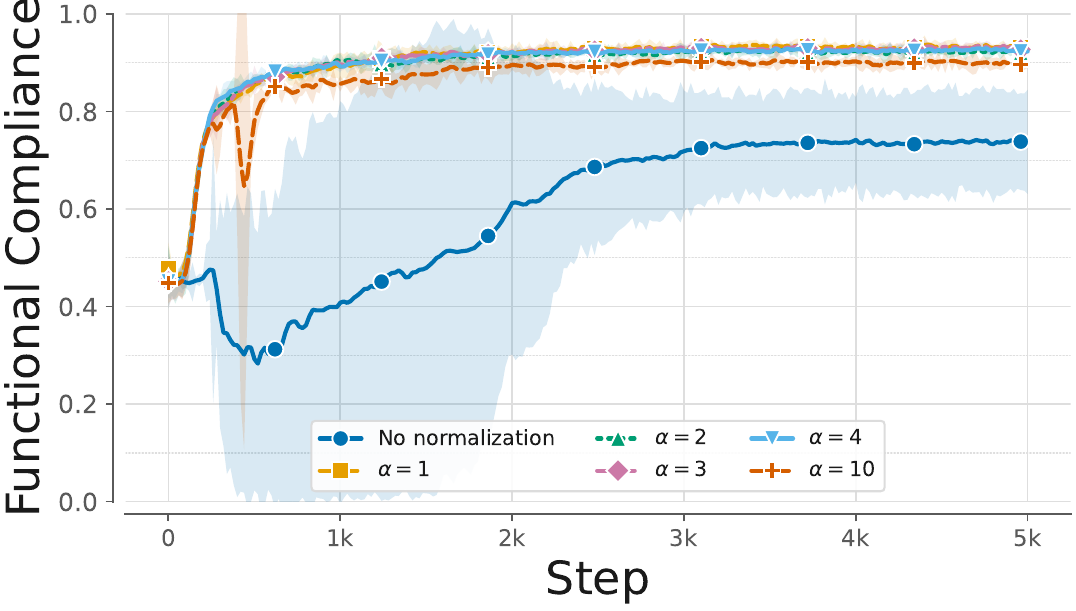}
        \caption{Functional compliance reward curves.}
    \end{subfigure}
    \vspace{1em}

    \centering
    \begin{subfigure}[b]{0.45\textwidth}
        \centering
        \includegraphics[width=\textwidth]{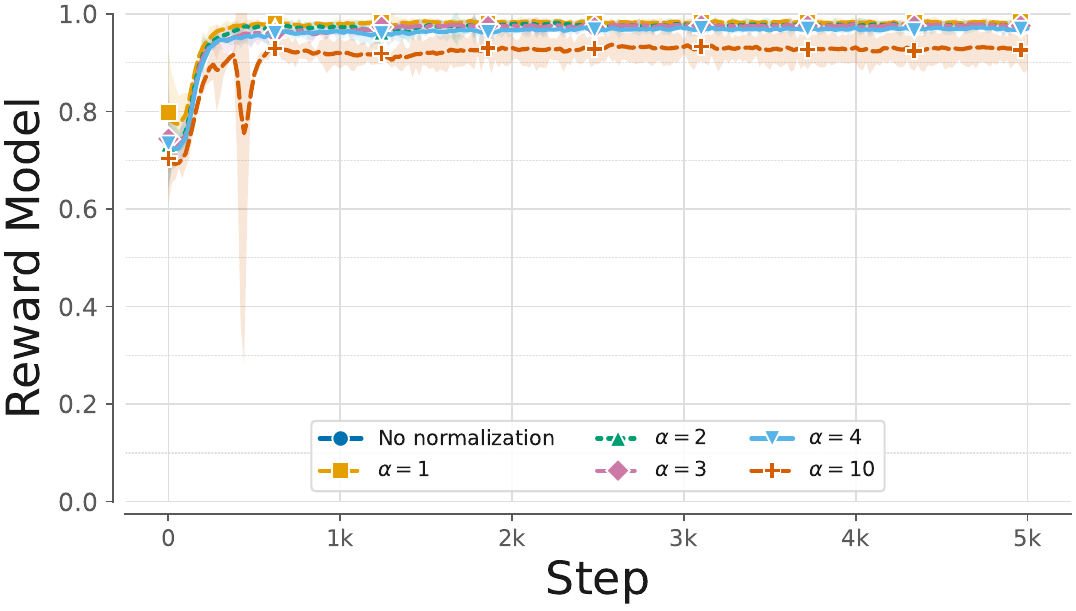}
        \caption{Reward model curves.}
    \end{subfigure}
    \hfill
    \begin{subfigure}[b]{0.45\textwidth}
        \centering
        \includegraphics[width=\textwidth]{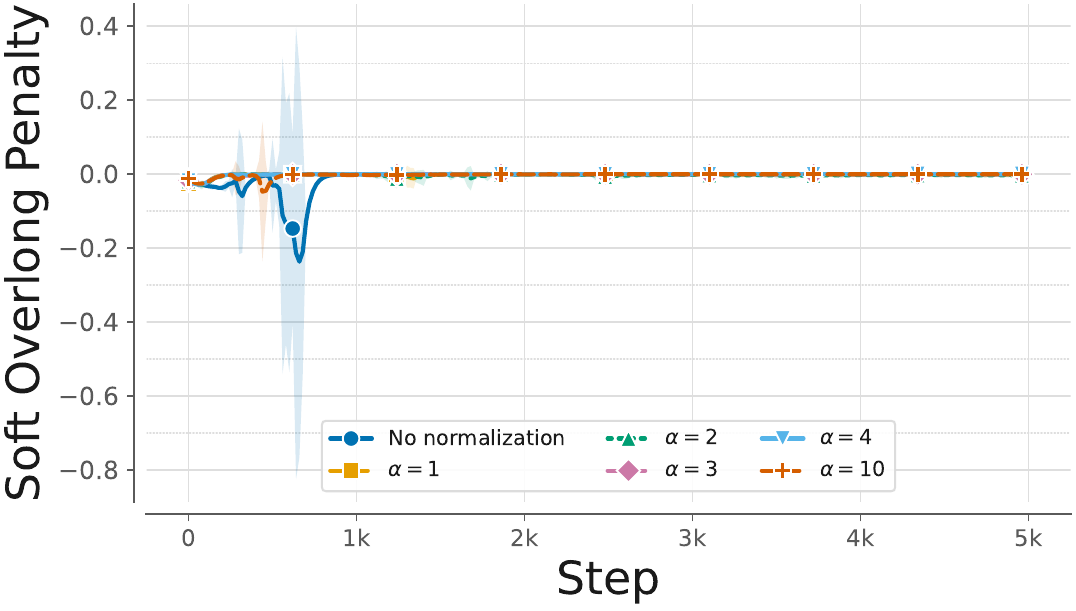}
        \caption{Soft overlong penalty curves.}
    \end{subfigure}
    
    \caption{Ablation of the RM normalization scale $\alpha$ during GRPO training with the FLOORA-0.6B (Qwen3) model. Moderate scales $\alpha \in [1,4]$ stabilize training and achieve high verifiable rewards. Shaded regions show 95\% CIs across 3 training seeds.}
    \label{fig:rm_normalization_ablation}
\end{figure}

The results in \cref{fig:rm_normalization_ablation,tab:rm_normalization_ablation} show that normalization is important for stable optimization. The experiments were conducted on the FLOORA-0.6B (Qwen3) model on 3 training seeds. Without normalization, the RM score grows to a much larger magnitude than the verifiable rewards, while geometric correctness and functional compliance remain substantially worse. In contrast, normalized settings with $\alpha \in [1,4]$ achieve consistently strong verifiable rewards. This indicates that a moderate normalization scale is sufficient to keep the learned reward aligned with the bounded verifiable rewards. Among the normalized settings, $\alpha \in [1,4]$ performs consistently well. We select $\alpha=3$ because it provides the best overall tradeoff between smooth RM learning dynamics and high verifiable rewards, while larger scales such as $\alpha=10$ degrade both geometric correctness and functional compliance.

\begin{table}[h!]
\centering
\caption{Final RM and verifiable reward scores for different normalization scales $\alpha$, reported with 95\% CIs across 3 training seeds.}
\label{tab:rm_normalization_ablation}
\scriptsize
\begin{tabular}{lcccc}
    \toprule
    \textbf{Setting} & \textbf{Reward Model} & \textbf{Geometric Correctness} & \textbf{Functional Compliance} & \textbf{Soft Overlong Penalty} \\
    \midrule
    No normalization & 28.719 $\pm$ 5.376 & 0.383 $\pm$ 0.192 & 0.761 $\pm$ 0.224 & 0.000 $\pm$ 0.000 \\
    $\alpha = 1$ & 0.994 $\pm$ 0.002 & 0.853 $\pm$ 0.026 & 0.951 $\pm$ 0.016 & 0.000 $\pm$ 0.000 \\
    $\alpha = 2$ & 0.990 $\pm$ 0.012 & 0.846 $\pm$ 0.017 & 0.944 $\pm$ 0.002 & 0.000 $\pm$ 0.000 \\
    $\alpha = 3$ & 0.988 $\pm$ 0.011 & 0.851 $\pm$ 0.027 & 0.948 $\pm$ 0.007 & 0.000 $\pm$ 0.000 \\
    $\alpha = 4$ & 0.982 $\pm$ 0.015 & 0.850 $\pm$ 0.017 & 0.943 $\pm$ 0.012 & 0.000 $\pm$ 0.000 \\
    $\alpha = 10$ & 0.954 $\pm$ 0.049 & 0.771 $\pm$ 0.032 & 0.922 $\pm$ 0.026 & 0.000 $\pm$ 0.000 \\
    \bottomrule
\end{tabular}
\end{table}

\subsection{Ablation on Original and DSL Tokenizers}
\label[appendix]{app:ablation_tokenizer}

As discussed in \cref{sec:pre-training,tab:tokenizer_stats}, the DSL tokenizer substantially reduces sequence length and pre-training cost. With the original tokenizer, the per-device batch size must be halved, and pre-training takes approximately 80 hours on 16 GPUs. In comparison, the DSL tokenizer completes pre-training in approximately 55 hours on 8 GPUs.

\begin{figure}[h!]
    \centering
    \begin{subfigure}[b]{0.32\textwidth}
        \centering
        \includegraphics[width=\textwidth]{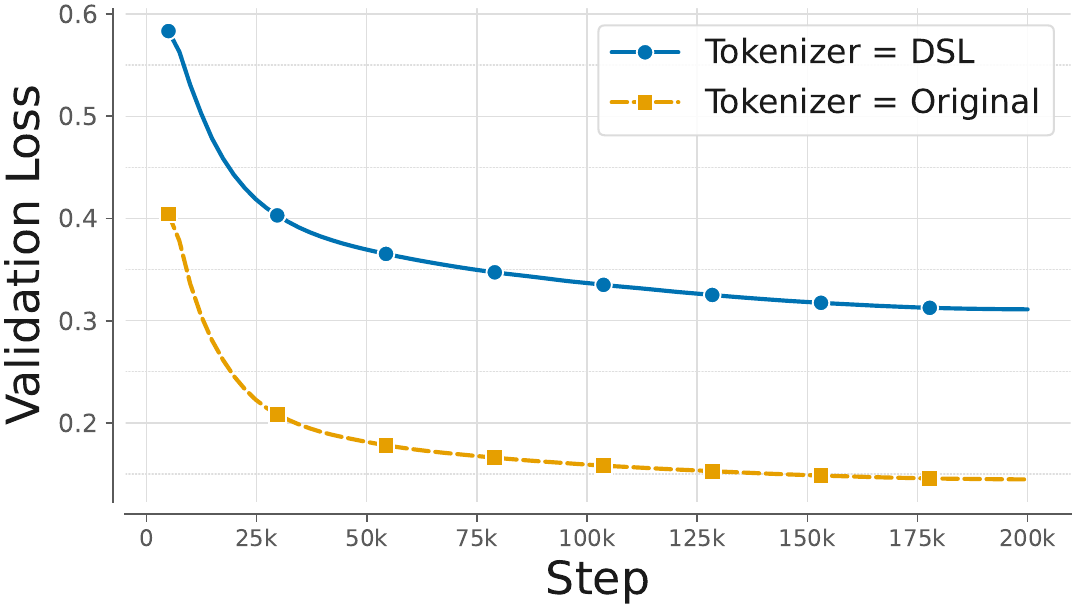}
        \caption{Pre-training validation loss.}
    \end{subfigure}
    \hfill
    \begin{subfigure}[b]{0.32\textwidth}
        \centering
        \includegraphics[width=\textwidth]{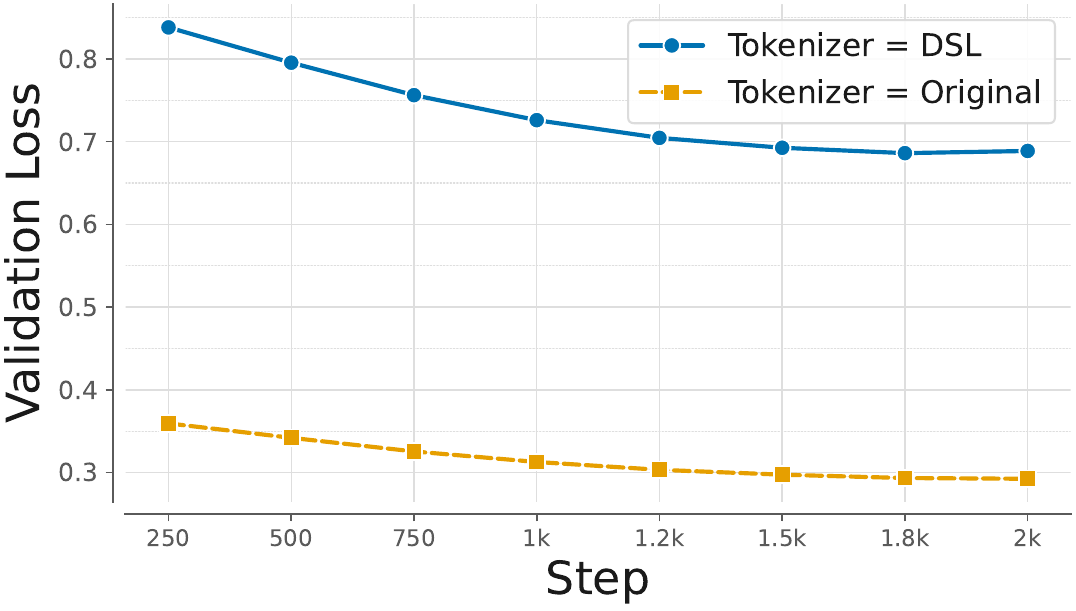}
        \caption{SFT validation loss.}
    \end{subfigure}
    \hfill
    \begin{subfigure}[b]{0.32\textwidth}
        \centering
        \includegraphics[width=\textwidth]{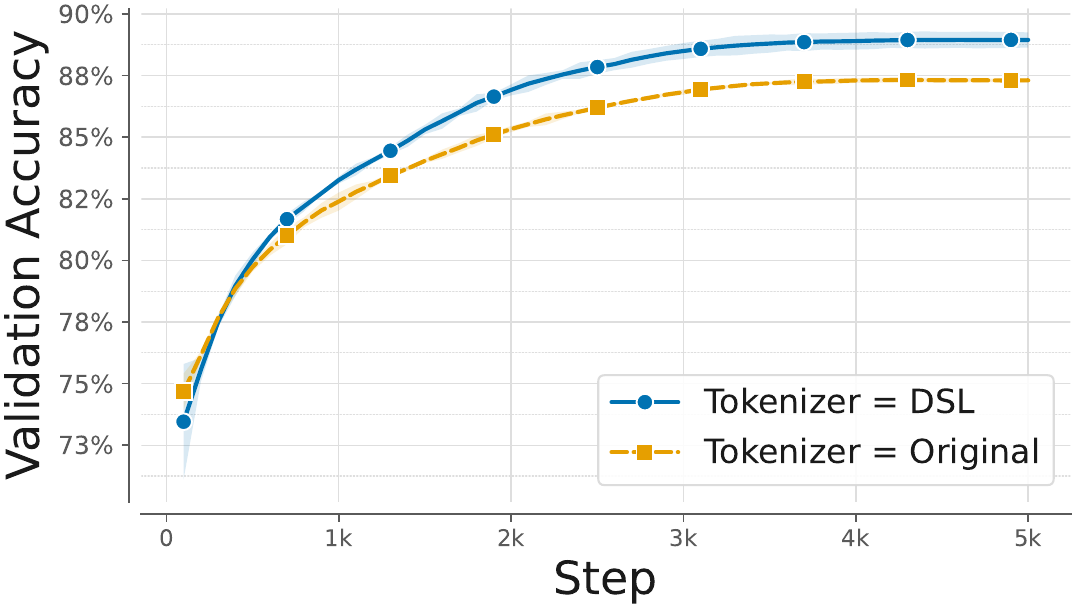}
        \caption{RM validation accuracy.}
    \end{subfigure}
    \caption{Tokenizer ablation for FLOORA-0.6B (Qwen3) over optimizer steps. Shaded regions show 95\% CIs across 3 training seeds, except pre-training which uses a single seed.}
    \label{fig:tokenizer_ablation_pt_sft}
\end{figure}

\begin{figure}[t!]
    \centering
    \begin{subfigure}[b]{0.45\textwidth}
        \centering
        \includegraphics[width=\textwidth]{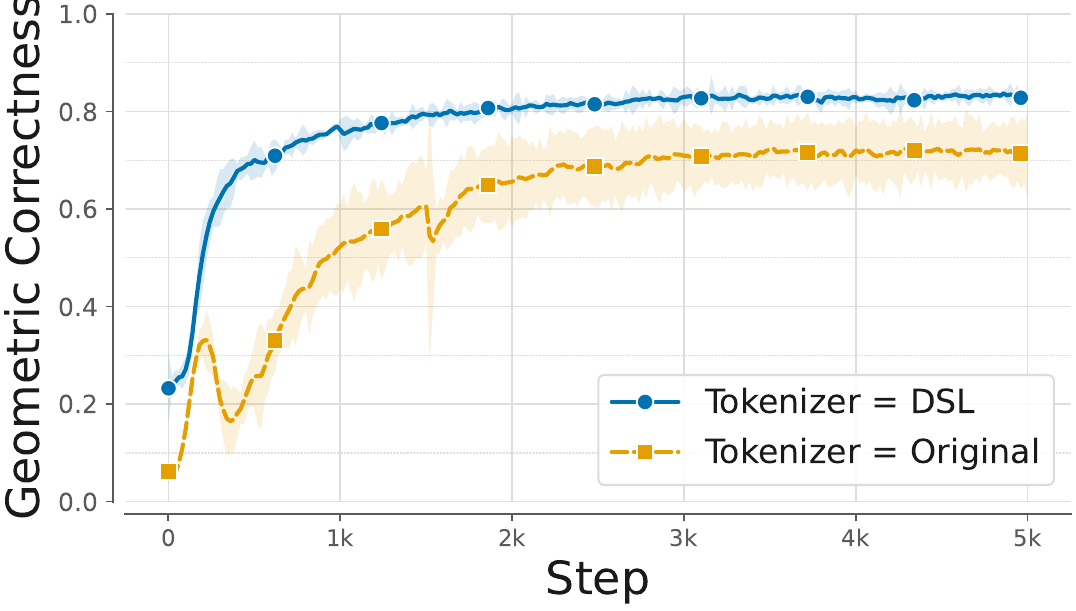}
        \caption{Geometric correctness reward curves.}
    \end{subfigure}
    \hfill
    \begin{subfigure}[b]{0.45\textwidth}
        \centering
        \includegraphics[width=\textwidth]{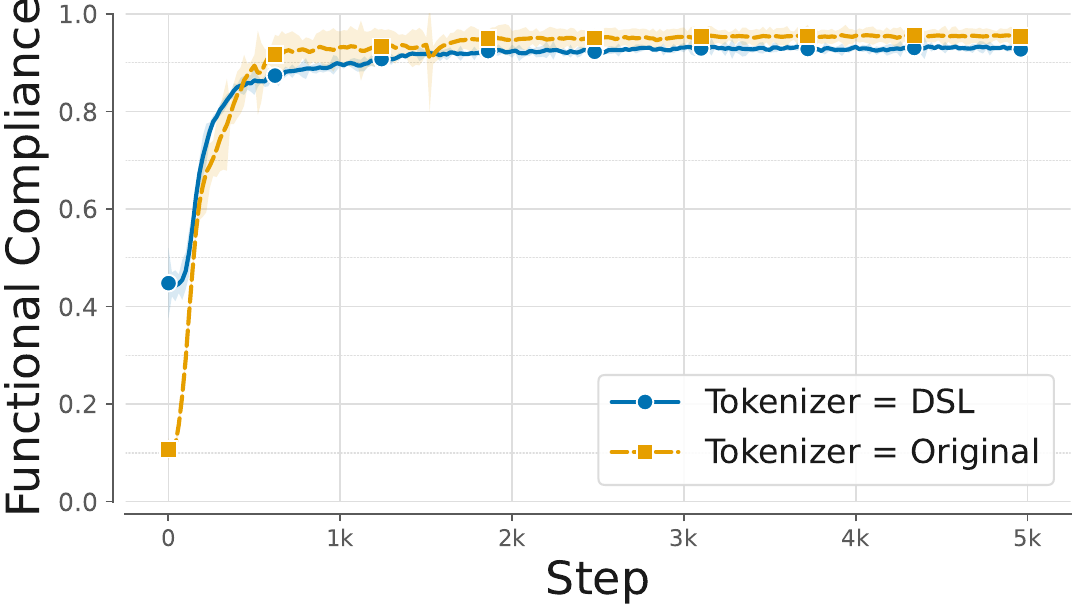}
        \caption{Functional compliance reward curves.}
    \end{subfigure}
    \vspace{1em}

    \centering
    \begin{subfigure}[b]{0.45\textwidth}
        \centering
        \includegraphics[width=\textwidth]{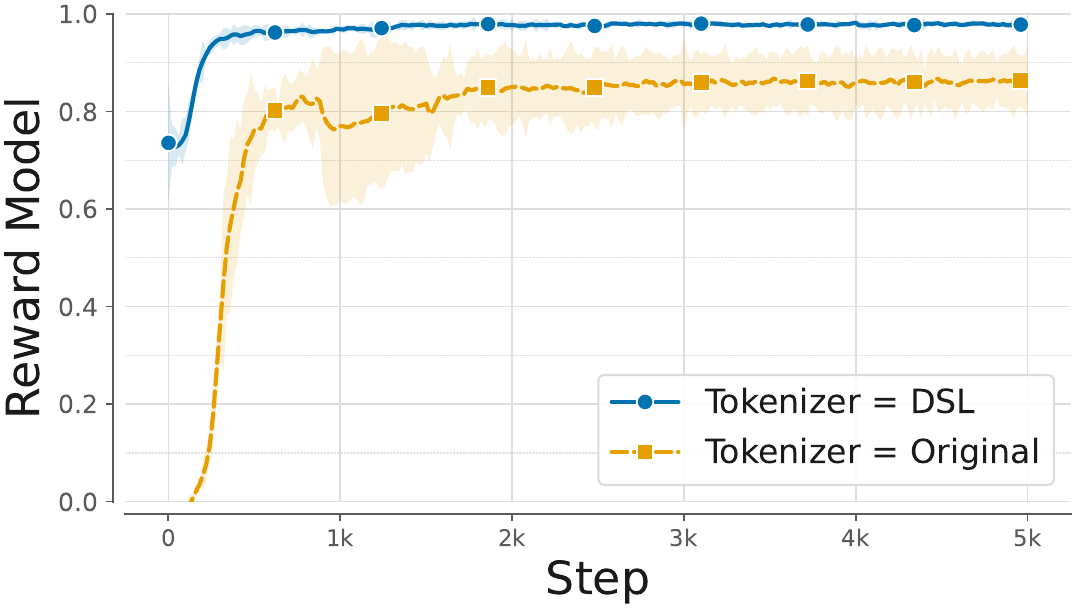}
        \caption{Reward model curves.}
    \end{subfigure}
    \hfill
    \begin{subfigure}[b]{0.45\textwidth}
        \centering
        \includegraphics[width=\textwidth]{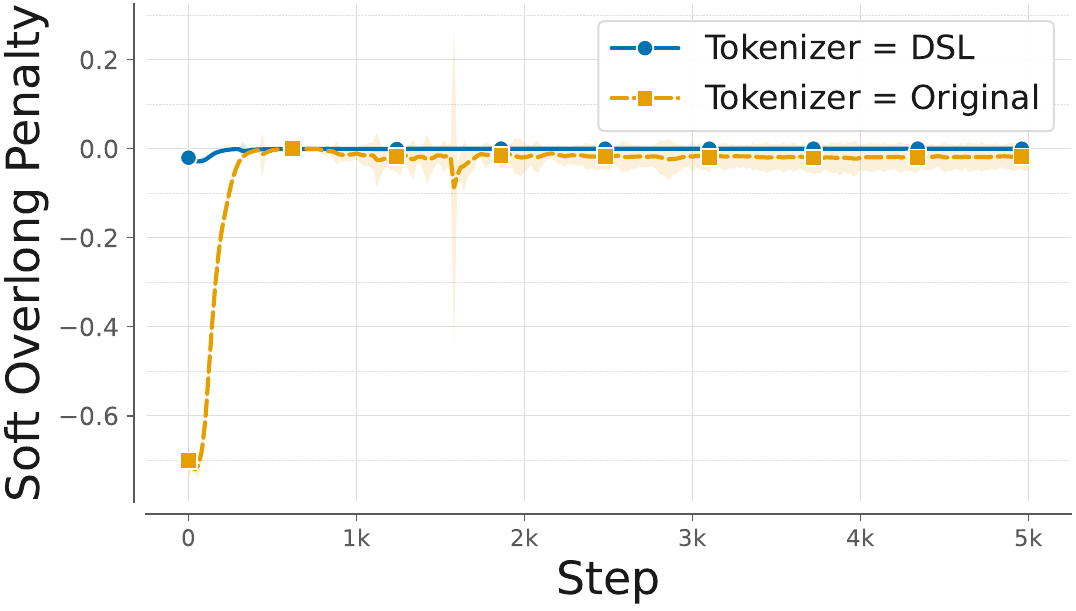}
        \caption{Soft overlong penalty curves.}
    \end{subfigure}
    
    \caption{Tokenizer ablation during GRPO (RM+VR) training with the FLOORA-0.6B (Qwen3) model. Shaded regions show 95\% CIs across 3 training seeds.}
    \label{fig:tokenizer_ablation_rl}
\end{figure}

\cref{fig:tokenizer_ablation_pt_sft,fig:tokenizer_ablation_rl} compare the training dynamics between the original and DSL tokenizers. Token-level loss and perplexity are not directly comparable across tokenizers because they segment sequences differently. For example, the original Qwen tokenizer represents digits individually, whereas the DSL tokenizer uses a single token for each quantized numeric value. Reward-model accuracy is tokenizer-independent and is higher with the DSL tokenizer.

Downstream results in \cref{tab:tokenizer_ablation_eval_osm,tab:tokenizer_ablation_eval_synthetic} show that the DSL tokenizer substantially improves OSM performance, with pass@5 gains of approximately 6.5-7.2 percentage points for the GRPO models. On the synthetic test set, the original tokenizer performs slightly better, but the largest pass@5 difference is only 0.9 percentage points.

\begin{table}[h!]
\centering
\scriptsize
\caption{Pass@1/3/5 performance of FLOORA-0.6B (Qwen3) across training stages using the original and DSL tokenizers on the \textbf{OSM} test set, reported with 95\% CIs across 3 training and 5 evaluation seeds. Best absolute value is shown in bold.}
\label{tab:tokenizer_ablation_eval_osm}
\begin{tabular}{l ccc}
\toprule
 & \multicolumn{3}{c}{\textbf{Total Reward}} \\
\cmidrule(lr){2-4}
\textbf{Model} & @1 & @3 & @5 \\
\midrule
FLOORA-0.6B (Qwen3) -- Base [DSL tok.] & 23.8 $\pm$ 0.1 & 33.4 $\pm$ 0.4 & 38.1 $\pm$ 0.7 \\
FLOORA-0.6B (Qwen3) -- Base [Orig. tok.] & 19.6 $\pm$ 0.4 & 27.3 $\pm$ 0.5 & 31.1 $\pm$ 0.5 \\
\midrule
FLOORA-0.6B (Qwen3) -- GRPO (RM+VR) [DSL tok.] & \textbf{79.3 $\pm$ 0.7} & \textbf{91.3 $\pm$ 0.5} & \textbf{94.0 $\pm$ 0.5} \\
FLOORA-0.6B (Qwen3) -- GRPO (RM+VR) [Orig. tok.] & 71.6 $\pm$ 2.9 & 83.2 $\pm$ 2.0 & 86.8 $\pm$ 1.6 \\
\midrule
FLOORA-0.6B (Qwen3) -- GRPO (VR only) [DSL tok.] & 74.7 $\pm$ 0.7 & 89.0 $\pm$ 0.5 & 92.5 $\pm$ 0.5 \\
FLOORA-0.6B (Qwen3) -- GRPO (VR only) [Orig. tok.] & 66.7 $\pm$ 0.5 & 81.5 $\pm$ 0.6 & 86.0 $\pm$ 0.6 \\
\bottomrule
\end{tabular}
\end{table}

\begin{table}[h!]
\centering
\scriptsize
\caption{Pass@1/3/5 performance of FLOORA-0.6B (Qwen3) across training stages using the original and DSL tokenizers on the \textbf{synthetic} test set, reported with 95\% CIs across 3 training and 5 evaluation seeds. Best absolute value is shown in bold.}
\label{tab:tokenizer_ablation_eval_synthetic}
\begin{tabular}{l ccc}
\toprule
 & \multicolumn{3}{c}{\textbf{Total Reward}} \\
\cmidrule(lr){2-4}
\textbf{Model} & @1 & @3 & @5 \\
\midrule
FLOORA-0.6B (Qwen3) -- Base [DSL tok.] & 90.2 $\pm$ 1.0 & 94.0 $\pm$ 1.0 & 94.9 $\pm$ 1.0 \\
FLOORA-0.6B (Qwen3) -- Base [Orig. tok.] & 88.9 $\pm$ 0.7 & 95.1 $\pm$ 0.6 & 96.2 $\pm$ 0.6 \\
\midrule
FLOORA-0.6B (Qwen3) -- GRPO (RM+VR) [DSL tok.] & 96.8 $\pm$ 0.2 & 98.1 $\pm$ 0.2 & 98.3 $\pm$ 0.2 \\
FLOORA-0.6B (Qwen3) -- GRPO (RM+VR) [Orig. tok.] & 93.5 $\pm$ 1.3 & 97.6 $\pm$ 0.5 & 98.4 $\pm$ 0.3 \\
\midrule
FLOORA-0.6B (Qwen3) -- GRPO (VR only) [DSL tok.] & \textbf{97.3 $\pm$ 0.1} & 98.1 $\pm$ 0.2 & 98.2 $\pm$ 0.2 \\
FLOORA-0.6B (Qwen3) -- GRPO (VR only) [Orig. tok.] & 95.2 $\pm$ 0.3 & \textbf{98.6 $\pm$ 0.1} & \textbf{99.1 $\pm$ 0.1} \\
\bottomrule
\end{tabular}
\end{table}

\subsection{Ablation on Human Feedback Data Scale}
\label[appendix]{app:ablation_feedback_scale}

As discussed in \cref{sec:post-training}, human feedback is collected in multiple rounds, with the collection model post-trained between rounds. We evaluate cumulative chronological subsets in increments of 1,500 examples, from 1,500 to 9,000, retraining SFT, the reward model, and GRPO for each subset. Because data quantity is coupled with collection stage and collection-model quality, this analysis reflects the practical collection trajectory rather than isolating the effect of data quantity alone. All subset sizes refer to the number of samples before rotation augmentation (see \cref{app:human_feedback_collection}).

\begin{figure}[h!]
    \centering
    \begin{subfigure}[b]{0.48\textwidth}
        \centering
        \includegraphics[width=\textwidth]{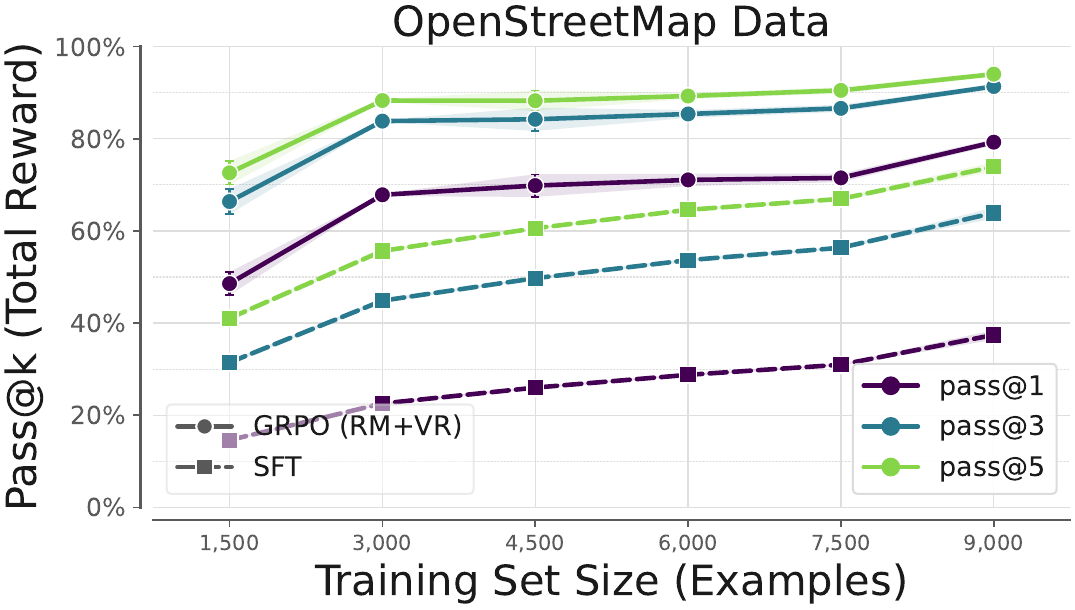}
        \caption{Pass@1/3/5 on the OSM test set.}
    \end{subfigure}
    \hfill
    \begin{subfigure}[b]{0.48\textwidth}
        \centering
        \includegraphics[width=\textwidth]{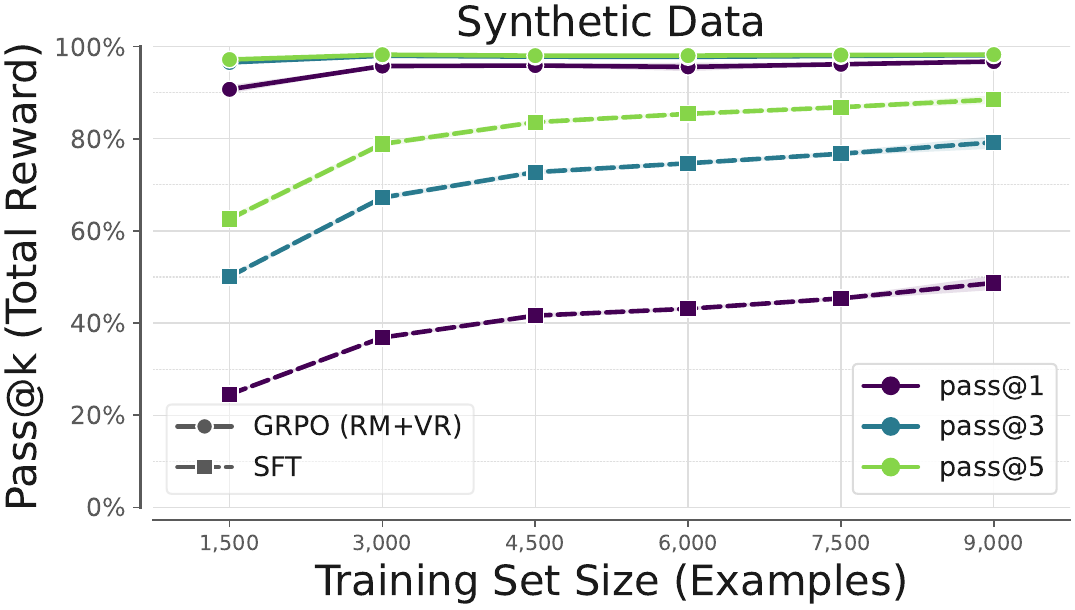}
        \caption{Pass@1/3/5 on the synthetic test set.}
    \end{subfigure}
    \caption{Human feedback data-scale ablation for FLOORA-0.6B (Qwen3) across SFT and GRPO (RM+VR) on the OSM and synthetic test sets. Shaded regions show 95\% CIs across 3 training seeds and 5 evaluation seeds.}
    \label{fig:data_ablation_app}
\end{figure}

As shown in \cref{fig:data_ablation_app}, performance generally improves across the collection trajectory for both SFT and GRPO. The largest gains occur at 3,000 and 9,000 examples, particularly on OSM, while synthetic performance saturates earlier. Overall, marginal gains diminish as the dataset grows, suggesting that tracking performance during collection can help identify when additional expert annotation provides limited benefit.



